\documentclass[fleqn,preprint,12pt]{elsarticle}
\usepackage{multirow}
\usepackage{amsmath}
\usepackage{amssymb}
\usepackage{amstext}
\usepackage{epsfig}
\usepackage{graphicx}
\usepackage{bm}
\usepackage{eufrak}
\usepackage{subfigure}
\usepackage{appendix}
\usepackage{color}
\usepackage{algorithm}
\usepackage{algorithmic}

\newcommand{\appsection}[1]{\let\oldthesection\thesection
  \renewcommand{\thesection}{Appendix \oldthesection}
  \section{#1}\let\thesection\oldthesection}

\begin{document}

\begin{frontmatter}

\title{Data-driven Reconstruction of Nonlinear Dynamics from Sparse Observation}

\author{Kyongmin Yeo${}^*$}
\cortext[cor]{Corresponding author}
\ead{kyeo@us.ibm.com}

\address{IBM T.J. Watson Research Center, Yorktown Heights, NY, USA}

\begin{abstract}
We present a data-driven model to reconstruct nonlinear dynamics from a very sparse times series data, which relies on the strength of the echo state network (ESN) in learning nonlinear representation of data. With an assumption of the universal function approximation capability of ESN, it is shown that the reconstruction problem can be formulated as a fixed-point problem, in which the trajectory of the dynamical system is a fixed point of the ESN. An under-relaxed fixed-point iteration is proposed to reconstruct the nonlinear dynamics from a sparse observation. The proposed fixed-point ESN is tested against both univariate and multivariate chaotic dynamical systems by randomly removing up to 95\% of the data. It is shown that the fixed-point ESN is able to reconstruct the complex dynamics from only 5 $\sim$ 10\% of the data. For a relatively simple non-chaotic dynamical system, the numerical experiments on a forced van der Pol oscillator show that it is possible to reconstruct the nonlinear dynamics from only 1$\sim$2\% of the data. 
\end{abstract}

\begin{keyword}
echo state network \sep reservoir computing \sep nonlinear dynamical system \sep time series \sep sparse observation \sep imputation
\end{keyword}

\end{frontmatter}

\section{Introduction}

Over the last decade, artificial neural network (ANN) has been extensively studied due to its strength in identifying complex nonlinear structure in data \cite{Bengio13,LeCun15,Schmidhuber15}. Recent studies on the application of ANN on the data-driven modeling of physical systems have shown promising results \cite{kutz17,Pathak18,RAISSI18,Yeo19}. In particular, in \cite{Lu17,Lu18}, it is shown that ANN is capable of predicting the dynamics of a chaotic attractor. When training an ANN for the modeling of a dynamical system, it is typical to use a sufficiently long time series data, so that the ANN fully explores the phase space. Given the huge amount of training data, it is unclear whether the long-term forecast capability of ANN is due to a high dimensional interpolation in a phase space, similar to the delay-coordinate embedding \cite{Packard80,Takens81}, or because ANN really learns the nonlinear dynamics. As a first step to answer this question, we first consider the problem, where only a partial observation of a time series data is available. We investigate if ANN can reconstruct the nonlinear dynamics from the incomplete information.

Reconstructing dynamics from an incomplete observation has practical applications in many physical, biological, and engineering problems, where the state of the system is accessible only through a sensor network. For example, in geophysics, it is not uncommon to find time series data, which contains a large amount of missing information or is irregularly sampled, due to sensor malfunction, atmospheric conditions, or physical limitations of the sensors \cite{Musial11,Shen15,Ozken18}. When a priori knowledge on the physical system is available, one of the standard approaches to reconstruct the nonlinear dynamics from an incomplete data set is to design a statistical model that incorporates the physical knowledge \cite{Gove06,MOFFAT07,OIKONOMOU18}. In \cite{Hamilton15}, it is shown that a high-resolution temporal dynamics can be reconstructed from a low-resolution time series data, which has a 10 times lower time resolution, by using only a subset of the governing equations. Incorporating the physics knowledge on the training of ANN, \cite{RAISSI18} proposed ``physics informed neural network'', which can recover the complex nonlinear dynamics from only a small fraction of the data.  

It is challenging to reconstruct nonlinear dynamics without having any prior knowledge on the underlying process. 
There are three major approaches to reconstruct the missing information; (a) identifying / modeling (auto-)correlation structures \cite{Sinopoli04,Kondrashov06,Pappas14}, (b) low-dimensional modeling \cite{Smith98,venturi04,RODRIGUES13}, and (c) pattern matching \cite{Phan17}. Although there is a significant progress in imputing the missing data, most of the classical approaches are based on strong assumptions, such as linearization, existence of a low-dimensional structure, or a priori knowledge on the correlation structures. The reconstruction accuracy of these models degrade quickly when the fraction of missing data is large \cite{Li18}, \emph{e.g.}, more than 50\%, and the underlying dynamical system is nonlinear.

Recently, a few ANN-based methods have been proposed for the data-driven reconstruction of the missing information by using recurrent neural networks (RNN). \cite{Lipton16} showed that simply filling in a missing period with the last known observation and using a missing data indicator as one of the input data to RNN can increase the prediction accuracy in a clinical binary classification problem. In \cite{Berglund15}, a Bidirectional-RNN based method is developed for a probabilistic reconstruction of the missing information for a binary time series data. \cite{Che18} proposed an unidirectional RNN model, which assumes a relaxation of the unobserved dynamics towards an empirical mean of the data. Based on the relaxation model of \cite{Che18}, \cite{Cao18} and \cite{Luo18} developed Bi-directional RNN and Generative Adversarial Network models, respectively. Those ANN models showed better performance compared to the conventional statistical imputation approaches for the applications in the health care domain. However, it should be noted that the relaxation towards an empirical mean is essentially based on the assumption of a stationary Markov process, which does not consider temporal structures or dynamics of the underlying process.
 
In this study, we develop a novel RNN model for the data-driven reconstruction of nonlinear dynamics from temporally sparse observations. For the modeling of nonlinear dynamics, the echo state network (ESN) \cite{Jaeger04} is employed. The reconstruction problem is formulated as a fixed-point problem. The behavior of the proposed fixed-point ESN is thoroughly studied by using chaotic time series data. It is shown that ESN can reconstruct the chaotic, nonlinear dynamics from as little as 5\% of the data. This paper is organized as follows; the echo state network is reviewed in section \ref{sec:esn_basic}. In \ref{sec:numeric}, a fixed-point iteration method is proposed. Numerical experiments of the fixed-point ESN model is presented in section \ref{sec:experiments}. Finally, conclusions are given in section \ref{sec:conclusions}.

\section{Echo State Network}

\subsection{Review of Echo State Network} \label{sec:esn_basic}

The echo state network (ESN) has been studied extensively due to its strength in learning complex dynamics from time series data \cite{Pathak18,Lu17,Inubushi17,Antonik18}. ESN is a nonlinear state-space model, of which latent state consists of a set of randomly generated dynamical systems. Following \cite{Jaeger04,LUKOSEVICIUS12}, the evolution equation of the latent state is
\begin{equation} \label{eqn:esn-internal}
\bm{s}_{t+1} = \lambda \bm{s}_t + (1-\lambda)\Psi(\bm{s}_t,\bm{x}_t).
\end{equation}
Here, $\bm{s}_t\in\mathbb{R}^{N_s}$ is the latent state of ESN, $\bm{x}_t\in\mathbb{R}^{N_x}$ is a vector of the input signal to ESN, $\lambda \in [0,1)$ is a temporal relaxation coefficient, and $\Psi(\cdot)$ is a nonlinear function. The subscript $t$ denotes a time stamp, $\phi_t = \phi(t \delta t)$, where $\delta t$ denotes a time step size, or data sampling interval.  The input signal may consist of the observation of the physical system, $\bm{y}_t\in\mathbb{R}^{N_y}$, and an exogenous forcing or auxiliary input, $\bm{u}_t\in\mathbb{R}^{N_u}$, such that $\bm{x}_t = (\bm{y}_t,\bm{u}_t)$. For the nonlinear function, usually a hyperbolic tangent function is used;
\begin{equation} \label{eqn:esn-internal-1}
\Psi(\bm{s}_t,\bm{x}_t) = \tanh(\bm{A}\bm{s}_t+\bm{B}\bm{x}_t).
\end{equation}
Here, $\bm{A}\in\mathbb{R}^{N_s\times N_s}$ and $\bm{B}\in\mathbb{R}^{N_s\times N_x}$ are randomly generated weight matrices. Independent uniform random variables are used to generate $\bm{A}$ and $\bm{B}$ \cite{Jaeger04}, e.g., $A_{ij},\,B_{ij} \sim \mathcal{U}(-\xi,\xi)$, in which $\mathcal{U}(a,b)$ denotes a uniform random distribution in $(a,b)$. It is worthwhile to note that the evolution equation of the standard recurrent neural network (RNN) has a very similar structure with (\ref{eqn:esn-internal}--\ref{eqn:esn-internal-1}). One of the most significant differences between RNN and ESN is that, in ESN, the weight matrices in (\ref{eqn:esn-internal-1}) are generated randomly, while they are computed by a maximum likelihood method in RNN. The weight matrix, $\bm{A}$, decides the connectivity between the latent state, $\bm{s}$. It is suggested that making the connection sparse results in a richer internal dynamics \cite{Jaeger04}. Typically,  $\bm{A}$ has only 1 $\sim$ 2\% of nonzero elements. 
A more detailed explanation about generating $\bm{A}$ is given in section \ref{sec:experiments}.

The echo state network solves an initial value problem. While it is straightforward to specify the initial condition of the physical variables, e.g., $\bm{y}_0$ and $\bm{u}_0$, it is challenging to find the correct initial condition of the latent state, $\bm{s}_0$. In \cite{Jaeger04}, it is proposed that an ESN must have the ``echo state property'', which is similar to a ``fading memory effect'' \cite{LUKOSEVICIUS12,Jaeger01}. 
For $\lambda > 0$, it is straightforward to show the fading memory effect. Observe that (\ref{eqn:esn-internal}) can be rewritten as
\begin{equation} \label{eqn:esn-euler}
\frac{\bm{s}_{t+1}-\bm{s}_t}{\delta t} = \frac{1-\lambda}{\delta t}\left( - \bm{s}_t + \bm{f}_t \right),
\end{equation}
in which $\bm{f}_t = \Psi(\bm{s}_t,\bm{x}_t)$. If we choose $\lambda = 1- \gamma \delta t$ for $0 < \gamma < 1/\delta t$, it is clear that the evolution equation of the latent state, (\ref{eqn:esn-internal}), is the forward Euler approximation to the system of relaxation equations,  
\begin{equation}\label{eqn:esn-relax}
\gamma^{-1} \frac{ d \bm{s}}{dt} = -\bm{s}(t)+\bm{f}(t).
\end{equation}
Since $|\bm{f}(t)| < \infty$,  (\ref{eqn:esn-relax}) implies that the effects of the misspecification of $\bm{s}_0$ will vanish after some initial transient period. It is also clear that ESN relies on a set of randomly generated relaxation processes to model the nonlinear behaviors in the time series data. 
When $\lambda = 0$, the ``echo state property'' can be guaranteed by making the spectral radius of $\bm{A}$ smaller than unity, i.e., $\rho(\bm{A}) < 1$ \cite{LUKOSEVICIUS12}.

Once the latent state of ESN is updated by (\ref{eqn:esn-internal}), the physical process at the next time step ($t+\delta t$) is computed by a linear projection of the latent state onto the physical space, 
\begin{equation}
\bm{y}_{t+1} = \bm{\theta}^T\bm{S}_{t+1},
\end{equation}
in which $\bm{S}_{t+1}\in\mathbb{R}^{N_s+1}$ is an augmented latent state, $\bm{S}_t = (1,\bm{s}^T_t)^T$ and $\bm{\theta}\in\mathbb{R}^{(N_s+1)\times N_y}$ is a linear map. 
Conceptually, ESN is similar to a ``kernel method'' in the machine learning literature \cite{Bishop06}, where a nonlinear projection is used to map the data into a high dimensional space with a goal of finding a linear pattern in the feature space. It should be noted that, although this class of methods has been successfully applied to solve complex problems in practice, it lacks theoretical analysis, such as the optimality of the projection and the convergence. 

In the standard ESN, ``model training'' is to find the linear map, $\bm{\theta}$, from the data. Let $\bm{Y}=(\bm{y}_0,\cdots,\bm{y}_N)$ be a time series data of length $N+1$. The linear map is computed by solving the following regularized optimization problem;
\begin{equation}\label{eqn:OLS}
\min_{\bm{\theta}} \sum_{i=1}^N \frac{1}{2}\| \bm{y}_i -  \bm{\theta}^T\bm{S}_i \|^2_2 + \frac{\beta}{2} \| \bm{\theta} \|^2_F,
\end{equation}
in which $\|\cdot\|_F$ denotes the Frobenius norm, and $\beta$ is a regularization parameter. The analytical solution of (\ref{eqn:OLS}) is
\begin{equation}\label{eqn:OLS-sol}
\bm{\theta} = \left( \sum_{i=1}^N \bm{S}_i\bm{S}_i^T + \beta\bm{I}\right)^{-1} \left( \sum_{i=1}^N\bm{S}_i \bm{y}_i^T\right).
\end{equation}
The second term in (\ref{eqn:OLS}) corresponds to the standard $L_2$ regularization, often called the Tikhonov regularization. In ESN, first the physical variable, $\bm{x}\in\mathbb{R}^{N_x}$, is projected onto a high dimensional  ($\mathbb{R}^{N_s}$) latent space by a nonlinear map, $\Psi$, where $N_s \gg N_x$. Because $\Psi$ is generated randomly, there is no guarantee that $\bm{s}_t$ is linearly independent. Hence, the $L_2$ regularization is usually required to deal with the possible multicollinearity.

\begin{table}
\center{
\caption{Free parameters of ESN.} 
\label{tbl:free_param}
\begin{tabular}{cc}
\hline \hline
$\lambda$ & Temporal relaxation parameter\\
$\beta$ & $L_2$-regularization parameter\\
$\nu$ & Fraction of nonzero elements of $\bm{A}$\\
$\rho_{max}$ & Spectral radius of $\bm{A}$\\
$\xi_A$ & $A_{ij} \sim \mathcal{U}(-\xi_A,\xi_A)$ \\
$\xi_B$ & $B_{ij} \sim \mathcal{U}(-\xi_B,\xi_B)$ \\
\hline \hline
\end{tabular}
}
\end{table}

In summary, an ESN consists of the following update equations;
\begin{align}
\bm{s}_{t+1} &= \lambda \bm{s}_t + (1-\lambda)\Psi(\bm{s}_t,\bm{x}_t),~\Psi(\bm{s}_t,\bm{x}_t) = \tanh (\bm{A}\bm{s}_t+\bm{B}\bm{x}_t), \nonumber\\
\bm{y}_{t+1} &= \bm{\theta}^T\bm{S}_{t+1}, \hspace{2.8cm} \bm{S}_{t+1} = (1,\bm{s}^T_t)^T. \nonumber
\end{align}
The free parameters of an ESN are listed in Table \ref{tbl:free_param}.

\subsection{Fixed-point ESN for Sparse Observation} \label{sec:numeric}

\begin{figure}
  \centering
  \includegraphics[width=0.8\textwidth]{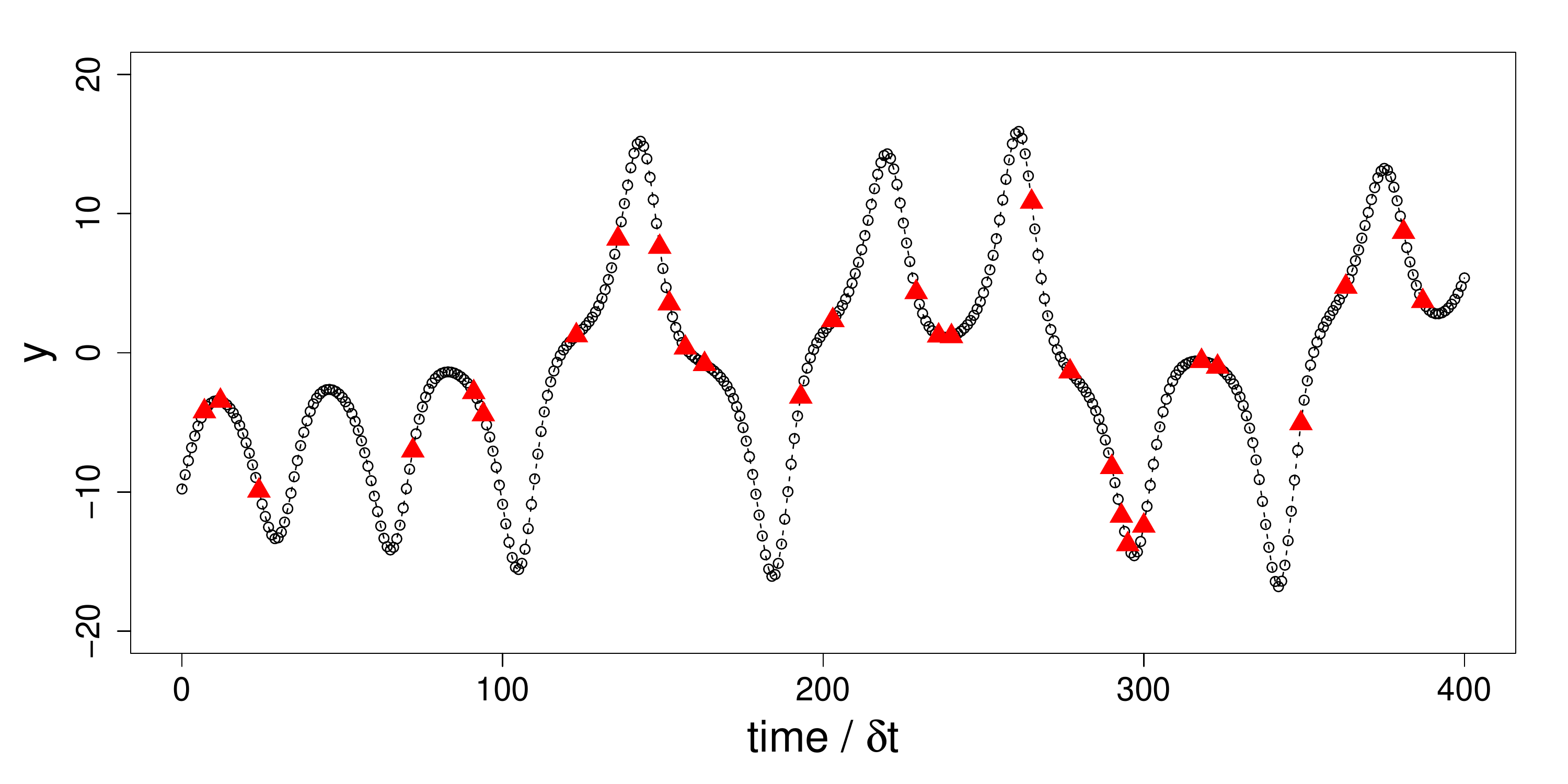}
  \caption{Example of a sparse observation. The hollow circles ($\circ$) correspond to the full observation ($\bm{Y}^*$) and the sparse observation ($\bm{Y}^O$) is denoted by the solid triangles ({\color{red}$\blacktriangle$}). The ground truth is shown as a dashed line. In this example, the missing fraction is $\omega = 0.95$. } \label{fig:sparse_example}
\end{figure}

Here, we consider the time series data generated by a nonlinear dynamical system,
\begin{equation}
\frac{d \bm{y}}{dt} = \mathcal{F}(\bm{y},\bm{u}),~~t \in [0,T\delta t ],
\end{equation}
in which $\mathcal{F}(\cdot)$ is an unknown time marching operator and $\delta t$ is the sampling interval of the observation. Let $\bm{Y}^*=(\bm{y}^*_0,\cdots,\bm{y}^*_T)$ be the full observation obtained by an equidistance sampling of $\bm{y}(t)$, i.e.,
\[
y^*_{l_j} = y_j(l \delta t),~~\text{for}~~l = 0,\cdots,T, ~\text{and}~~ j = 1,\cdots,N_y. 
\]
We assume that the full observation, $\bm{Y}^*$, is not accessible, and only a partial observation of $\bm{Y}^*$, which is denoted by $\bm{Y}^O$, is known. The missing (unobserved) elements of $\bm{Y}^O = (\bm{y}^O_0,\cdots,\bm{y}^O_T)$ are marked by an arbitrarily large number, $\eta$, i.e.,
\[
y^O_{l_j} = \begin{cases} y^*_{l_j}& \text{if observed,} \\ \eta & \text{otherwise.} \end{cases}
\]
Based on $\bm{Y}^O$, we can define an indicator set of the observation, $\bm{M} = (\bm{m}_0,\cdots,\bm{m}_T)$, in which
\[
m_{l_j} = \begin{cases} 0 & \text{if}~y^O_{l_j} = \eta, \\ 1 & \text{otherwise}.  \end{cases}
\]
Here, we are interested in reconstructing $\bm{Y}^*$ from a sparse observation, $\bm{Y}^O$, where the fraction of the missing data,
\begin{equation} \label{eqn:omega}
\omega = \frac{1}{T \times N_y} \sum_{t=1}^T \sum_{j=1}^{N_y} (1-m_{t_j}),
\end{equation}
is larger than 0.5. We assume that the exogenous forcing, $\bm{U}=(\bm{u}_0,\cdots,\bm{u}_T)$, is fully known. Figure \ref{fig:sparse_example} shows an example of such a sparse observation.

The standard method of training an ANN, so called the ``teacher forcing'' method, consists of three steps; 1) provide an input data $\bm{x}_{t-1}$, 2) make a prediction, $\hat{\bm{y}}_t$, and 3) compare the difference between $\hat{\bm{y}}_t$ and $\bm{y}^*_t$ at every time step, i.e., for $t = 1,\cdots,T$. 
However, when there are missing entries in the data set, such sequential process is not possible. One of the conventional approaches to circumvent the problem is to fill the missing entries by using a statistical imputation method and to train an RNN against the imputed data set, assuming that the imputed data is close to $\bm{Y}^*$ and has no systemic bias. When the missing fraction is small, e.g., $\omega < 0.2 \sim 0.3$, the imputation-and-training method works fairly well. For a very high missing fraction, e.g., $\omega > 0.8$, the standard imputation methods become increasingly unreliable. 

To develop an ESN for the reconstruction of complex dynamics from a sparse observation, we exploit the capability of ESN as a universal function approximator \cite{Schrauwen07,Funahashi93,GRIGORYEVA18}. We assume that, for a large enough $N_s$, there is an ESN, which can accurately computes the time series data, $\bm{Y}^*$, such that
\begin{equation} \label{eqn:y_star-1}
\bm{y}^*_t = {\bm{\theta}^*}^T\bm{S}^*_t, ~~\text{for}~~t=1,\dots,T
\end{equation}
and
\begin{equation} \label{eqn:y_star-2}
\bm{s}^*_t =  \lambda \bm{s}^*_{t-1} + (1-\lambda)\tanh(\bm{A} \bm{s}^*_{t-1} + \bm{B}_y\bm{y}^*_{t-1} + \bm{B}_u \bm{u}_{t-1}),
\end{equation}
in which $\bm{B}_y\in\mathbb{R}^{N_s\times N_y}$ and $\bm{B}_u\in\mathbb{R}^{N_s\times N_u}$ are the submatrices of $\bm{B}$, which correspond to $\bm{y}$ and $\bm{u}$, respectively. We can rewrite (\ref{eqn:y_star-1} -- \ref{eqn:y_star-2}) as
\begin{equation}\label{eqn:fixed_Y}
\bm{Y}^* = \bm{\mathcal{E}}(\bm{Y}^*,\bm{U}),
\end{equation}
where
\begin{align}
\mathcal{E}_0(\bm{Y}^*,\bm{U}) &= \bm{y}^*_0,\\
\mathcal{E}_l(\bm{Y}^*,\bm{U}) &= {\bm{\theta}^*}^T\bm{S}^*_l~~\text{for}~~l = 1,\cdots,T.
\end{align}
Note that $\bm{S}^*_l$ is a function of $(\bm{y}^*_0,\cdots,\bm{y}^*_{l-1},\bm{u}_0,\cdots,\bm{u}_{l-1})$ and $\bm{\theta}^*$ depends on $(\bm{Y}^*,\bm{U})$ through (\ref{eqn:OLS}). It is clear from (\ref{eqn:fixed_Y}) that $\bm{Y}^*$ is a fixed point of the nonlinear operator, $\bm{\mathcal{E}}(\bm{\cdot},\bm{U})$.

\begin{algorithm}[t]
	\caption{Fixed-point iteration for the nonlinear imputation}
	\label{alg:fixed_point}
	\hspace*{\algorithmicindent} \textbf{Input}: $\bm{Y}^O$, $\bm{M}$, tolerance level ($\epsilon$), maximum iteration ($k_{max}$)\\
	\hspace*{\algorithmicindent} \textbf{Output}: $\bm{Y}^R$ 
	\begin{algorithmic}
		\STATE Set an initial condition, $\bm{Y}^{R,\,0}$, by a linear interpolation
		\STATE
		\STATE Set $k = 0$	
		\WHILE{ ($e^k > \epsilon$) and ($k < k_{max}$)} 
			\STATE Compute $\mathcal{S}^k$ from $\bm{Y}^{R,\,k}$
			\STATE Compute $\bm{\theta}^k$:
			\[
			\bm{\theta}_j^{k} = \left( \sum_{t=T_s}^T m_{t_j}\bm{S}^{k}{\bm{S}^{k}}^T + \beta \bm{I} \right)^{-1} \left( \sum_{t=T_s}^T m_{t_j} y^O_{t_j}\bm{S}^k_t \right),~\text{for}~j=1,\cdots,N_y.
			\]
			\STATE Update $\bm{Y}^{R,\,k+1}$:
			\[
			\bm{Y}^{R,\,k+1} = \alpha \bm{Y}^{R,\,k} + (1-\alpha) \mathcal{E}(\bm{Y}^{R,\,k},\bm{U})
			\]
			\STATE Compute the $l_2$-improvement
			\[
			e^k = \left[ \frac{1}{N_y T}\sum_{j=1}^T \|\bm{y}^{R,\,k+1}_j -\bm{y}^{R,\,k}_j\|^2_2 \right]^{1/2}  
			\]
			\STATE $k = k + 1$
		\ENDWHILE
	\end{algorithmic}
\end{algorithm}

Now, we define a ``reconstructed'' observation vector,
\begin{equation} \label{eqn:yR}
\bm{y}^R_t = (\bm{1}-\bm{m}_t) \circ \hat{\bm{y}}_t + \bm{m}_t \circ\bm{y}^O_t,
\end{equation}
in which $\circ$ is the Hadamard product, $\bm{1} \in \mathbb{R}^{N_y}$ is a vector of ones, and $\hat{\bm{y}}_t$ is  a model prediction, or reconstruction. Simply, $\bm{y}^R_t$ takes the observation if available, otherwise the missing entry is filled by the output of the ESN. 
From (\ref{eqn:fixed_Y}), the goal of the nonlinear reconstruction of the dynamics is equivalent to finding $\bm{Y}^R=(\bm{y}^R_0,\cdots,\bm{y}^R_T)$ and $\bm{\theta}$, which satisfy
\begin{equation}
\bm{Y}^R = \bm{\mathcal{E}}(\bm{Y}^R,\bm{U}).
\end{equation}
Hence, we propose the following fixed-point iteration method,
\begin{equation} \label{eqn:esn_fixed}
\bm{Y}^{R,\,k+1} = \bm{\mathcal{G}}(\bm{Y}^{R,\,k},\bm{U}) = \alpha \bm{Y}^{R,\,k} + (1-\alpha)\bm{\mathcal{E}}(\bm{Y}^{R,\,k},\bm{U}),~~\text{for}~k\ge0,
\end{equation}
in which $0 < \alpha < 1$ is an under-relaxation parameter of the fixed point iteration, and the superscript $k$ is the iteration count.
The under relaxation is required to guarantee that the Jacobian of $\bm{\mathcal{G}}$ exists (see \ref{eqn:Jacobian_G}). It is trivial to show that $\bm{Y}^* = \bm{Y}^R$ upon convergence.

The linear map at $k$-th iteration is computed by solving the following optimization problem,
\begin{equation} \label{eqn:opt_fixed}
\bm{\theta}^k=\min_{\bm{\theta}} \sum_{t=T_s}^T \frac{1}{2}\| \bm{m}_t\circ(\bm{y}^O_t -  {\bm{\theta}}^T\bm{S}^k_t) \|^2_2 + \frac{\beta}{2} \| \bm{\theta}^k \|^2_F.
\end{equation}
In other words, $\bm{\theta}^k$ is computed by comparing only with the available observation data. Note that the time index starts from $T_s$. As discussed in section \ref{sec:esn_basic}, we cannot specify the correct initial condition of the latent state, $\bm{s}_0$. Hence, we disregard the ESN outputs in  the first $T_s-1$ time steps to remove the effects of the initial condition. The analytical solution of (\ref{eqn:opt_fixed}) is
\begin{equation} \label{eqn:opt_fixed_sol}
\bm{\theta}_j^{k} = \left( \sum_{t=T_s}^T m_{t_j}\bm{S}^{k}{\bm{S}^{k}}^T + \beta \bm{I} \right)^{-1} \left( \sum_{t=T_s}^T m_{t_j} y^O_{t_j}\bm{S}^k_t \right), ~~\text{for}~j=1,\cdots,N_y.
\end{equation}
Here, $\bm{\theta}_j$ is the $j$-th column of $\bm{\theta}$. 
Note that after a proper rescaling, the summations in the original ESN formulation (\ref{eqn:OLS-sol}) are time averaging, while the summations in (\ref{eqn:opt_fixed_sol}) correspond to an ensemble averaging, or a Monte Carlo sampling. Hence, when the dynamical system is ergodic, the time series data is missing in random, and the number of observation, $\omega T$, is large enough, the linear map of the fixed-point method, $\bm{\theta}^k$, approaches to $\bm{\theta}^*$ as the fixed-point iteration converges.


The fixed point iteration procedure is outlined in Algorithm \ref{alg:fixed_point}. Computing $\bm{Y}^{R,\,k+1}$ involves updating the ESN twice over the entire time series ($t = 0 \sim T$); first to compute $\bm{\theta}^k$ and then to update $\bm{Y}^{R,\,k+1}$. It is possible to reduce the computation cost by storing $\mathcal{S}^k=(\bm{S}^k_1,\cdots,\bm{S}^k_T)$ in the first passage and updating $\bm{Y}^R$ by computing only the projection ${\bm{\theta}^k}^T\mathcal{S}^k$. However, this approach requires a huge memory space. For example, when $T \sim O(10^4)$ and $N_s \sim O(10^3)$, storing $\mathcal{S}^k$ with double precision requires a few gigabytes of memory. 

One of the advantages of a fixed point iteration method is that it does not require derivative information of the nonlinear operator, $\mathcal{G}$. Computing the derivatives requires an additional storage space of $O(N_s \times T)$ and has computational complexity of $O(N_s^2 \times T^2)$. However, it is challenging to analyze the convergence of a fixed point iteration and it usually exhibits a linear convergence. A theoretical analysis of the proposed fixed point iteration is provided in \ref{sec:convergence}. 

\section{Numerical experiments} \label{sec:experiments}

In this section, a series of numerical experiments on the reconstruction of nonlinear dynamical systems is presented. The same ESN parameters are used across all the numerical experiments, except for $\beta$;
\begin{equation} \label{eqn:ref_param}
\lambda = 0.6, ~ \nu = 0.01,~ \rho_{max} = 0.9, ~\xi_A = 1,~\xi_B = 0.4, ~\text{and}~T_s = 200.
\end{equation}
The size of the ESN is also fixed at $N_s = 10^3$. To generate the sparse matrix $\bm{A}$, fist, $\nu\times N_s^2 = 10^4$ elements of $\bm{A}$ are randomly selected and filled by sampling from the uniform distribution, $\mathcal{U}(-\xi_A,\xi_A)$. Then, $\bm{A}$ is scaled by multiplying a constant to make $\rho(\bm{A}) = \rho_{max}$. 
As shown in \cite{Lu17}, the accuracy of an ESN depends on the choice of those free parameters. However, we do not make an attempt to fine tune those parameters, because, without having a priori knowledge on the dynamical system, hand-tuning those parameters with a sparse observation is impractical. 

\subsection{Mackey--Glass Time Series} \label{sec:MG}

\begin{figure}
  \centering
  \includegraphics[width=0.48\textwidth]{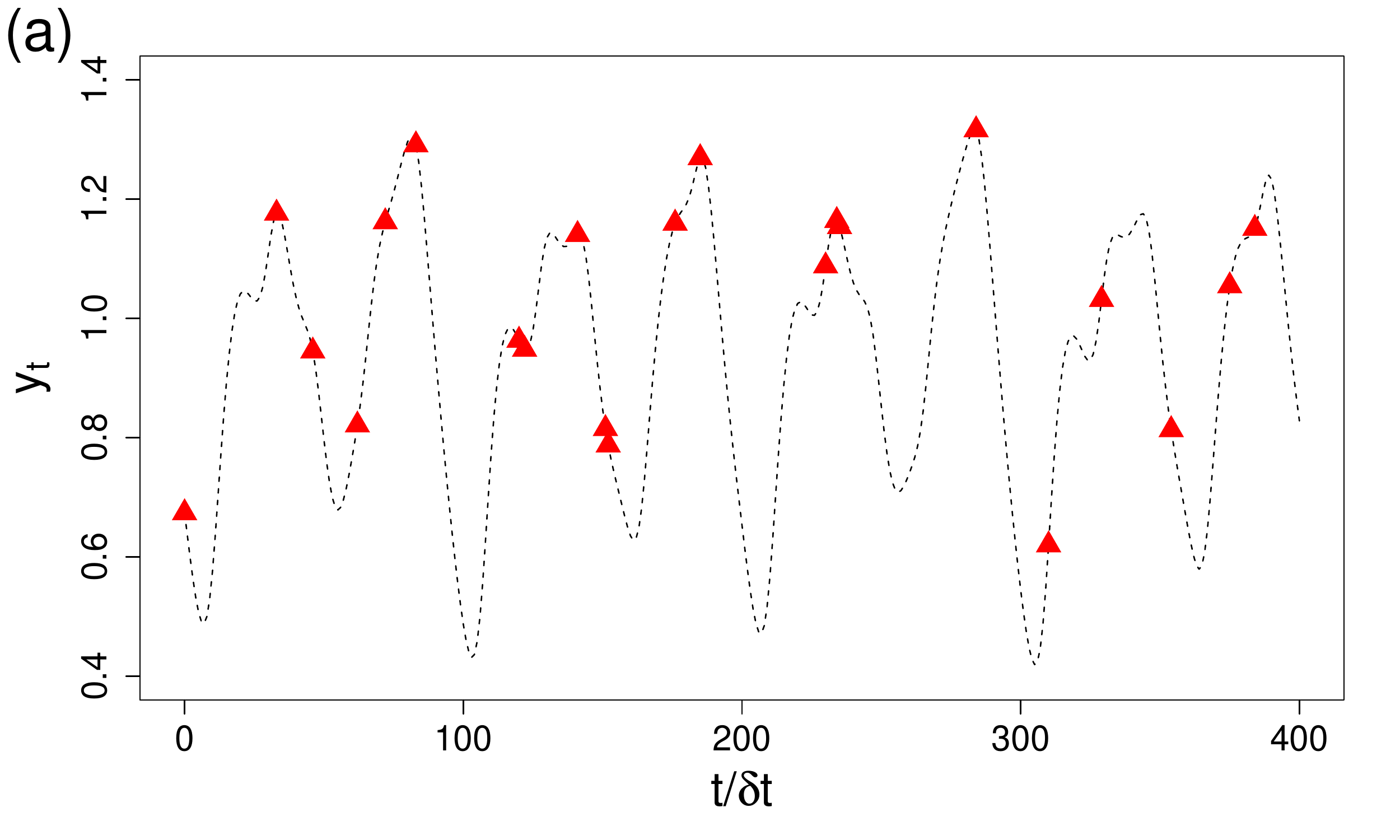}
  \includegraphics[width=0.48\textwidth]{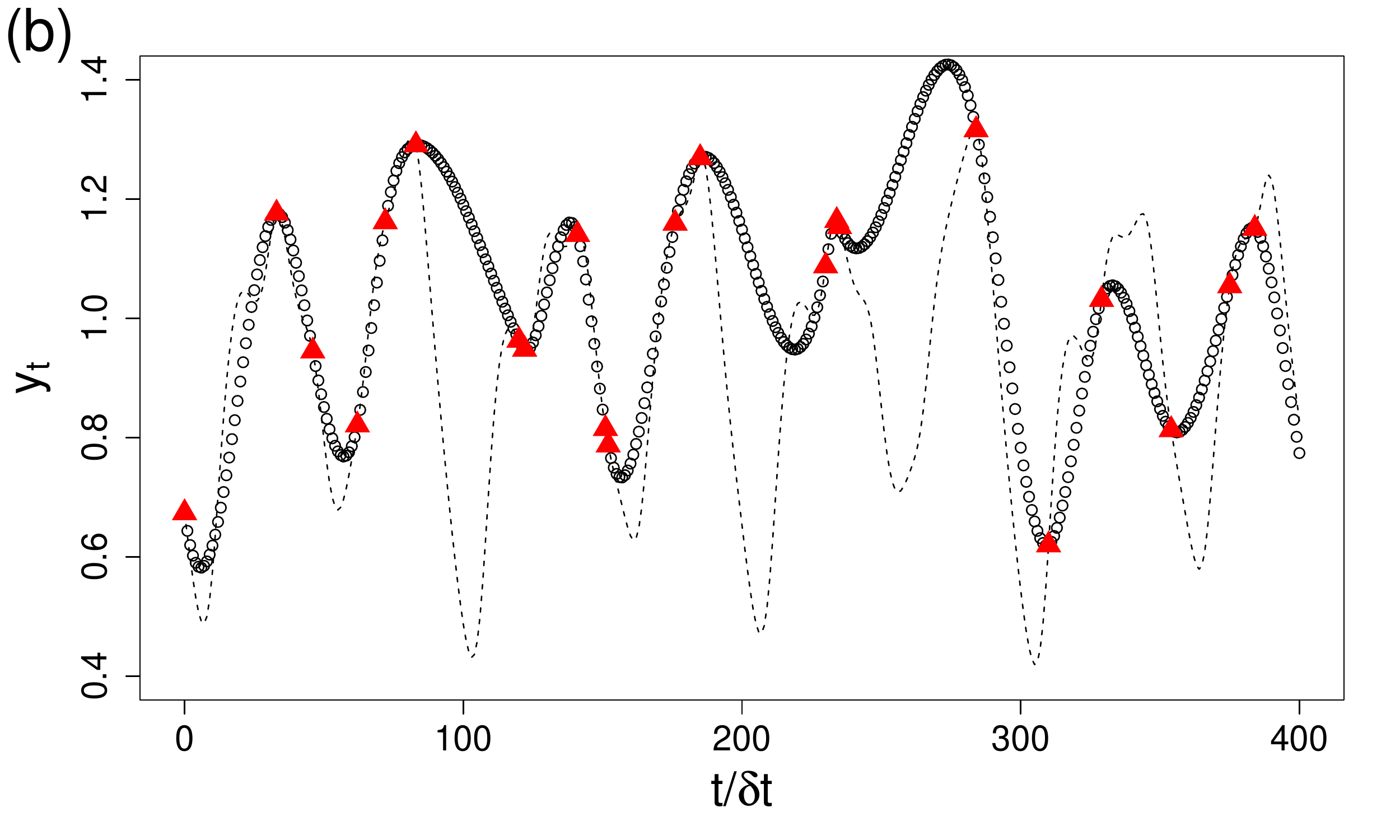}
  \caption{Sparse Mackey-Glass time series for the missing fraction $\omega = 0.95$. The dashed line is the ground truth ($\bm{Y}^*$) and the sparse observation ($\bm{Y}^O$) is denoted by the solid triangles ({\color{red}$\blacktriangle$}). In (b), the hollow circles ($\circ$) are the cubic spline interpolation of $\bm{Y}^O$.} \label{fig:MG_data}
\end{figure}

First, the behavior of the proposed fixed-point ESN is investigated by using a sparse observation of a univariate time series. The ground truth is generated by a discrete sampling of the Mackey-Glass delay-time dynamical system with a sampling interval of $\delta t = 1$. The Mackey-Glass equation is \cite{Mackey77},
\begin{equation} \label{eqn:MG}
\frac{d y(t)}{dt} = \frac{\gamma_1 y(t-\tau)}{1+y^{\gamma_2}(t-\tau)} - \gamma_3 y(t),
\end{equation}
where we use $\gamma_1 = 0.2$, $\gamma_2 = 10$, $\gamma_3 = 0.1$, and $\tau = 17$. For this set of parameters, the Mackey-Glass time series becomes chaotic \cite{Farmer82}.  The length of the time series is $T = 5\times10^4$. The sparse observation is generated by randomly removing $\omega$ fraction of the data. Figure \ref{fig:MG_data} (a) shows the sparse observation when 95\% of the time series data is  removed. In general, it is very challenging to reconstruct the missing information of a univariate time series \cite{Phan17}. Without a prior knowledge, the usual practice is to use an interpolation. However, as shown in figure \ref{fig:MG_data} (b), a standard cubic spline interpolation results in a poor estimation. 

\begin{figure}
  \centering
  \includegraphics[width=0.48\textwidth]{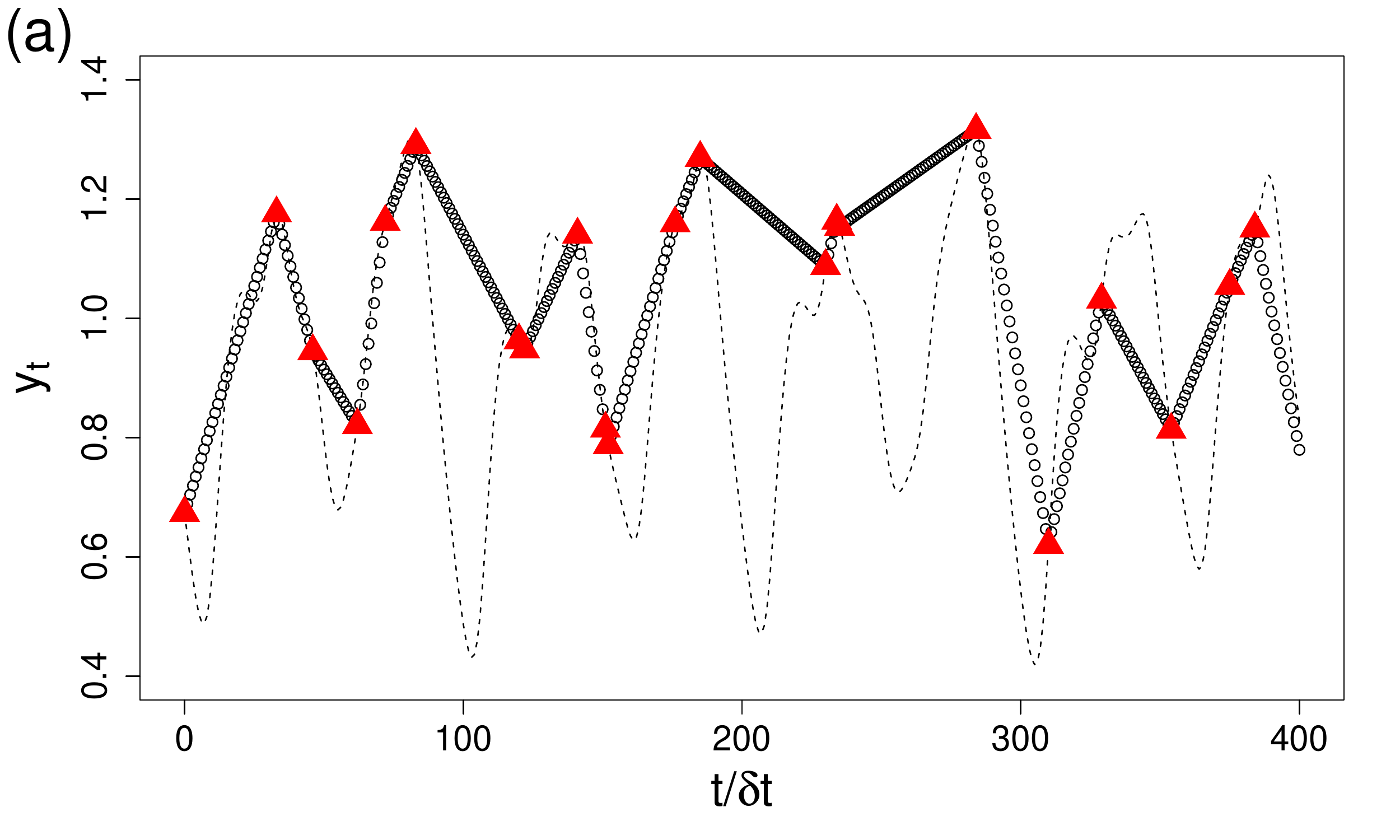}
  \includegraphics[width=0.48\textwidth]{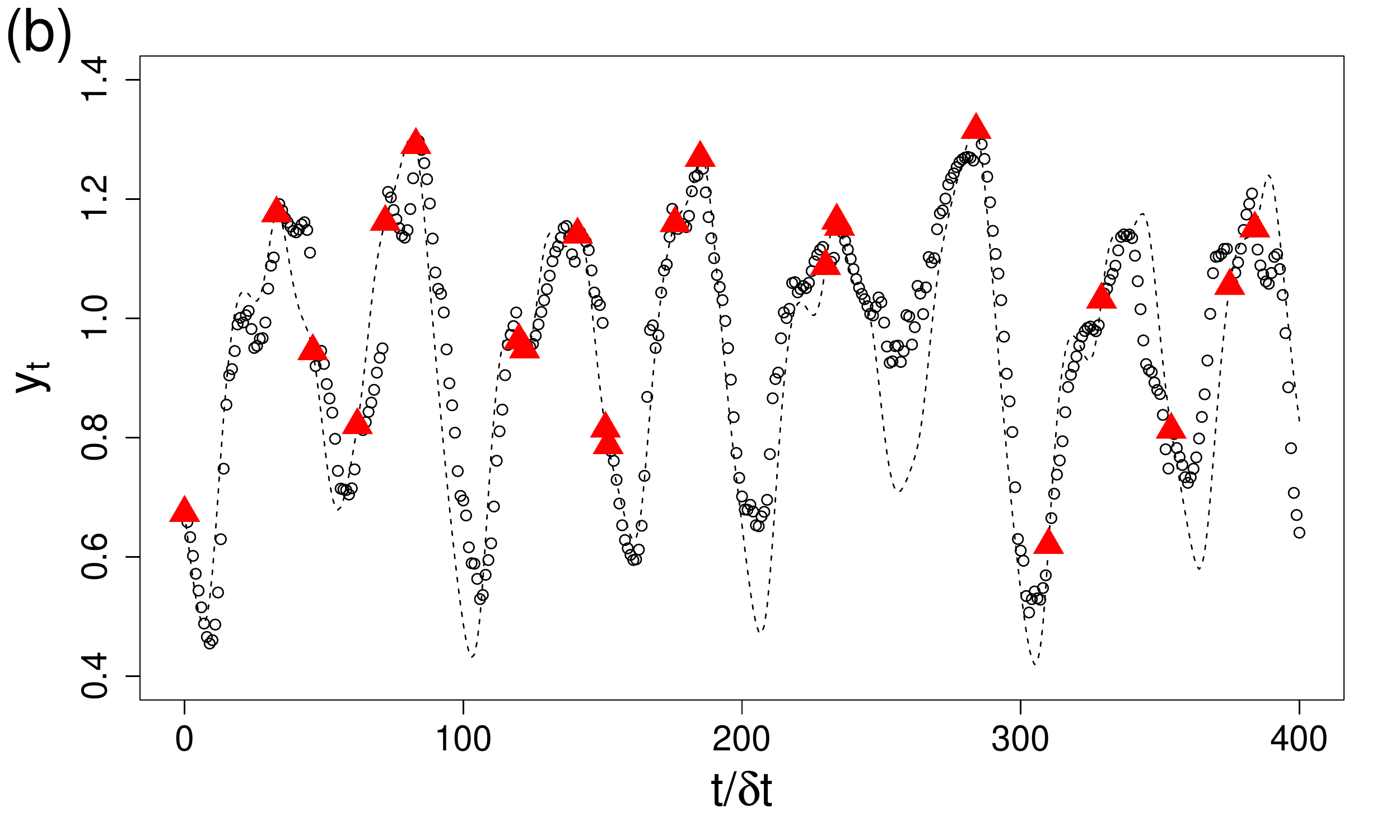}\\
  \includegraphics[width=0.48\textwidth]{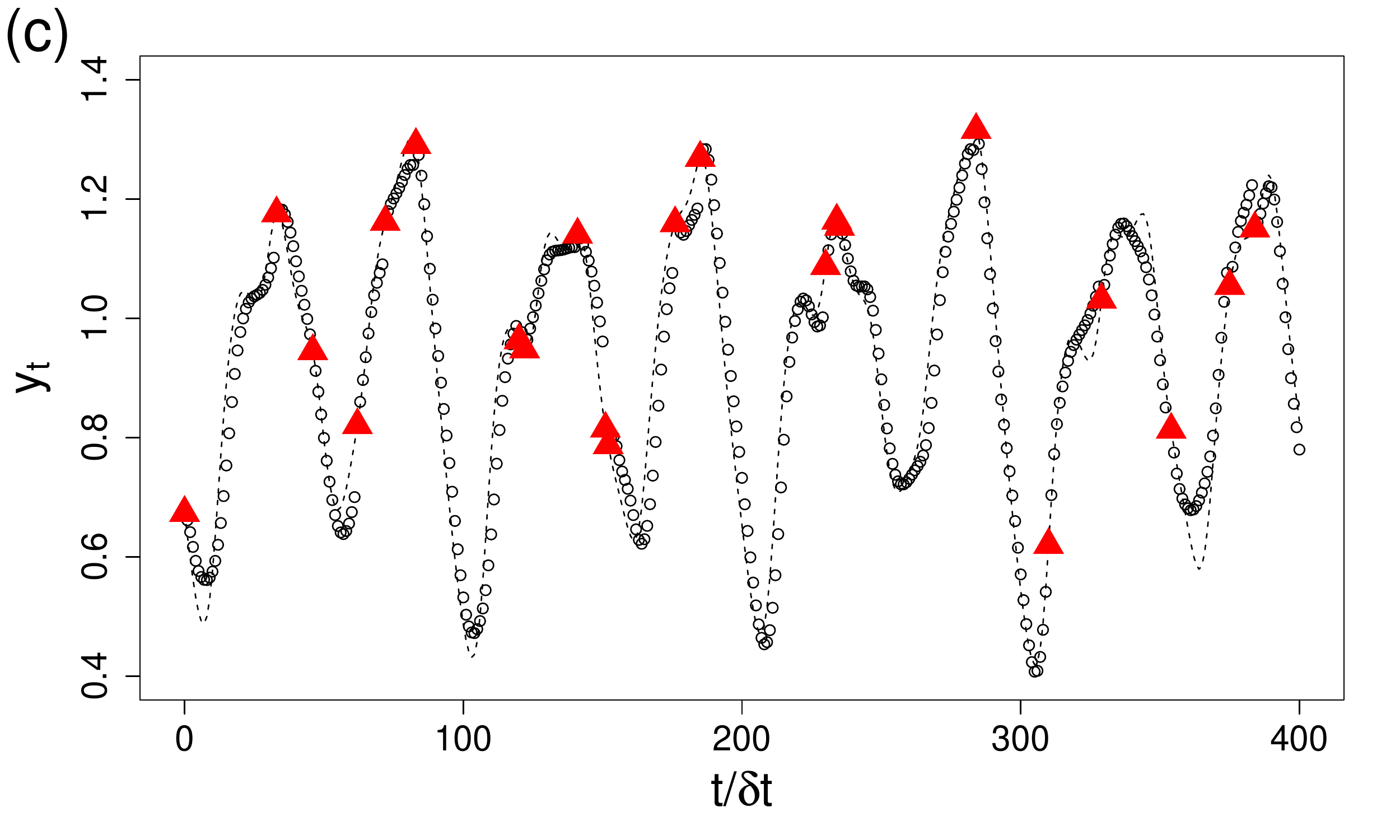}
  \includegraphics[width=0.48\textwidth]{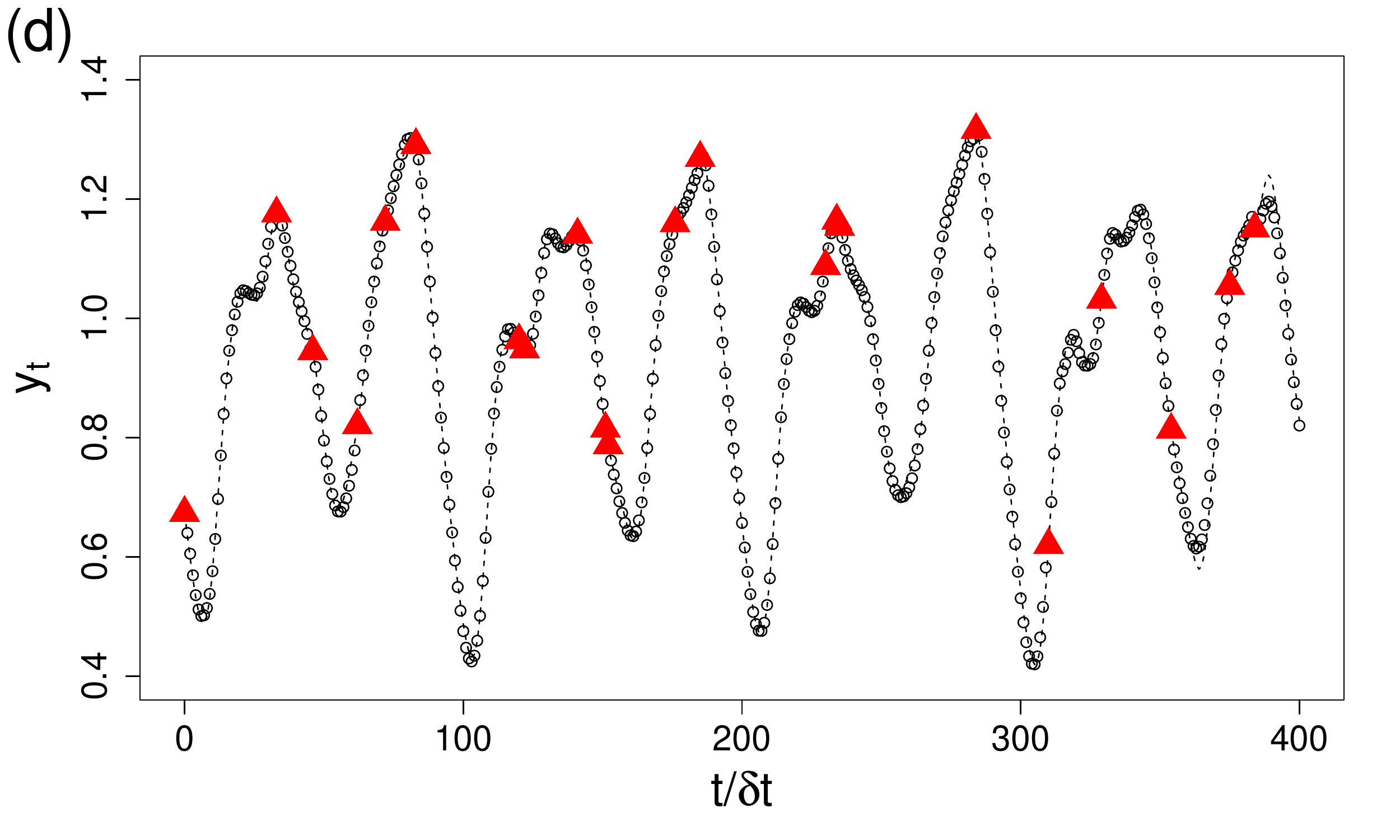}\\
  \caption{Reconstruction of the sparse Mackey-Glass time series ($\omega = 0.95$). The dashed line is the ground truth ($\bm{Y}^*$) and the sparse observation ($\bm{Y}^O$) is the solid triangles ({\color{red}$\blacktriangle$}). The hollow circles ($\circ$) are the reconstructed dynamics, $\bm{Y}^{R,\,k}$, at (a) $k =0$, (b) 25, (c) 50, and (d) 200.} \label{fig:MG_iter_95}
\end{figure}

Figure \ref{fig:MG_iter_95} shows the improvement of $\bm{Y}^{R,\,k}$ over the iterations for $\omega=0.95$. The initial condition, $\bm{Y}^{R,\,0}$, is given as a linear interpolation (fig. \ref{fig:MG_iter_95} a). It is clearly shown that, over the iterations, the fixed-point ESN learns the complex nonlinear, time-delay dynamics closer and closer to the ground truth. By the 200-th iteration, the reconstructed trajectory, $\bm{Y}^{R,200}$, becomes almost indistinguishable from $\bm{Y}^*$.

\begin{figure}
  \centering
  \includegraphics[width=0.4\textwidth]{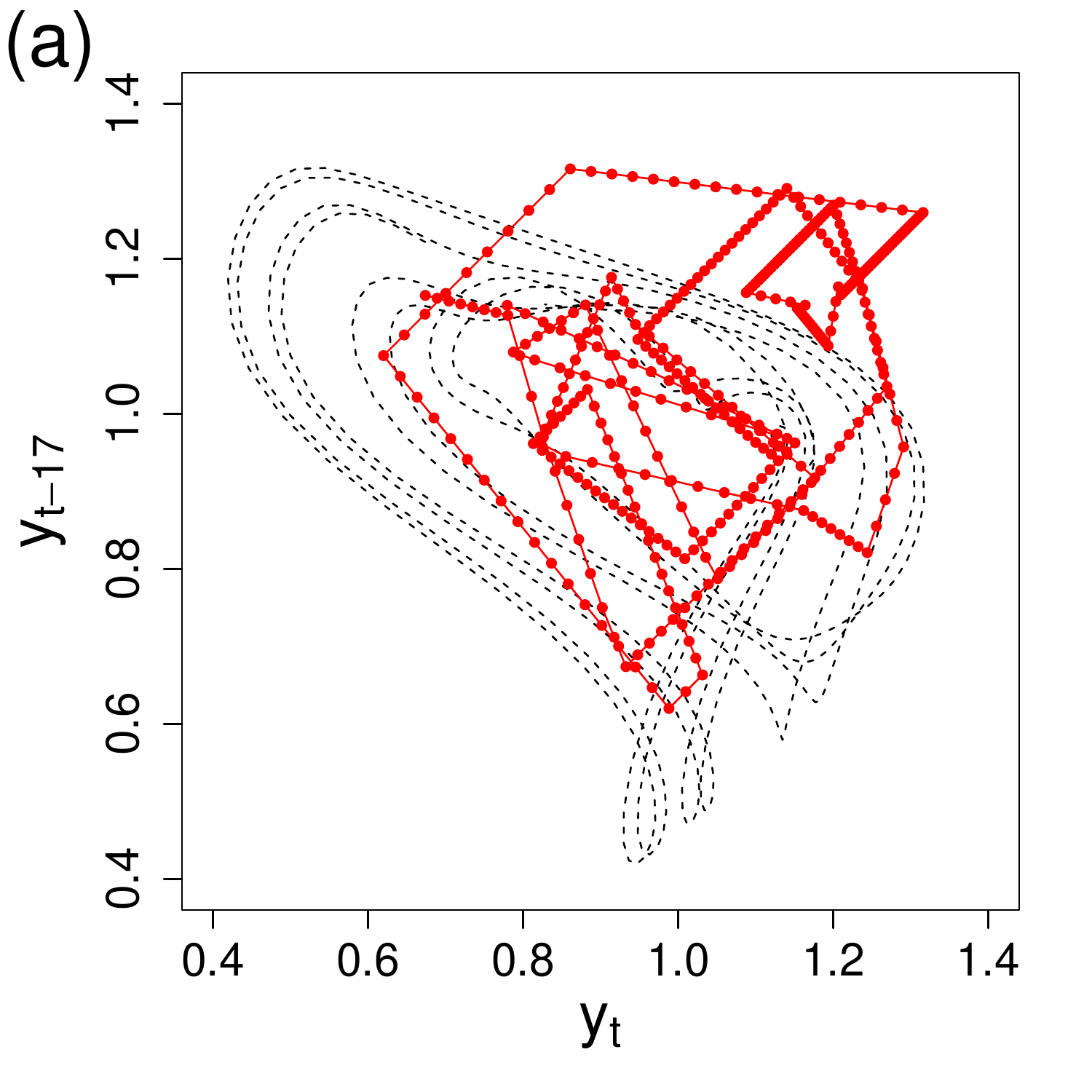}\hspace{1em}
  \includegraphics[width=0.4\textwidth]{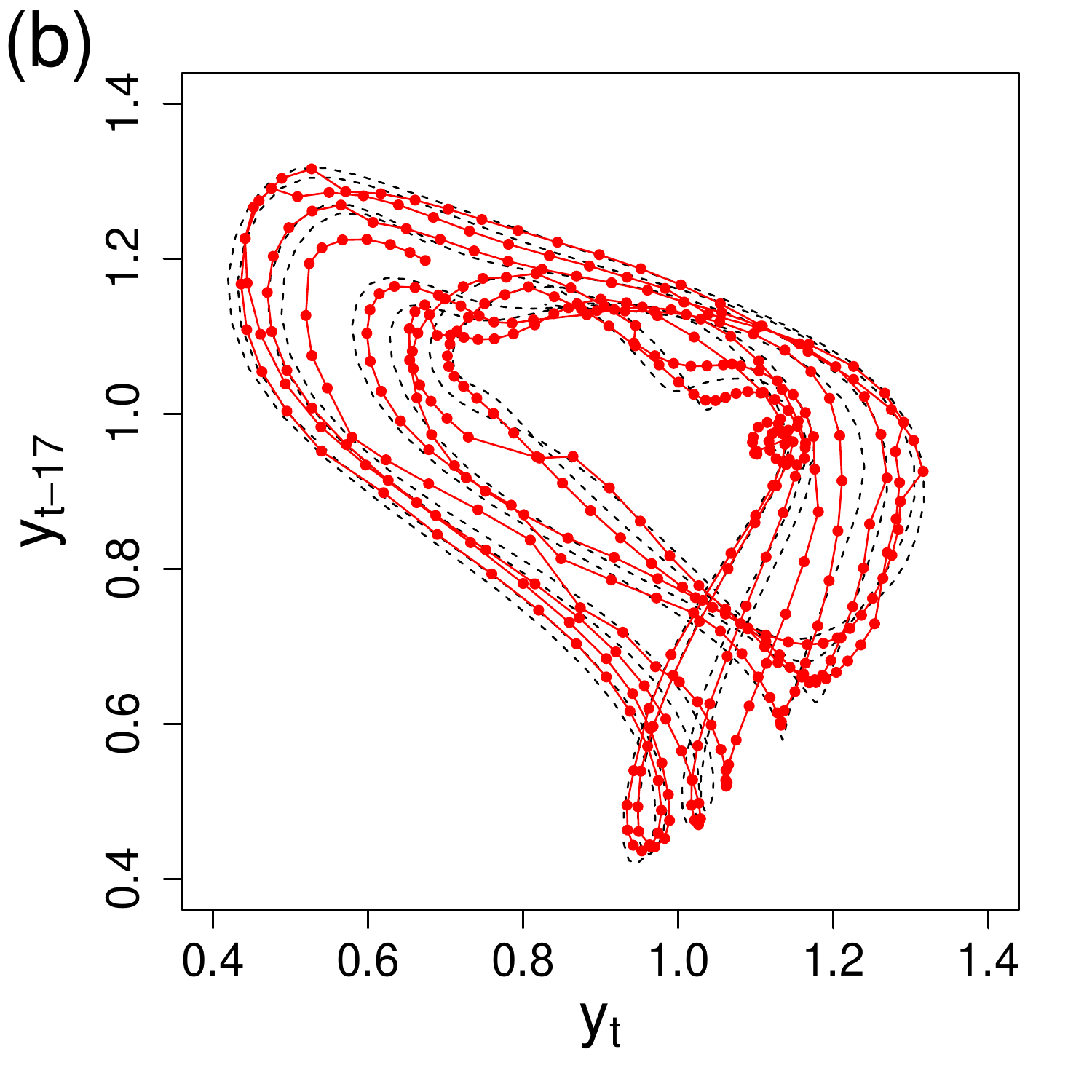}
  \caption{Reconstruction of the Mackey-Glass time series ($\omega = 0.95$) in the delay-coordinate phase space. The dashed line is the ground truth ($\bm{Y}^*$) and the solid circles ({\color{red}$\bullet$}) denote (a) the linear interpolation and (b) $\bm{Y}^{R}$ at $k = 200$.} \label{fig:MG_phase_95}
\end{figure}

Figure \ref{fig:MG_phase_95} shows the reconstructed chaotic attractor of the Mackey-Glass time series. Note that only the linearly interpolated time series (fig. \ref{fig:MG_phase_95} a) together with the locations of missing data is provided to the fixed-point ESN. After 200 iterations, the fixed-point ESN is able to closely reconstruct the chaotic attractor. This result suggests that the capability of ESN is beyond a high-dimensional interpolation in the phase space.

\begin{figure}
  \centering
  \includegraphics[width=0.8\textwidth]{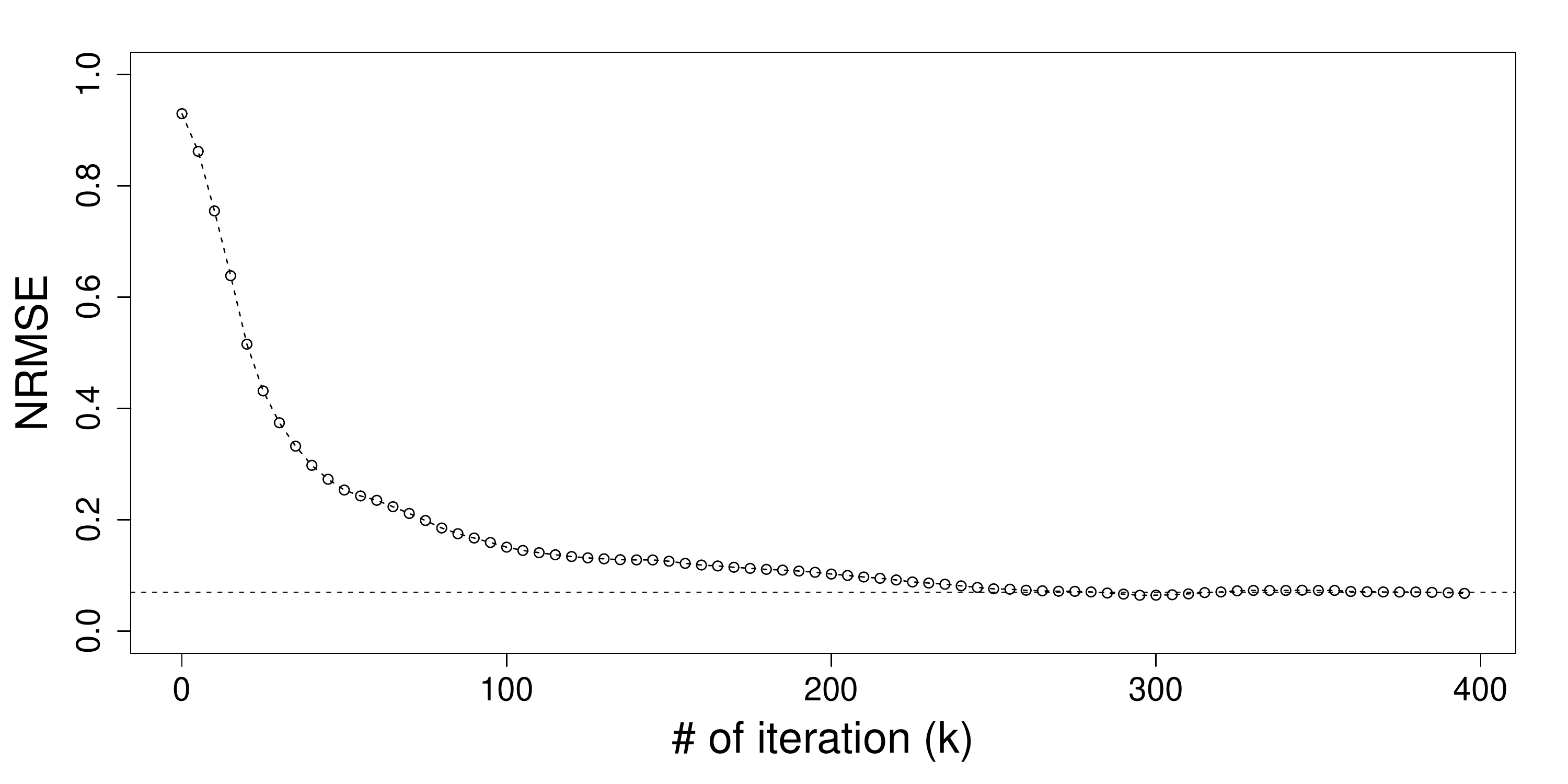}
  \caption{Root mean-square error (RMSE) of $\bm{Y}^{R,\,k}$ with respect to the ground truth, $\bm{Y}^*$, as a function of the iteration count. RMSE is normalized by the standard deviation of $\bm{Y}^*$.  The dashed line is the final NRMSE $\simeq 0.07$.} \label{fig:MG_improv_rmse}
\end{figure}

In figure \ref{fig:MG_improv_rmse}, a normalized root mean-square error of $\bm{Y}^{R}$ is shown as a function of the iteration number. The normalized root mean-square error (NRMSE) is defined as,
\begin{equation}
\text{NRMSE} = \left[  \frac{\sum_i (y_i^{R,\,k}-y^*_i)^2}{\sum_i {y^*_i}^2 } \right]^{1/2}.
\end{equation}
There is a rapid decrease in NRMSE in the first 100 iterations. NRMSE is decreased from 0.93 of the initial condition to 0.15 at $k=100$. After $k=280$, $\bm{Y}^R$ seems to be converged and there is no further improvement in NRMSE.

\begin{figure}
  \centering
  \includegraphics[width=0.48\textwidth]{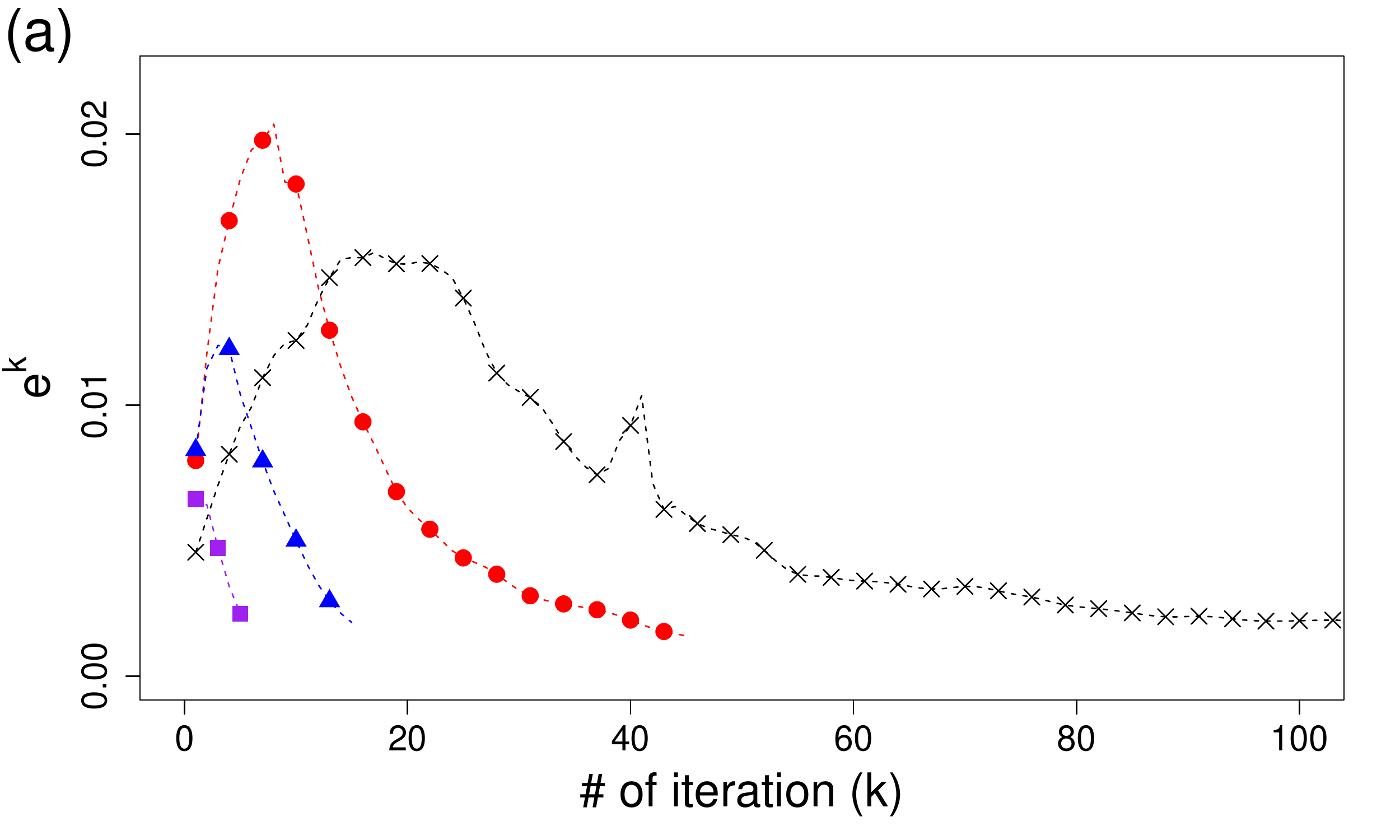}
  \includegraphics[width=0.48\textwidth]{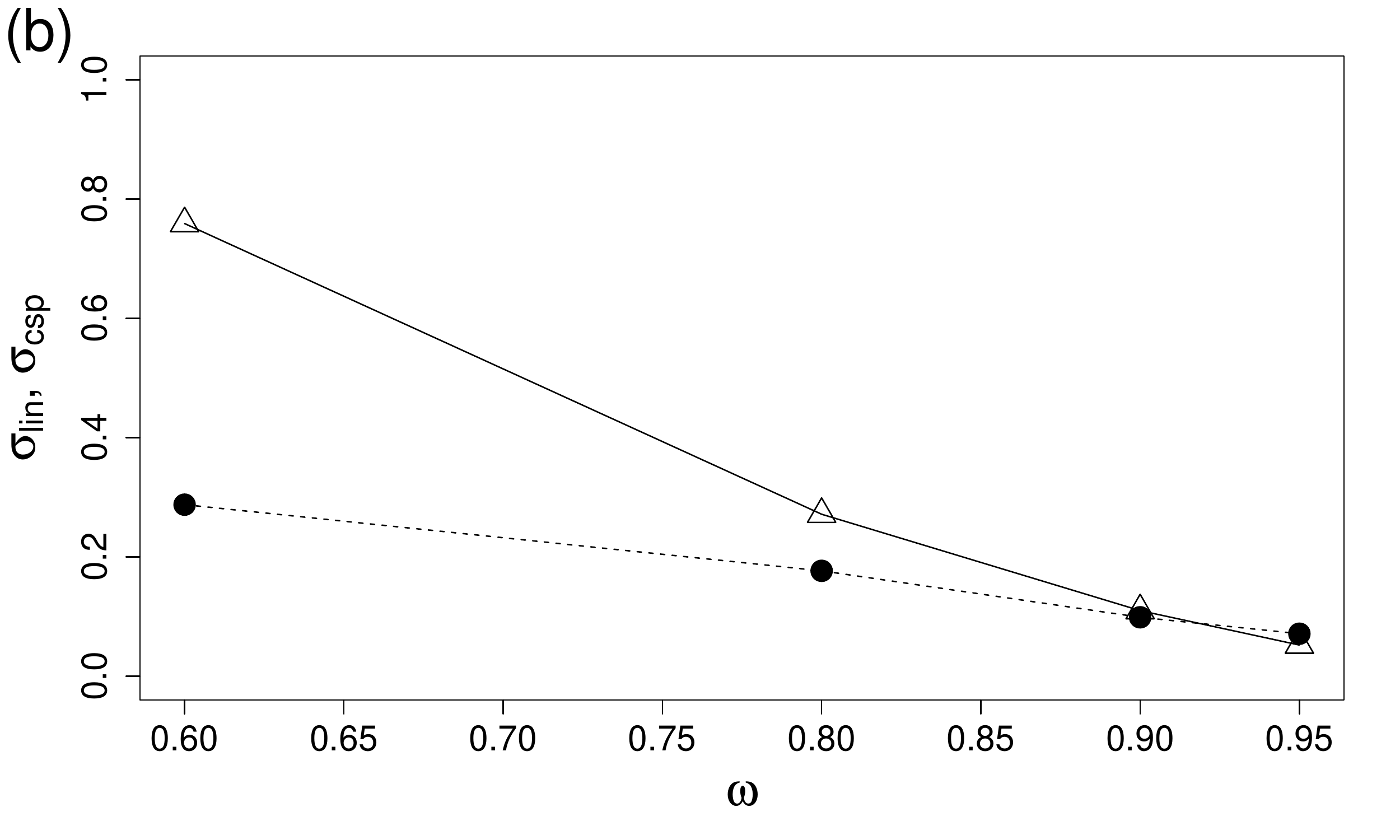}
  \caption{(a) Changes in the $l_2$-improvement with respect to the number of iterations for $\omega = 0.60$ ($\color{magenta}\blacksquare$), 0.80 ($\color{blue}\blacktriangle$), 0.90 ($\color{red}\bullet$), and 0.95 ($\times$). (b) RMSE of $\bm{Y}^R$ with respect to $\bm{Y}^*$ normalized by the RMSEs of linear ($\bullet$) and cubic spline interpolations ($\vartriangle$). } \label{fig:MG_converge}
\end{figure}

\begin{table}
\center{
\caption{Relaxation ($\alpha$) and regularization ($\beta$) parameters of the fixed-point ESN} 
\label{tbl:MG_param}
\begin{tabular}{c|cccc}
\hline \hline
$\omega$ & 0.60 & 0.80 & 0.90 & 0.95\\
\hline
$\alpha$ & 0.2 & 0.2 & 0.2 & 0.4 \\
$\beta$ & $10^{-9}$ & $10^{-9}$ & $10^{-9}$ & $10^{-8}$\\
\hline \hline
\end{tabular}
}
\end{table}

In figure \ref{fig:MG_converge}, the behavior of the fixed-point ESN is investigated for a range of the missing rate, $0.6 \le \omega \le 0.95$. The relaxation and regularization parameters are listed in table \ref{tbl:MG_param}. It is found that, when $\omega$ is large, both $\alpha$ and $\beta$ need to be increased to make the fixed-point iteration stable. The data sets have the equal length, $T=5\times10^4$. The number of observations changes from $2\times10^4$ at $\omega = 0.6$ to $2.5\times10^3$ at $\omega = 0.95$.

Figure \ref{fig:MG_converge} (a) shows the convergence of the fixed-point ESN for the range of $\omega$. During the fixed point iteration, the convergence is checked by the $l_2$-improvement,
\begin{equation}
e^k = \left[ \frac{1}{T-T_s}\sum_{j=T_s+1}^T (y^{R,\,k+1}_j - y^{R,\,k}_j)^2 \right]^{1/2}.  
\end{equation}
It is shown that, when $\omega$ is modest, the fixed point iteration converges very rapidly. For $\omega = 0.6$, i.e., 40\% of the data is available, the fixed-point ESN converges under 5 iterations, which increases to 50 iterations for $\omega = 0.90$. 

In figure \ref{fig:MG_converge} (b), the accuracy of the fixed-point ESN is compared with linear and cubic spline interpolations. The normalized RMSE is computed as,
\begin{equation}
\sigma_{lin} = \left[  \frac{\sum_i (y_i^{R}-y^*_i)^2}{\sum_i (y_i^{lin}-y^*_i)^2 } \right]^{1/2},\text{and}~~\sigma_{csp} = \left[  \frac{\sum_i (y_i^{R}-y^*_i)^2}{\sum_i (y_i^{csp}-y^*_i)^2 } \right]^{1/2}.
\end{equation}
Here, $y^{lin}$ and $y^{csp}$, respectively, denote the linear and cubic spline interpolations. The fixed-point ESN outperforms both the linear and cubic spline interpolations even at low missing rate, $\omega = 0.6$. The root mean-square error of the fixed-point ESN is about 30\% of the linear interpolation and 80\% of the cubic spline interpolation. The strength of the fixed-point ESN is more pronounced at higher $\omega$. When $\omega = 0.95$, RMSE of the fixed-point ESN is less than 8\% of the linear and cubic spline interpolations.

\begin{figure}
  \centering
  \includegraphics[width=0.48\textwidth]{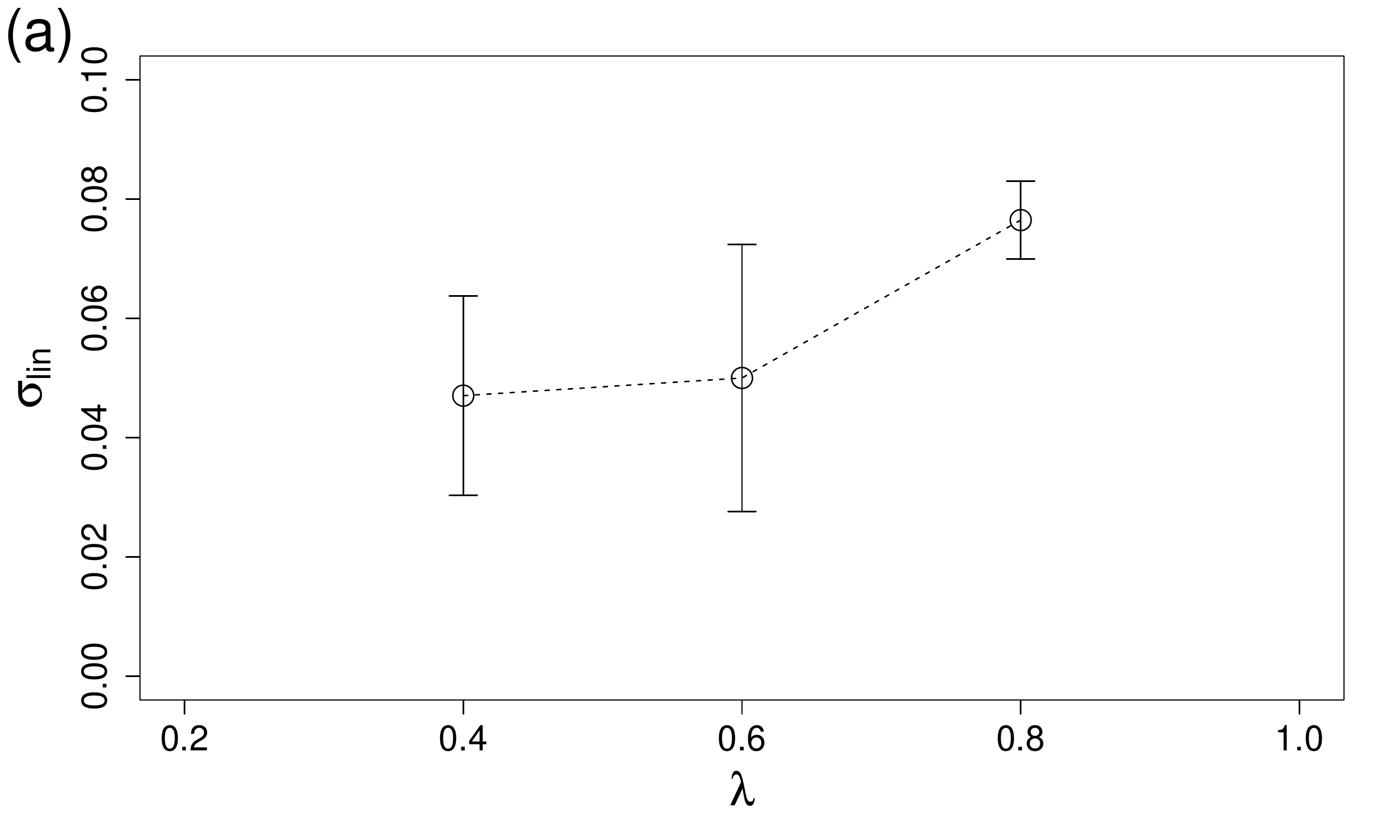}
  \includegraphics[width=0.48\textwidth]{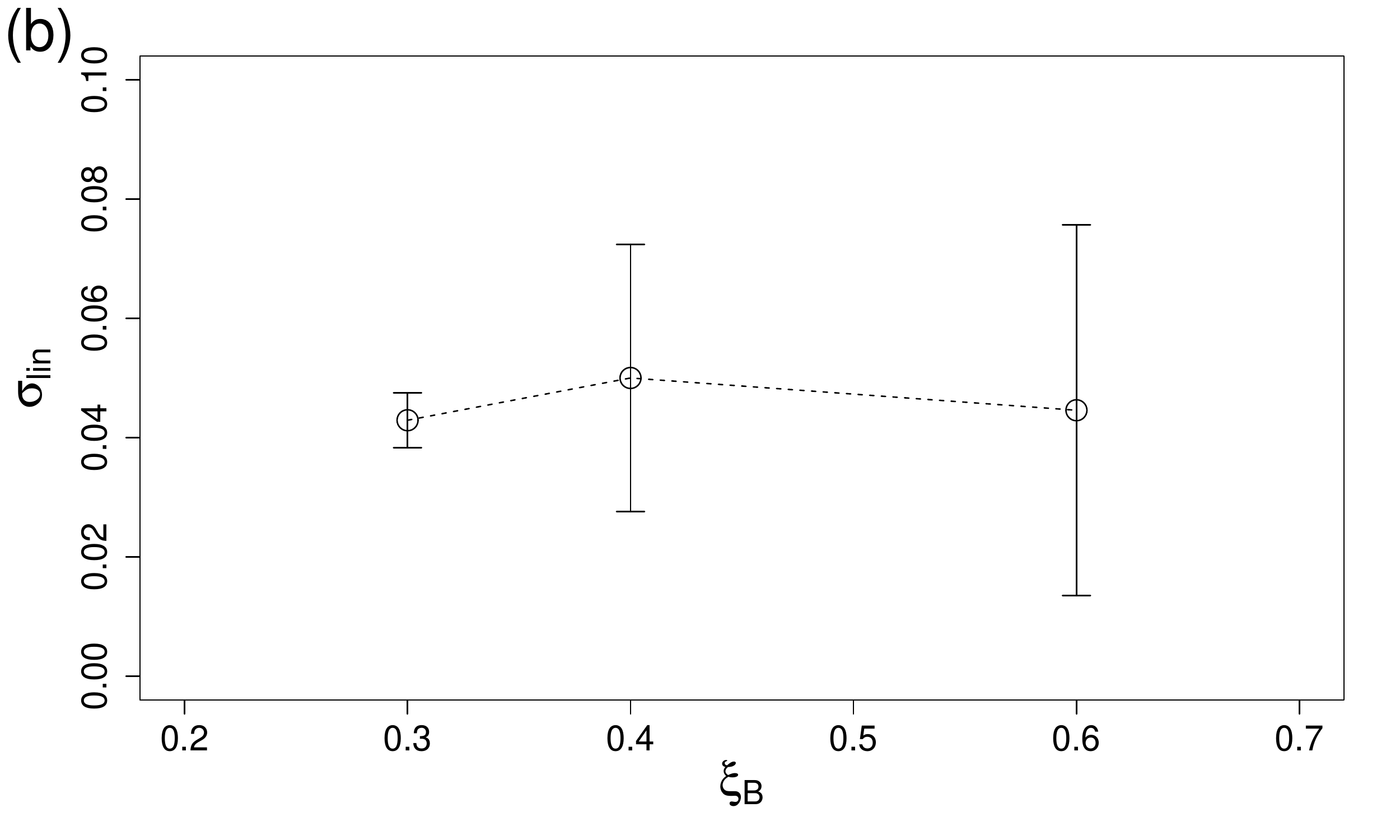}\\
  \includegraphics[width=0.48\textwidth]{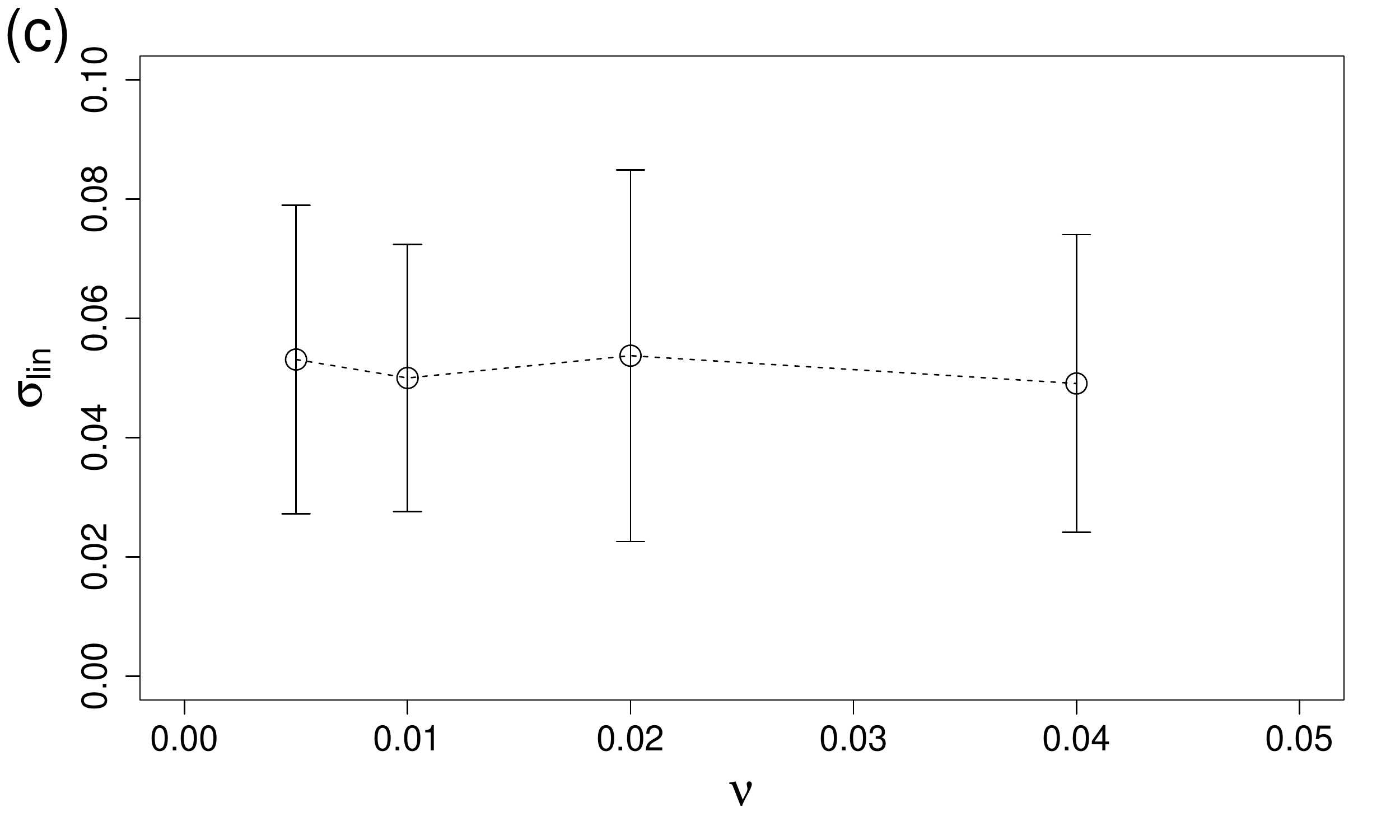}
  \includegraphics[width=0.48\textwidth]{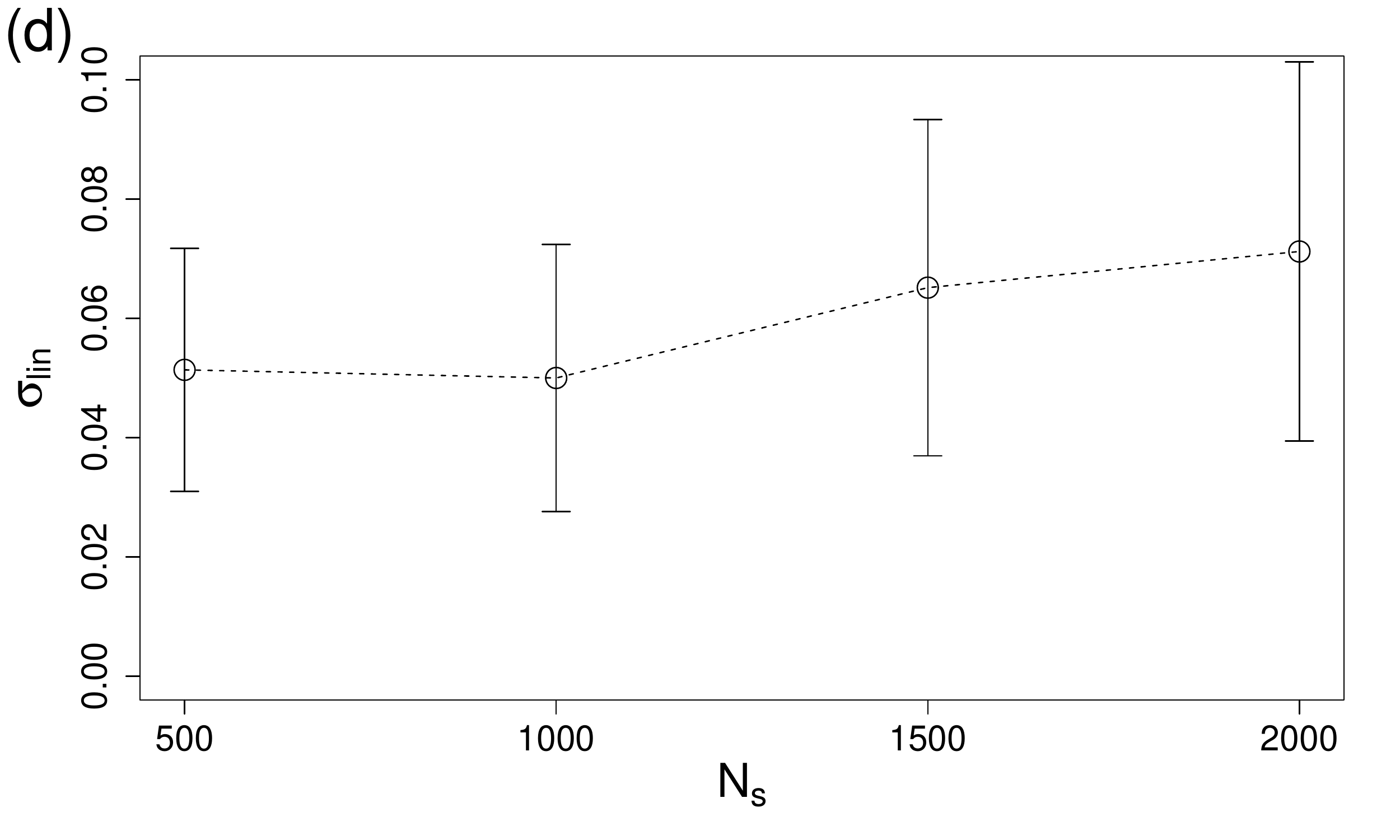}
  \caption{Sensitivity of the fixed-point ESN reconstruction with respect to the free parameters for $\omega = 0.9$; (a) temporal relaxation ($\lambda$), (b) scale of the weight matrix $\bm{B}$ ($\xi_B$), (c) fraction of non-zero elements ($\nu$), and (d) size of ESN ($N_s$). The hollow symbol and error bar denote the expectation and one standard deviation of $\sigma_{lin}$ from an ensemble of 20 ESNs. } \label{fig:MG_param}
\end{figure}

The effects of the free parameters and the random generation of ESN are shown in figure \ref{fig:MG_param}. In the experiments, four free parameters considered, $\lambda$, $\xi_B$, $\nu$, and $N_s$. In each experiment, all of the free parameters are fixed at the reference values (\ref{eqn:ref_param}) except for the one being tested. The normalized RMSEs with respect to the linear interpolation ($\sigma_{lin}$) are computed from 20 randomly generated fixed-point ESNs. It is shown that the expectation of $\sigma_{lin}$, for the reference parameter set, e.g., $\lambda = 0.6$ in figure \ref{fig:MG_param} (a), is much lower than $\sigma_{lin}$ from a single realization in figure \ref{fig:MG_converge} (b); $E[\sigma_{lin}] \simeq 0.05$ versus $\sigma_{lin} \simeq 0.10$. Note that, for the range of parameters tested, the variations of $E[\sigma_{lin}]$ is less than 0.04, indicating that the fixed-point ESN is not so sensitive to the changes in the free parameters.

The effects of the temporal relaxation parameter, $\lambda$, are shown in figure \ref{fig:MG_param} (a). In general, the error decreases at smaller $\lambda$ for the Mackey-Glass problem. However, it is found that, when $\lambda \le 0.2$, the fixed-point ESN does not converge. The temporal relaxation parameter, $0 \le \lambda < 1$, works effectively as a regularization. Increasing $\lambda$ enforces a longer autocorrelation, or higher inertia, which results in a smoother trajectory. On the other hand, $\lambda = 0$ does not impose such constraint in the temporal dynamics. When the missing fraction, $\omega$, is high, a regularization in the temporal structure is required to guarantee temporal smoothness in the solution, which explains why the fixed-point ESN does not converge when $\lambda \le 0.2$.

Figure \ref{fig:MG_param} (b) shows the changes in $\sigma_{lin}$ as a function of the scale of the weight matrix, $B_{ij} \sim \mathcal{U}(-\xi_B,\xi_B)$. It is shown that, while $E[\sigma_{lin}]$ is not so sensitive to $\xi_B$, the standard deviation of $\sigma_{lin}$ is an increasing function of $\xi_B$. The scale of $\bm{B}$ determines the relative contribution between the input variable, $\bm{x}_t$, and the latent state, $\bm{s}_t$, to the time evolution of the latent state, $\bm{s}_{t+1}$. Because the time evolution of $\bm{s}$ becomes more autonomous when $\xi_B$ is small, the standard deviation of $\sigma_{lin}$ also becomes smaller \cite{Yeo19b}.

It is argued that making the connection of the latent state sparse leads to a richer dynamics \cite{Jaeger04}. In figure \ref{fig:MG_param} (c), $\sigma_{lin}$ is computed as a function of the sparsity, $\nu$. It is shown that, in the range of $\nu$ explored, the fixed-point ESN is not sensitive to the changes in $\nu$.

In figure \ref{fig:MG_param} (d), $E[\sigma_{lin}]$ is shown to be an increasing function of $N_s$ when $N_s > 1000$. In general, it is expected that increasing $N_s$ makes the nonlinear representation capability of an ESN stronger. However, when the number of data is fixed, using a large $N_s$ makes it difficult to train an ESN because the number of unknown parameters increases linearly proportional to $N_s$. For $\omega = 0.9$, the number of observations is $(1-\omega)T = 5\times10^3$, while the number of unknown parameters is $\text{dim}(\bm{\theta}) = N_s + 1$. The increase in the reconstruction error for $N_s > 1000$ seems to be related with the decrease in the ratio between the number of unknown parameters to the number of data.

\subsection{Lorenz-63 System}\label{sec:Lorenz_63}
In this section, we use the Lorenz-63 system \cite{Lorenz63} to validate the fixed-point ESN. The Lorenz-63 system is given by the following set of equations,
\begin{equation}  \label{eqn:Lorenz}
\frac{d}{dt}
\begin{bmatrix}
x\\y\\z
\end{bmatrix}
=
\begin{bmatrix}
\gamma_1(y-x)\\
x(\gamma_2 - z) - y\\
xy - \gamma_3 z
\end{bmatrix}.
\end{equation}
We used the coefficients from \cite{Lorenz63}, $\gamma_1 = 10$, $\gamma_2 = 8/3$, and $\gamma_3 = 28$. The time series data is generated by sampling at every,  $\delta t = 0.02$. The length of the time series is $T = 5\times10^4$.  

\begin{figure}
  \centering
  \includegraphics[width=0.48\textwidth]{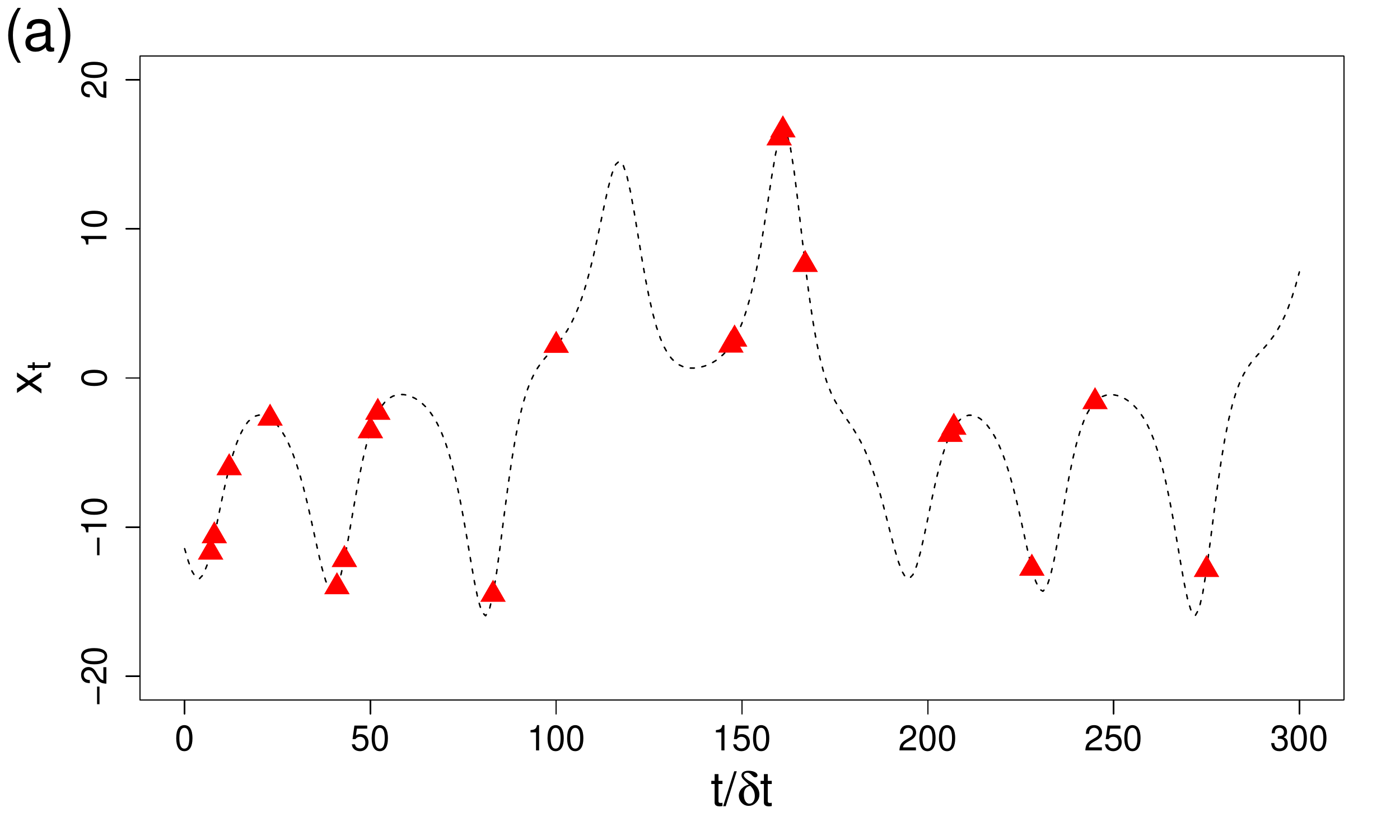}
  \includegraphics[width=0.48\textwidth]{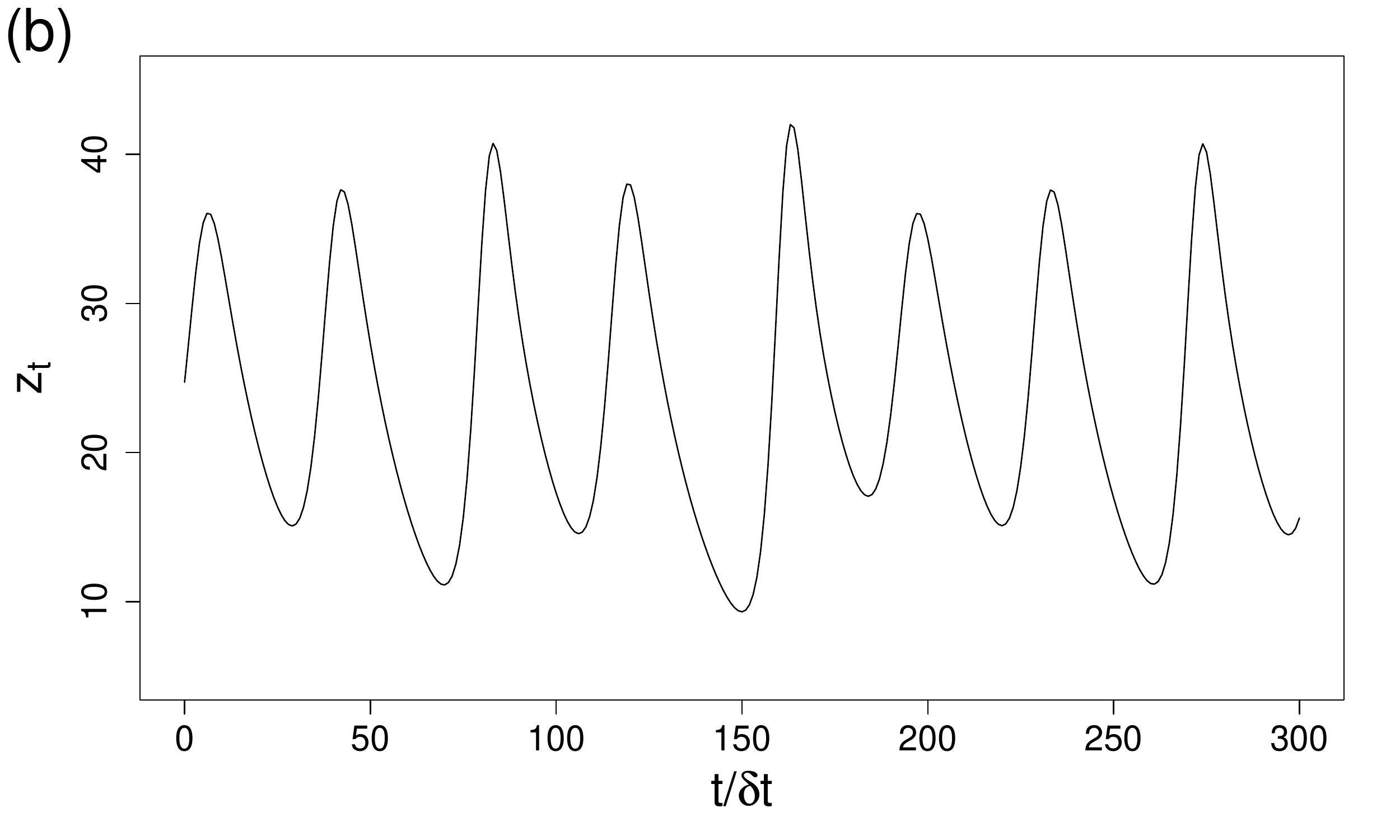}
  \caption{Sparse, partial observation of the Lorenz 63 system. The time series data consists of only (a) $x_t$ and (b) $z_t$, and 95\% of $x_t$ is missing. In (a), the dashed line is the ground truth and the solid symbols (${\color{red}\blacktriangle}$) denote the sparse observation.} \label{fig:Lorenz63_data_X}
\end{figure}

First, we consider a partial observation of the Lorenz-63 system \cite{Lu17}. The data set consists of only ($x_t,z_t$). We assume that $z_t$ is completely known, while $\omega$ fraction of the data is randomly missing in $x_t$ . Figure \ref{fig:Lorenz63_data_X} shows an example of the partial observation with missing data. Here, we aim to recover the missing data in $x_t$ from ($x^O_t,z_t$). 

\begin{figure}
  \centering
  \includegraphics[width=0.48\textwidth]{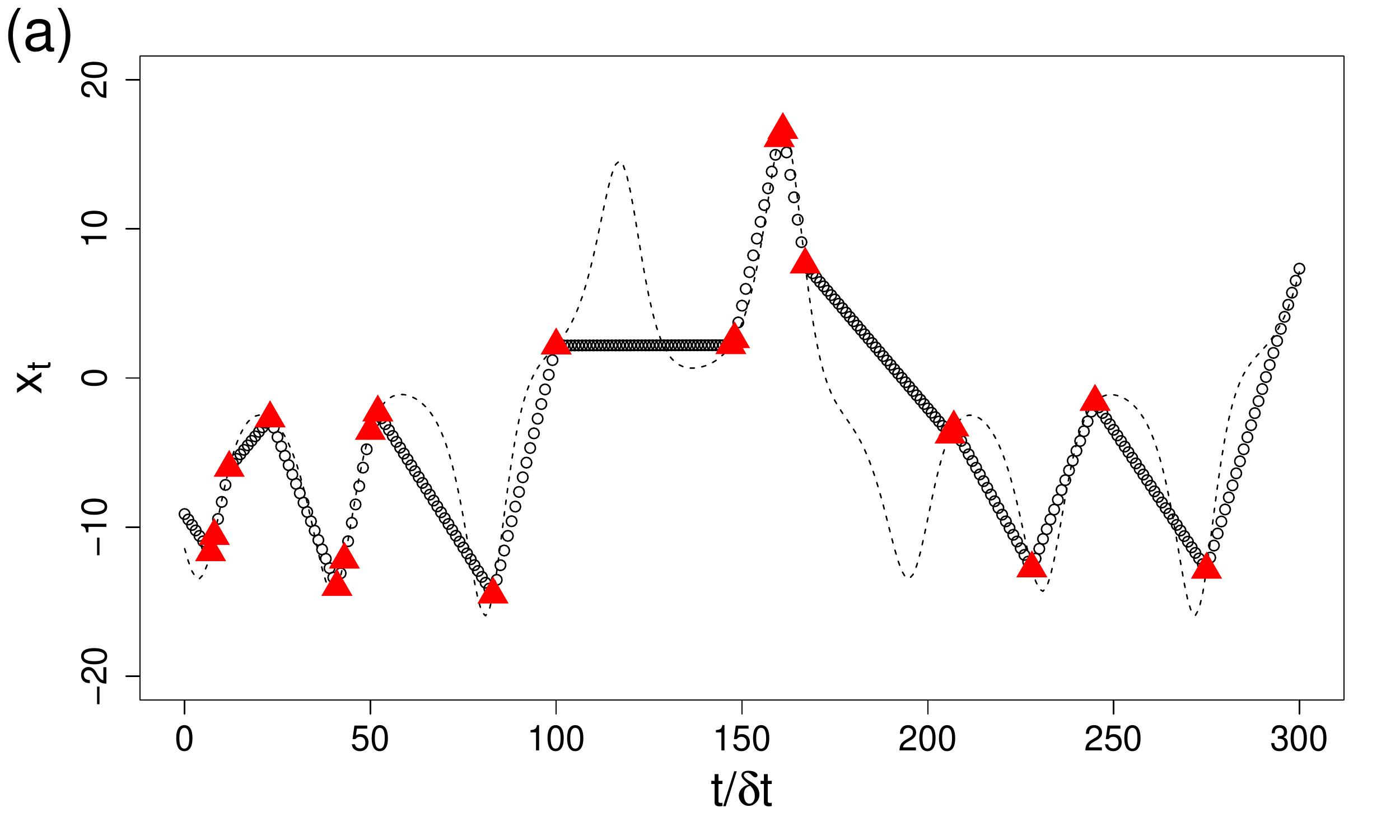}
  \includegraphics[width=0.48\textwidth]{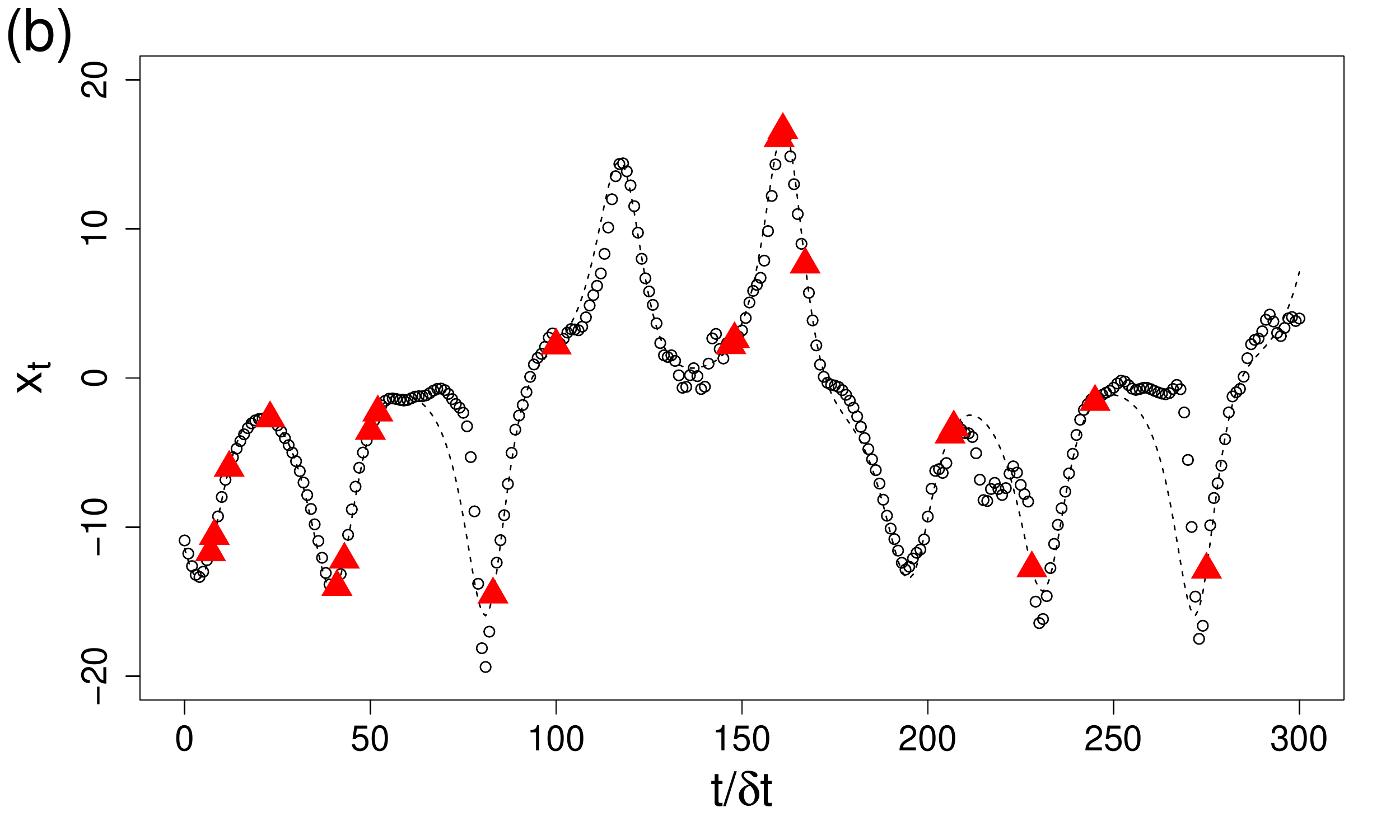}\\
  \includegraphics[width=0.48\textwidth]{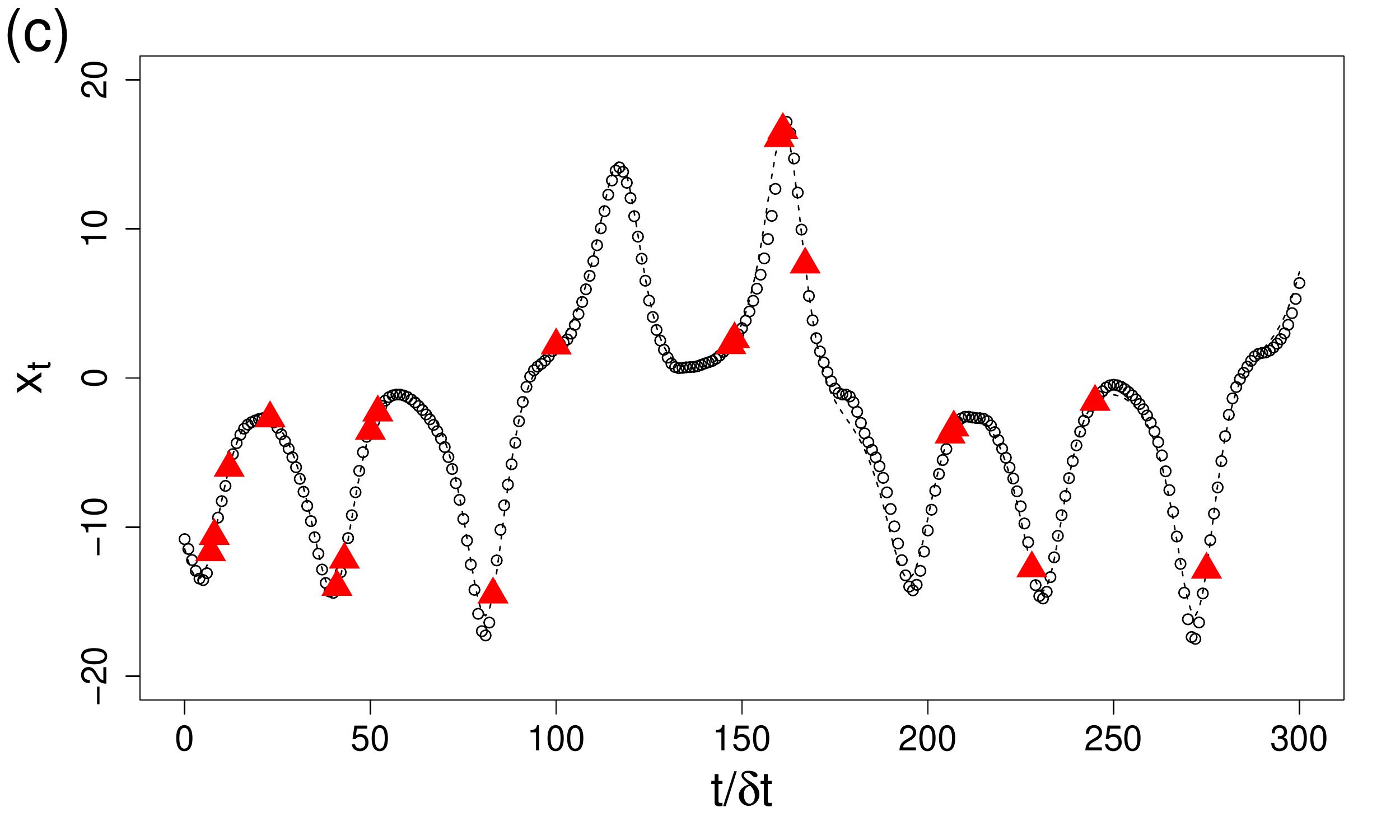}
  \includegraphics[width=0.48\textwidth]{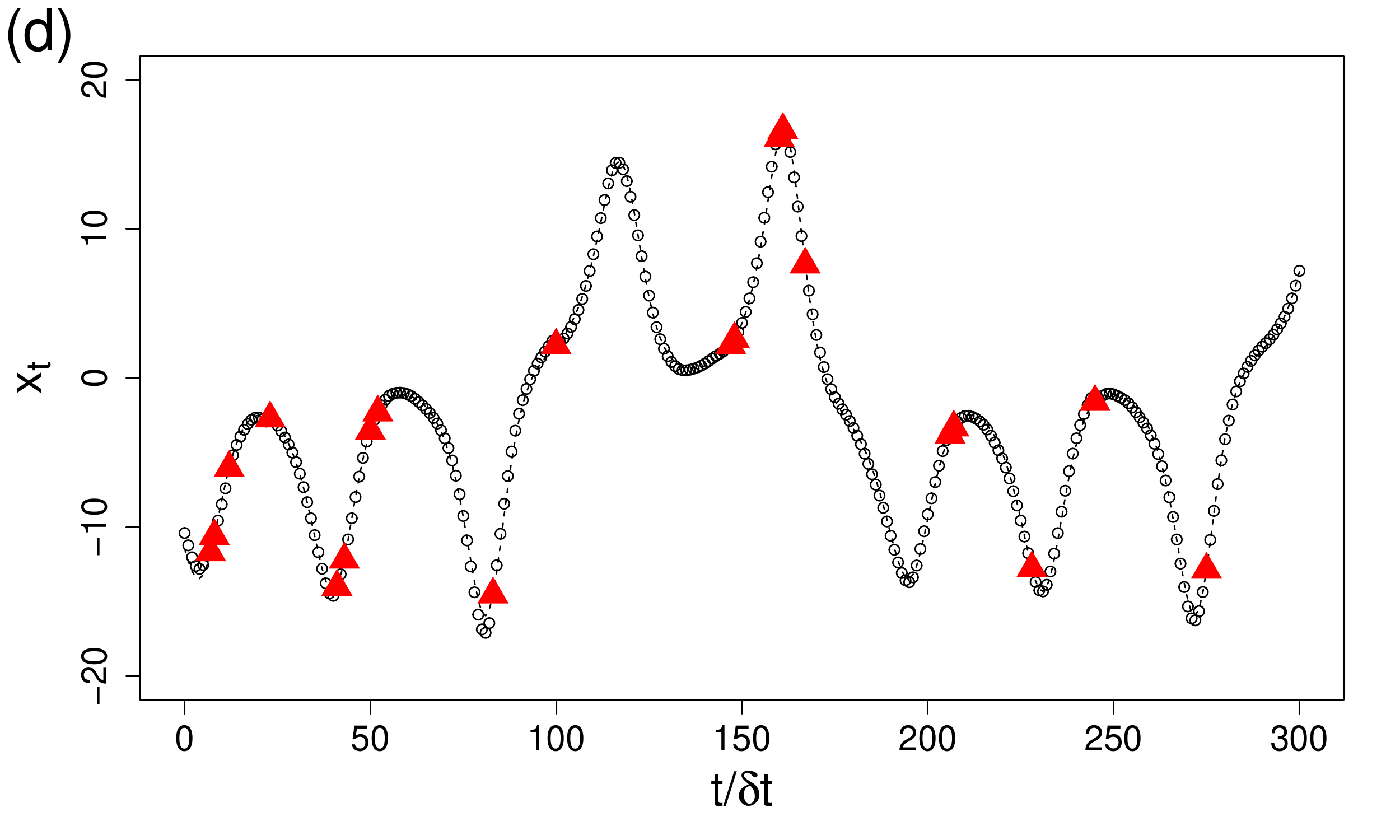}\\
  \caption{Reconstruction from the sparse, partial observation of Lorenz-63 time series. $z_t$ is fully observed, while 95\% of $x_t$ is missing. The dashed line is the ground truth ($\bm{Y}^*$) and the sparse observation ($\bm{Y}^O$) is the solid triangles ({\color{red}$\blacktriangle$}). The hollow circles ($\circ$) are the reconstructed trajectory, $\bm{Y}^{R,\,k}$, at (a) $k =0$, (b) 50, (c) 100, and (d) 300.} \label{fig:Lorenz63_iter_X}
\end{figure}

Figure \ref{fig:Lorenz63_iter_X} shows the reconstruction, $\bm{Y}^R$, at four different iteration counts. The fixed-point ESN starts from a linear interpolation (Fig. \ref{fig:Lorenz63_iter_X} a). Similar to the Mackey-Glass time series (Fig. \ref{fig:MG_improv_rmse}), the reconstruction error quickly reduces in the first 100 iterations (Fig. \ref{fig:Lorenz63_iter_X} b-c).  By $k=100$, $\bm{Y}^R$ is already very close to the ground truth, $\bm{Y}^*$. At $k = 300$, $\bm{Y}^R$ becomes almost indistinguishable from $\bm{Y}^*$. The RMSE normalized by the linear interpolation is $\sigma_{lin} = 0.07$ for the time window shown in figure \ref{fig:Lorenz63_iter_X} (d).

\begin{figure}
  \centering
  \includegraphics[width=0.96\textwidth]{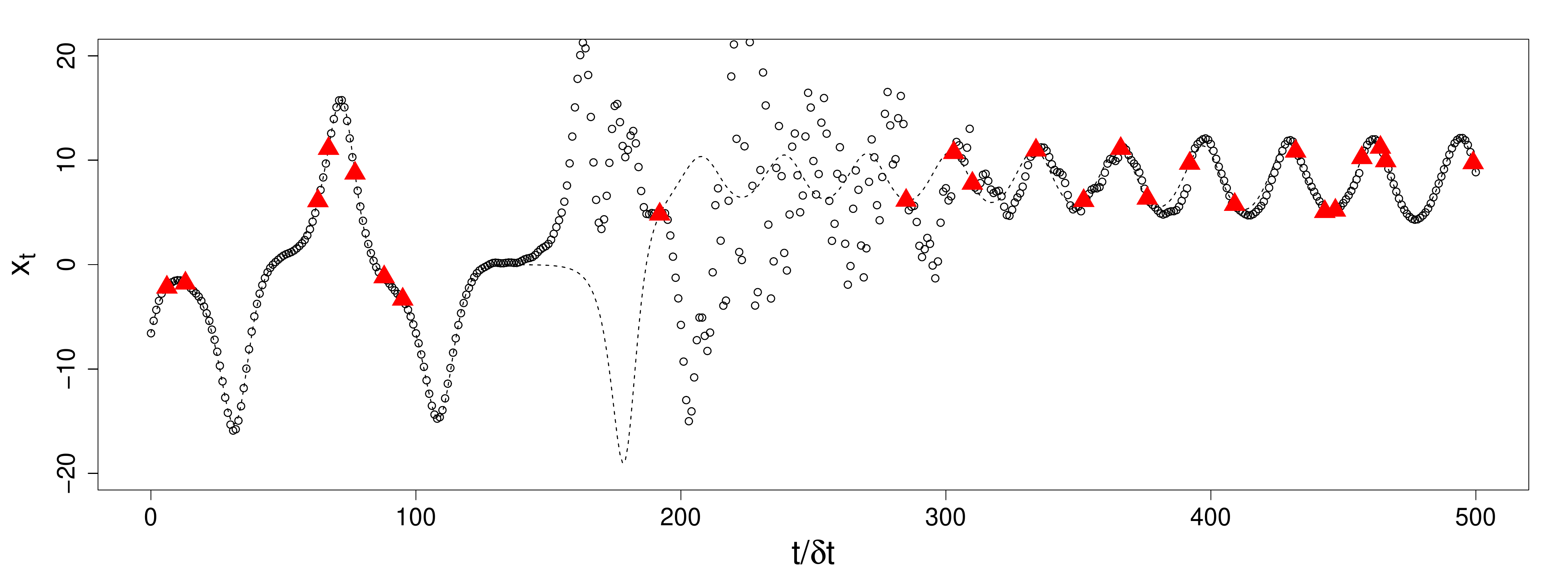}
  \caption{Section of the largest error in the reconstructed Lorenz-63 time series from the 95\% missing data. The dashed line is the ground truth ($\bm{Y}^*$), the sparse observation ($\bm{Y}^O$) is the solid triangles ({\color{red}$\blacktriangle$}), and the hollow circles ($\circ$) detnoe the reconstructed dynamics.}\label{fig:Lorenz63_reconst_X}
\end{figure}

\begin{table}
\center{
\caption{Relaxation ($\alpha$) and regularization ($\beta$) parameters and NRMSE for the partial observation of the Lorenz-63 time series.} 
\label{tbl:partial_Lorenz63_param}
\begin{tabular}{c|ccc}
\hline \hline
$\omega$ & $\alpha$ & $\beta$ & $\sigma_{lin}$\\
\hline
0.90 & 0.2 & $10^{-7}$ & 0.06  \\
0.95 & 0.4 & $10^{-6}$ & 0.37 \\
\hline \hline
\end{tabular}
}
\end{table}

The normalized RMSEs ($\sigma_{lin}$) for $\omega = 0.9$ and 0.95 are shown in table \ref{tbl:partial_Lorenz63_param}, together with the parameters ($\alpha$, $\beta$). $\sigma_{lin}$ is computed over the entire trajectory, $T = 5\times10^4$. It is shown that, when $\omega=0.9$, RMSE of the fixed-point ESN is only about 6\% of the linear interpolation, which changes to 37\% at $\omega = 0.95$. Figure \ref{fig:Lorenz63_reconst_X} explains why there is a significant increase in RMSE at $\omega = 0.95$. Because of the random sampling, in the time window shown in figure \ref{fig:Lorenz63_reconst_X}, there is only one observation at $t = 193\delta t$ for $t \in [96\delta t, 285\delta t]$. In other words, two extremely large missing intervals ($97\delta t$ and $92 \delta t$) exist consecutively. This is a rare case, because the average interval of the missing observation is $20 \delta t$ for $\omega = 0.95$. As discussed in \cite{Lu17}, it is not possible to correctly predict $x(t)$ given only $z(t)$, because, when $y(t)$ is not observed, the symmetry of the Lorenz-63 system implies that $x(t)$ and $-x(t)$ are equally possible. Hence, the reconstructed trajectory, $\bm{Y}^R$, predicts a positive peak at $t/\delta t = 165$, while the ground truth, $\bm{Y}^*$, reaches a negative peak at $t/\delta t = 180$. Then, for $193 \le t/\delta t \le 285$, the fixed-point ESN cannot adjust its dynamics due to the lack of the data. On one hand, this observation demonstrates the limitation of the fixed-point ESN, which fails to capture the dynamics when the missing interval is too long. One the other hand, the Lorenz system is chaotic, of which long-term predictability is quite limited. 

\begin{figure}
  \centering
  \includegraphics[width=0.48\textwidth]{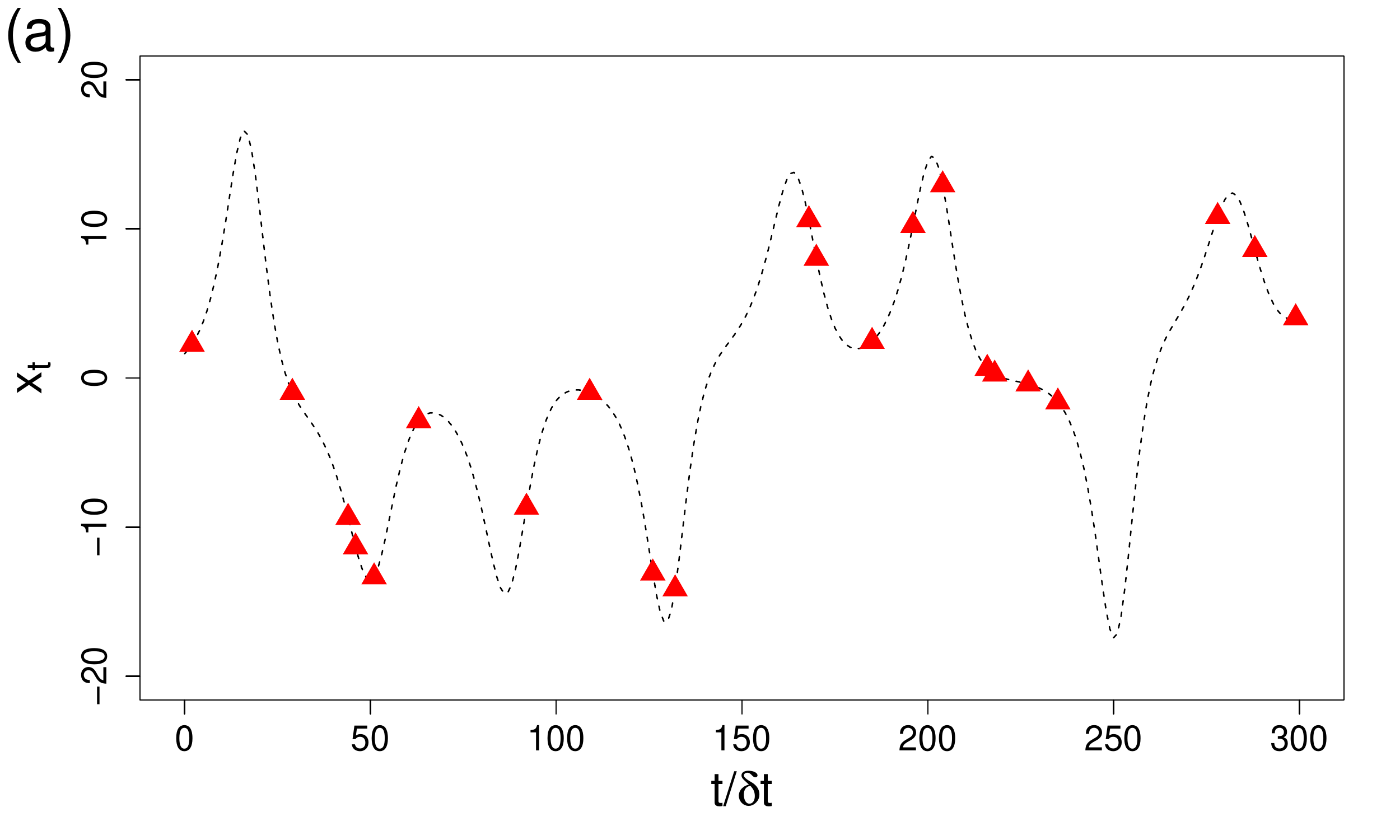}
  \includegraphics[width=0.48\textwidth]{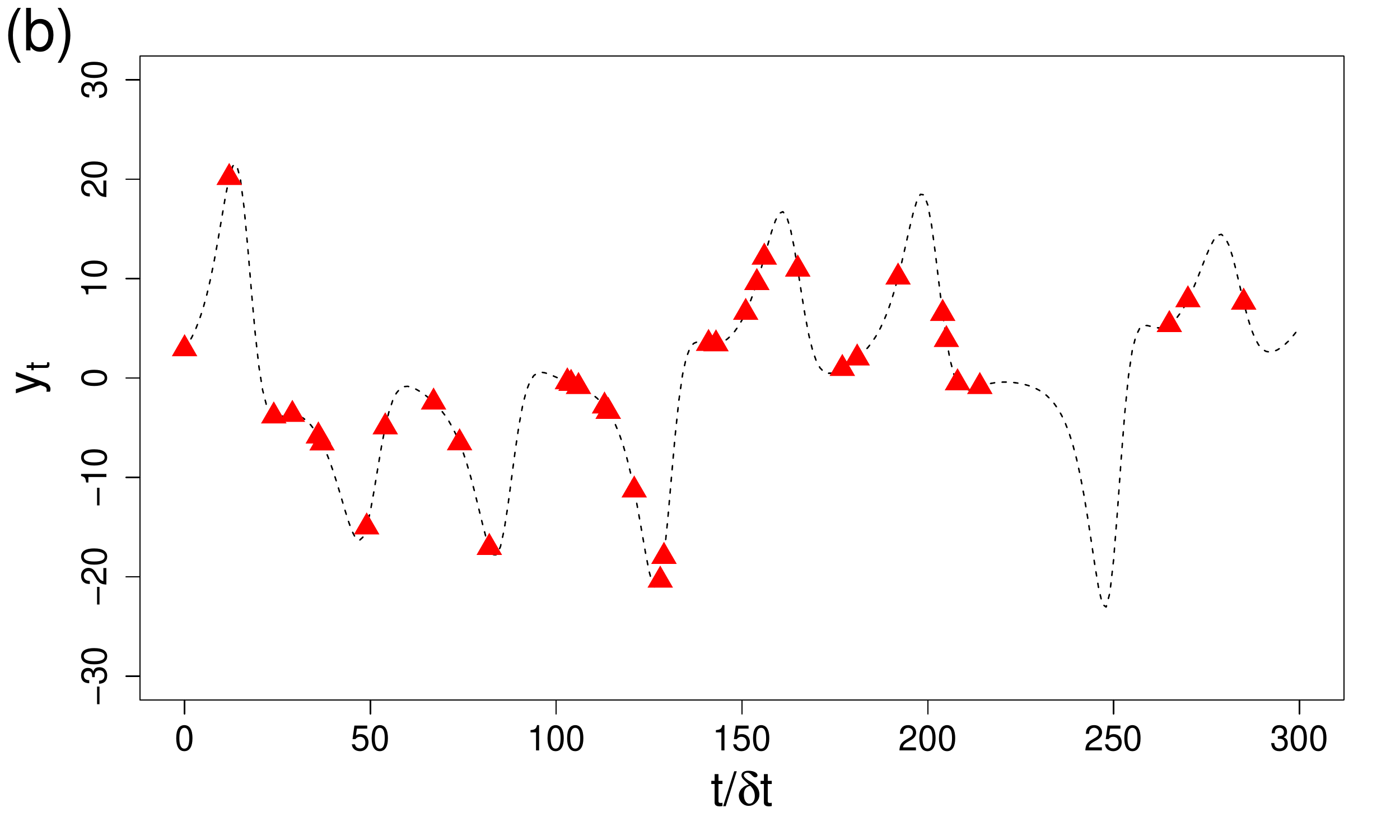}\\
  \includegraphics[width=0.48\textwidth]{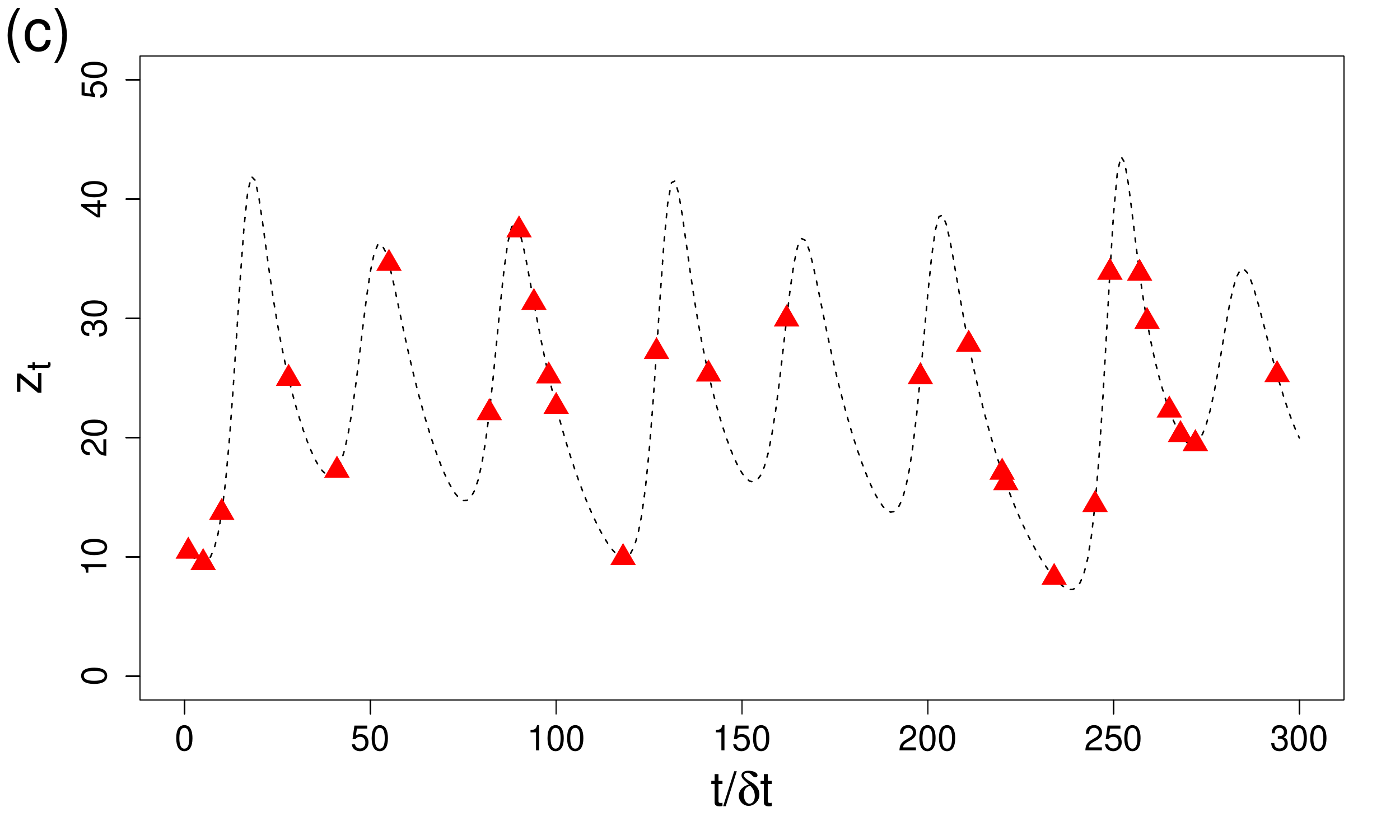}
  \includegraphics[width=0.48\textwidth]{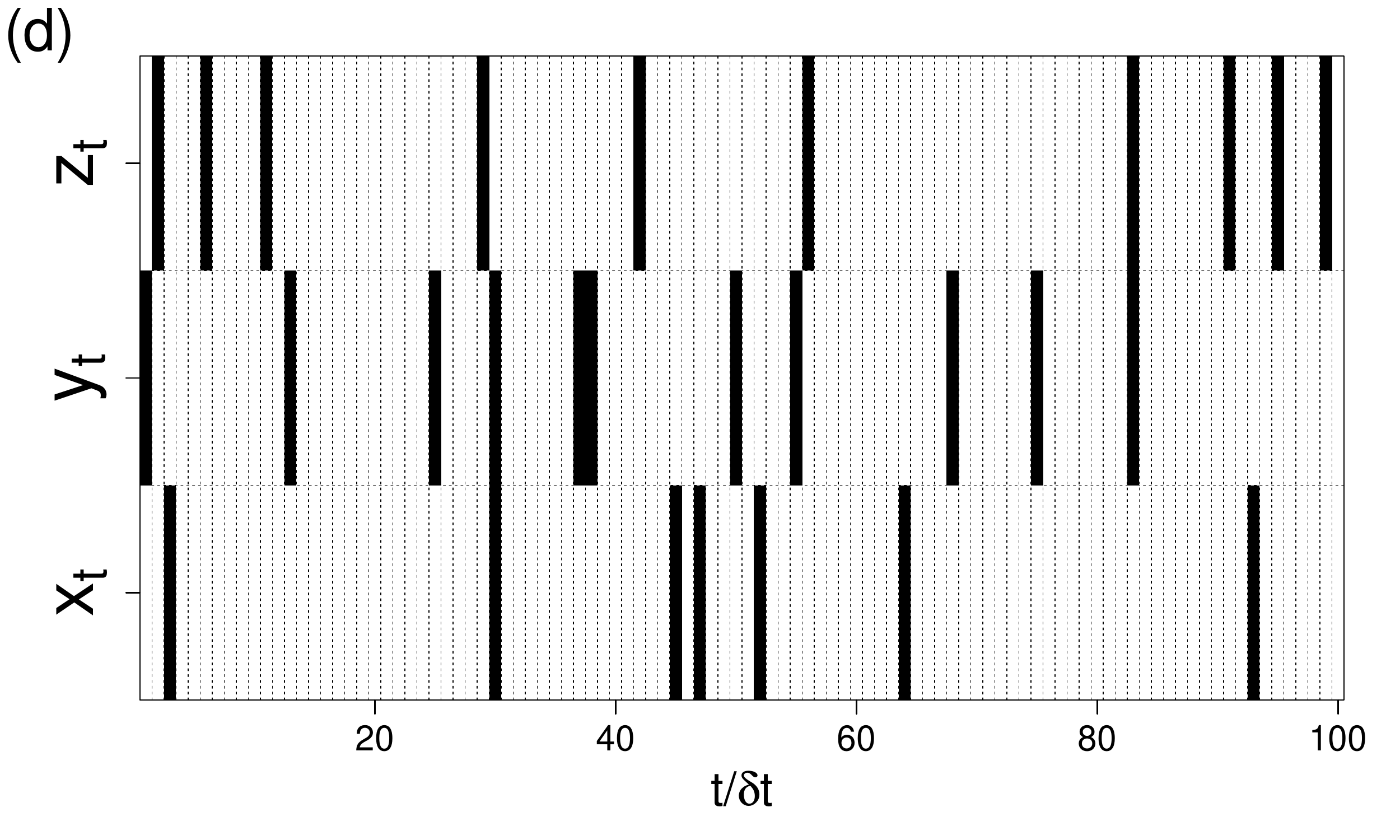}
  \caption{Sparse observation of the Lorenz-63 system with $\omega = 0.9$; (a) $x_t$, (b) $y_t$, and (c) $z_t$.  In (a--c), the dashed line is the ground truth and the solid symbols (${\color{red}\blacktriangle}$) denote the sparse observation. An example of the data availability is shown in (d). The location of the observation is marked by the black bars.} \label{fig:Lorenz63_data_all}
\end{figure}

In the next experiment, the sparse observation of all three variables, $(x_t,y_t,z_t)$, are given, and we aim to reconstruct the dynamics of the three-dimensional time series simultaneously. Figure \ref{fig:Lorenz63_data_all} (a--c) shows an example of the sparse observation of the Lorenz-63 time series. All three variables have the same missing rate, i.e., $\omega_x = \omega_y = \omega_z = \omega$. Figure \ref{fig:Lorenz63_data_all} (d) shows the availability of the data. Since the time series data is removed at random, the probability that all three variables are observed at the same time frame is 0.001 for $\omega = 0.9$. 

\begin{figure}
  \centering
  \includegraphics[width=0.7\textwidth]{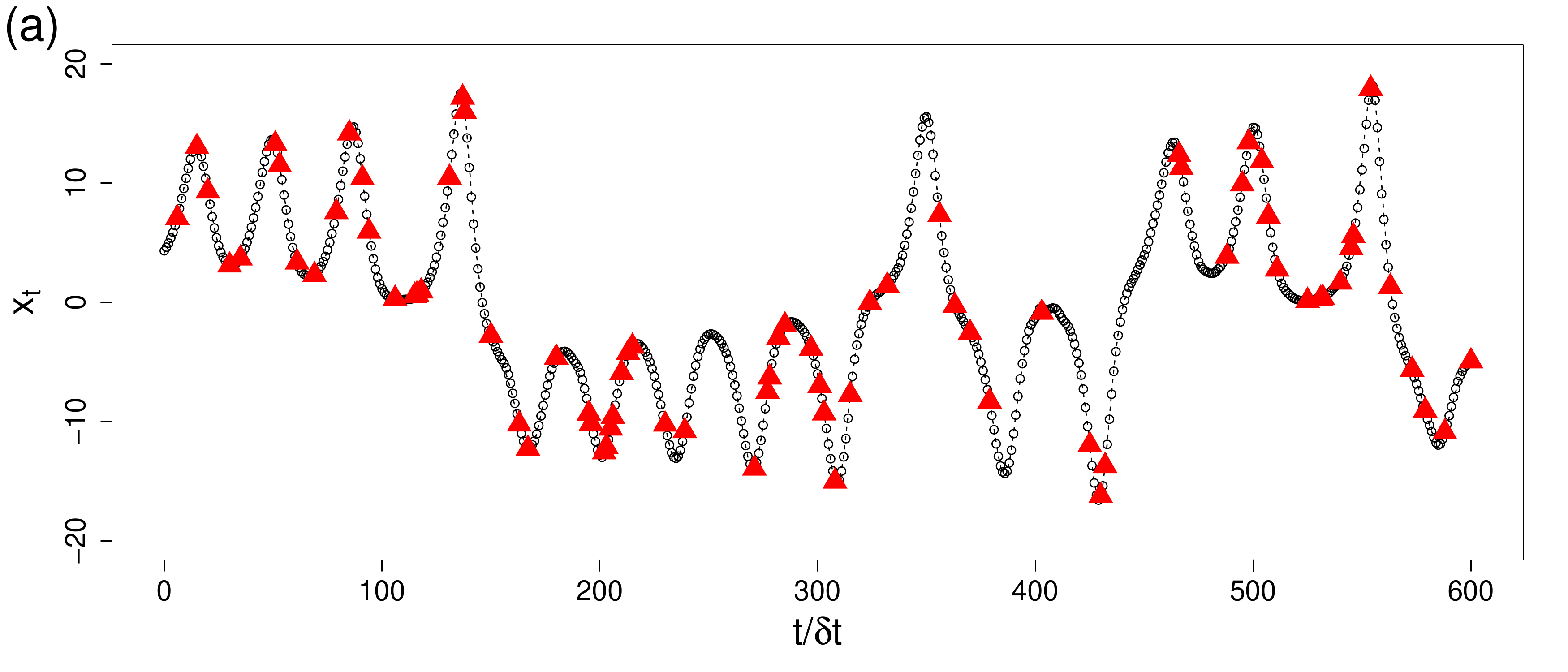}\\
  \includegraphics[width=0.7\textwidth]{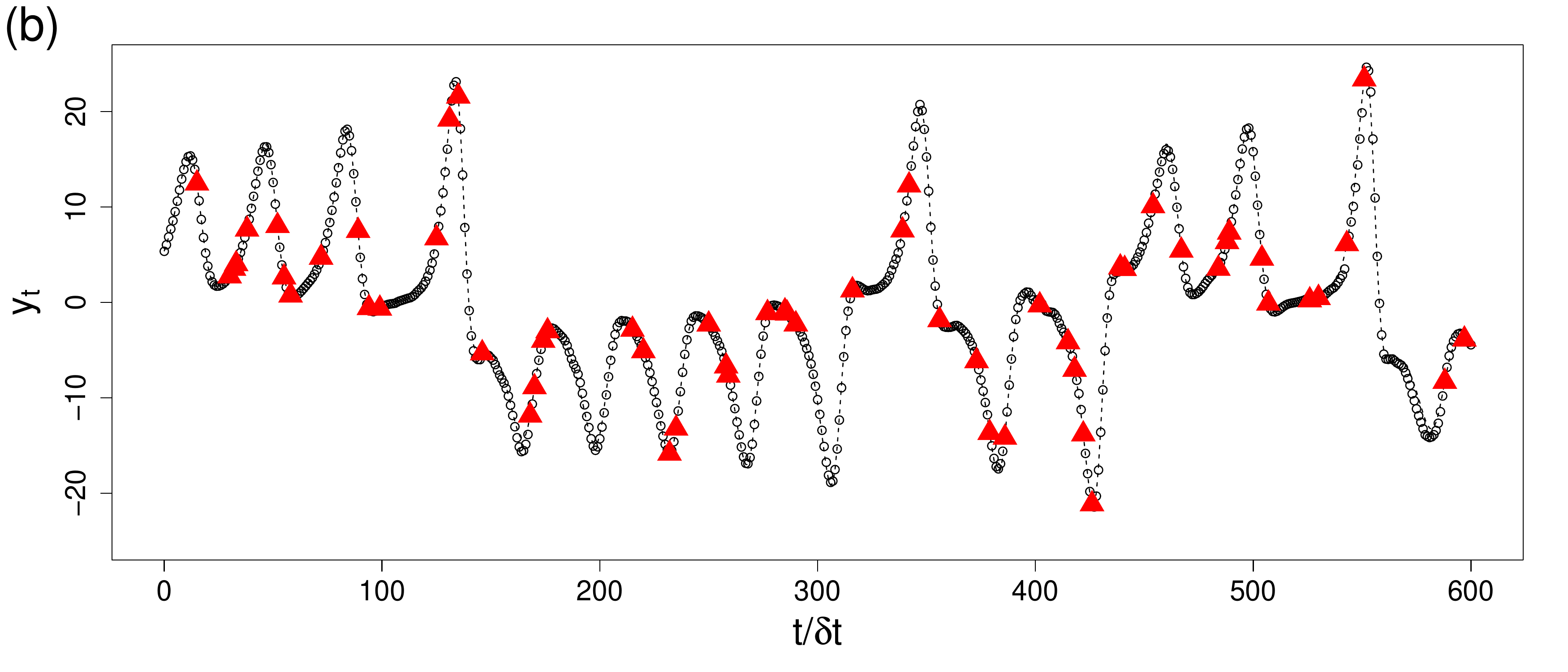}\\
  \includegraphics[width=0.7\textwidth]{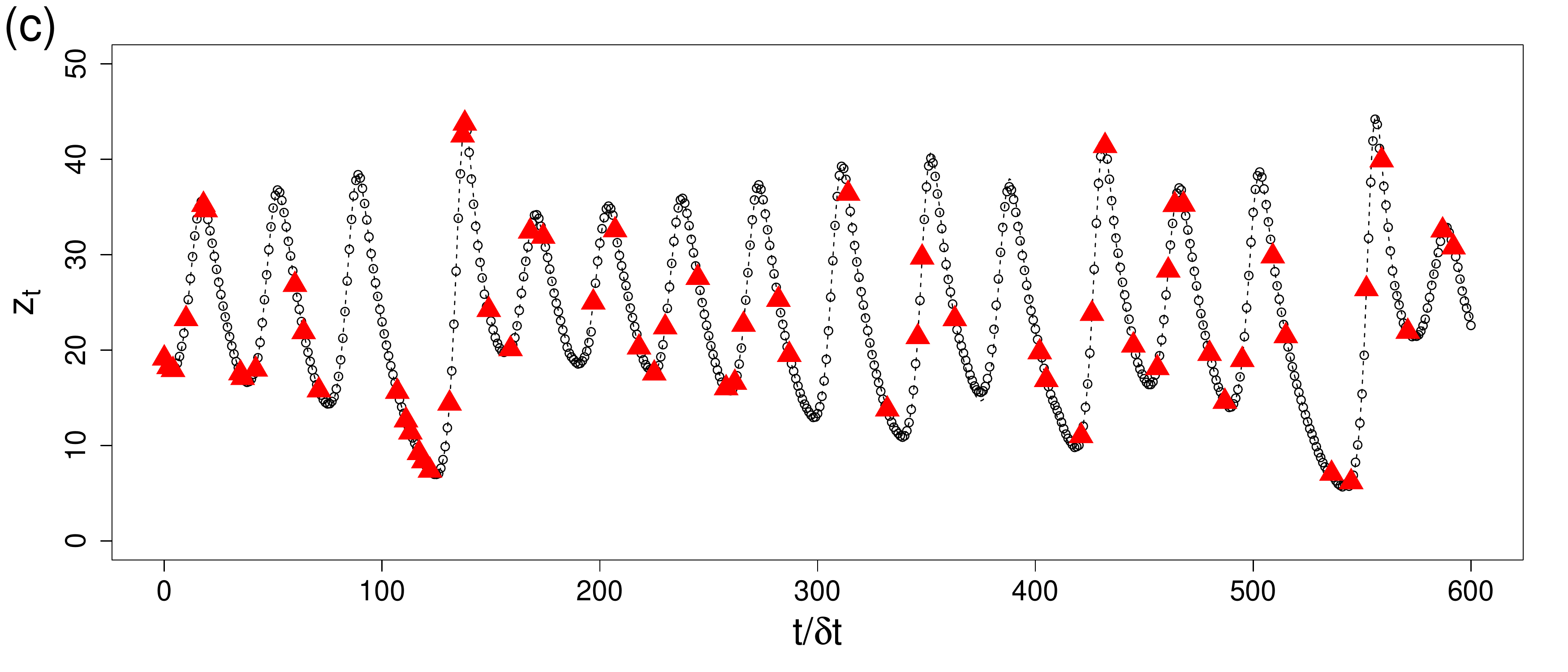}
  \caption{Reconstruction of the sparse Lorenz-63 system ($\omega = 0.9$); (a) $x_t$, (b) $y_t$, and (c) $z_t$.  In (a--c), the dashed line is the ground truth, the solid symbols (${\color{red}\blacktriangle}$) denote the sparse observation, and the reconstructed dynamics is shown as the hollow symbols ($\circ$).} \label{fig:Lorenz63_reconst_all}
\end{figure}

Figure \ref{fig:Lorenz63_reconst_all} shows the reconstructed Lorenz-63 time series from the 90\% randomly missing data after 90 iterations. The relaxation and regularization parameters are, respectively, $\alpha = 0.2$ and $\beta = 10^{-6}$. It is shown that the fixed-point ESN is capable of accurately reconstruct the nonlinear dynamics of the Lorenz-63 system.

\begin{table}
\center{
\caption{Normalized RMSE of the multivariate reconstruction of the Lorenz time series.} 
\label{tbl:full_Lorenz63_error}
\begin{tabular}{c|cc}
\hline \hline
$\omega$ & 0.90 & 0.95\\
\hline
$\sigma_{lin}$ & 0.18 & 0.51  \\
$\sigma_{csp}$ & 0.17 & 0.36 \\
\hline \hline
\end{tabular}
}
\end{table}

The normalized RMSEs with respect to $\bm{Y}^*$ are shown in table \ref{tbl:full_Lorenz63_error} for $\omega = 0.90$ and 0.95. The normalized RMSE is defined as,
\[
\sigma_{intp} = \left[ \frac{1}{N_y} \sum_{k=1}^{N_y} \left\{ \frac{\sum_i (y_{i_k}^{R} - y^*_{i_k})^2}{\sum_i (y_{i_k}^{intp} - y^*_{i_k})^2} \right\}\right]^{1/2}.
\]
For $\omega = 0.90$, the fixed-point ESN makes a much more accurate reconstruction compared to the linear and cubic spline interpolations. However, the accuracy of the fixed-point ESN significantly decreases at $\omega = 0.95$. The fixed-point ESN still outperforms those interpolation methods. But, there is about three-fold increase in $\sigma_{lin}$ as $\omega$ changes from 0.9 to 0.95. At $\omega = 0.95$, the cubic spline interpolation results in a large error due to the overshoot, which makes $\sigma_{csp}$ smaller than $\sigma_{lin}$.


\subsection{Lorenz-96 System}\label{sec:Lorenz_96}

In this experiment, the fixed-point ESN is tested against a 6-node Lorenz-96 system \cite{Lorenz96}; 
\begin{equation}
\frac{d y^{(i)}}{dt} = -y^{(i-2)}y^{(i-1)}+y^{(i-1)}y^{(i+1)}-y^{(i)}+F,~\text{for}~i=1,\cdots,N_y.
\end{equation}
The Lorenz-96 system is periodic, i.e., $y^{(i-N_y)} = y^{(i+N_y)} = y^{(i)}$. Here, we consider $N_y = 6$ and $F = 8$. The time series data is generated by sampling at every $\delta t = 0.02$ and the length of the time series data is $T = 5\times10^4$. 

\begin{figure}
  \centering
  \includegraphics[width=0.48\textwidth]{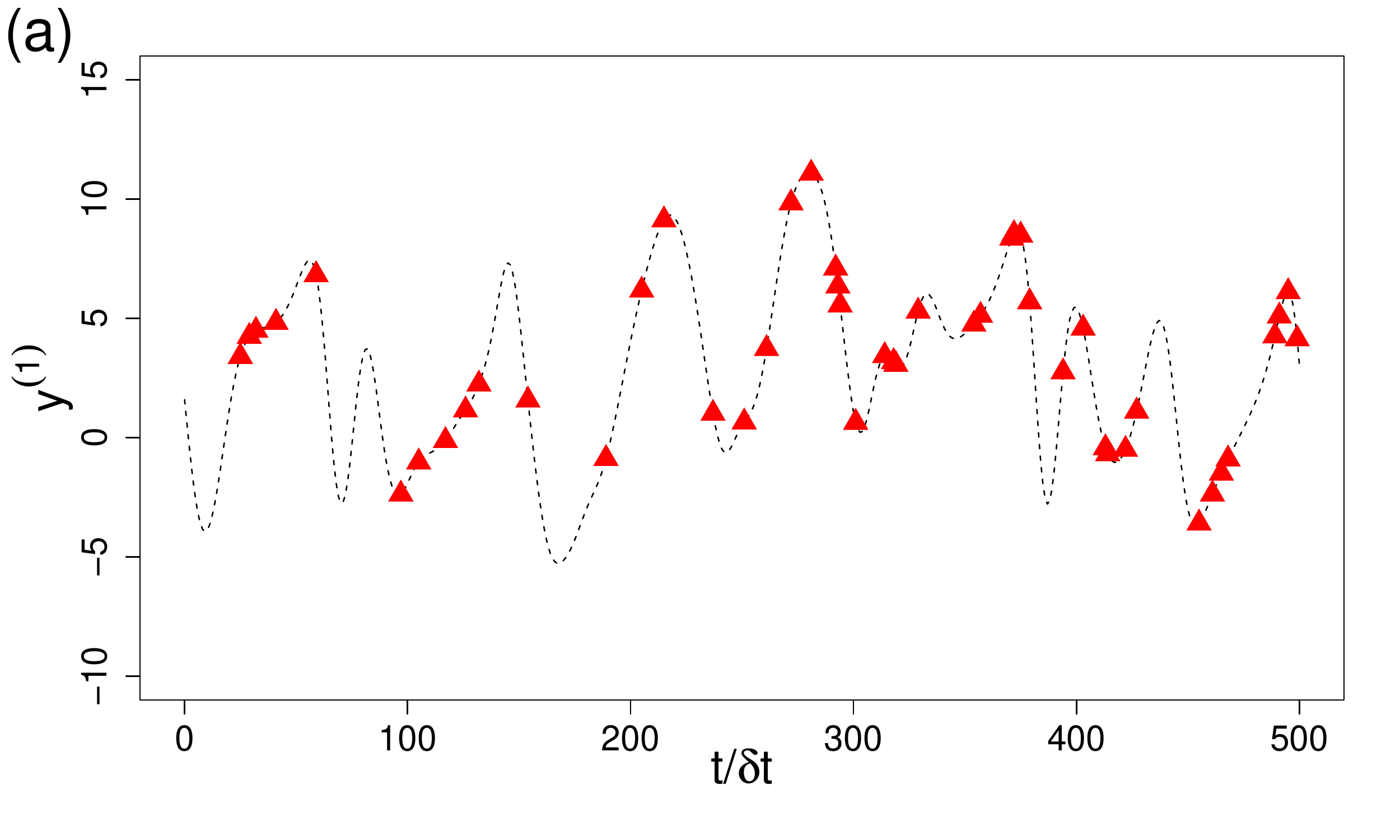}
  \includegraphics[width=0.48\textwidth]{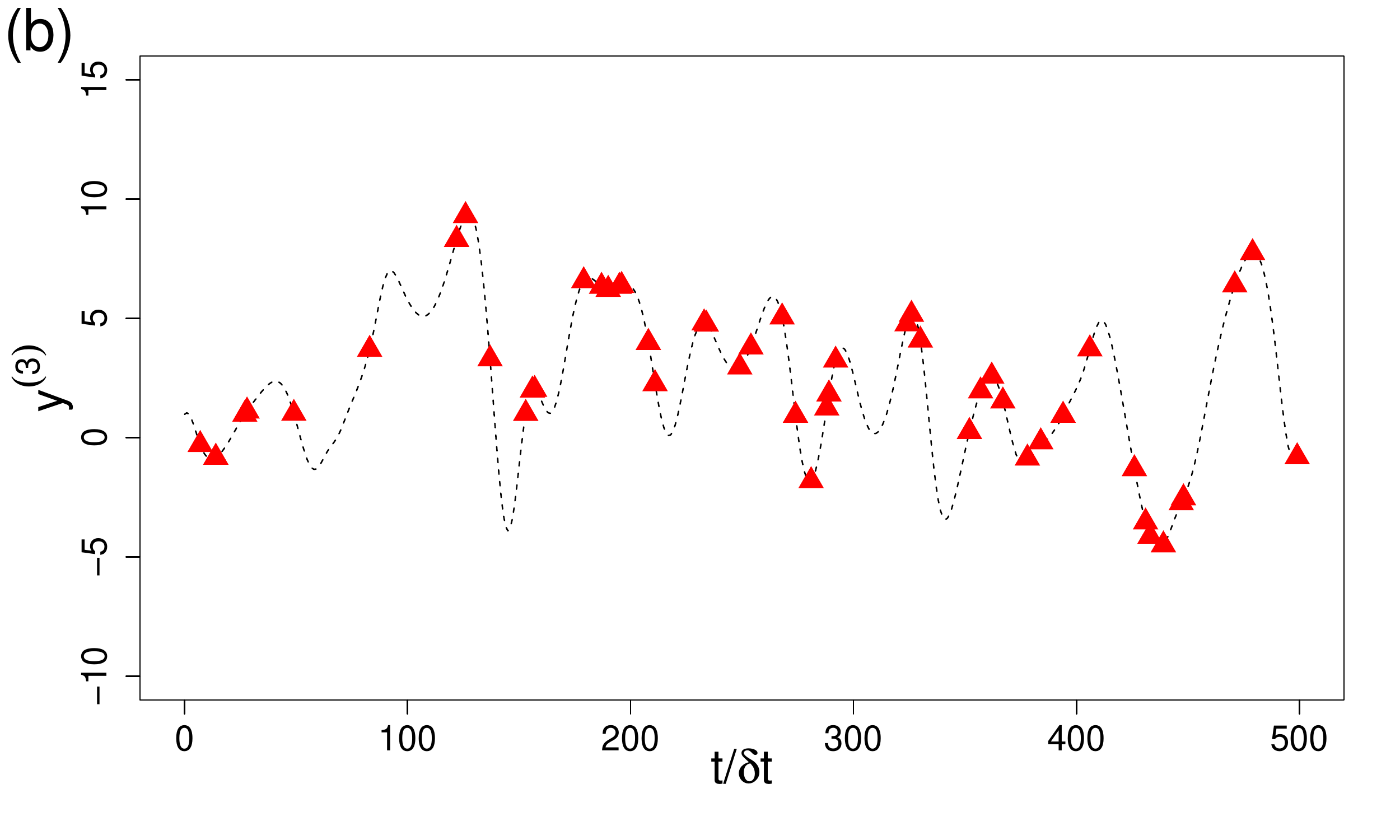}\\
  \includegraphics[width=0.48\textwidth]{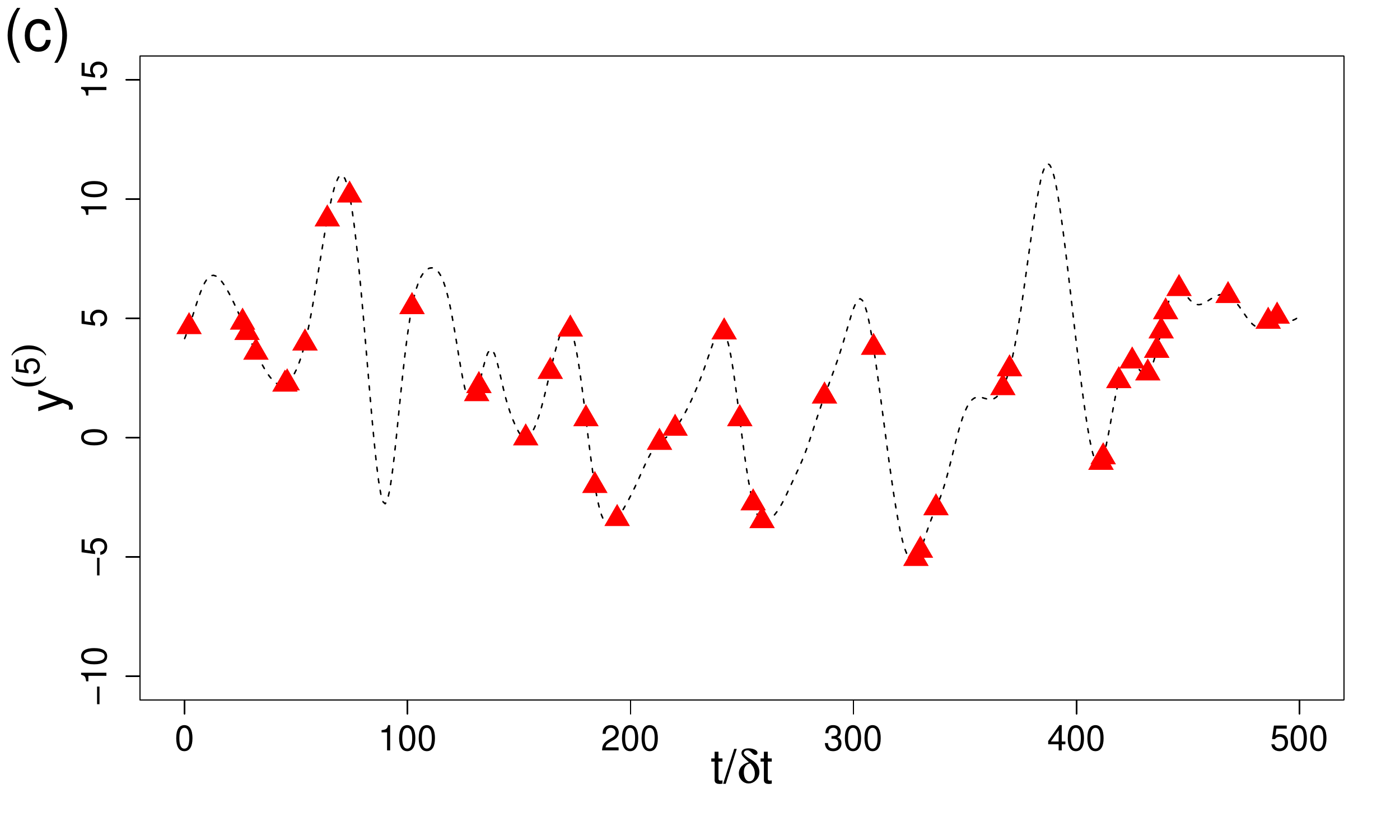}
  \includegraphics[width=0.48\textwidth]{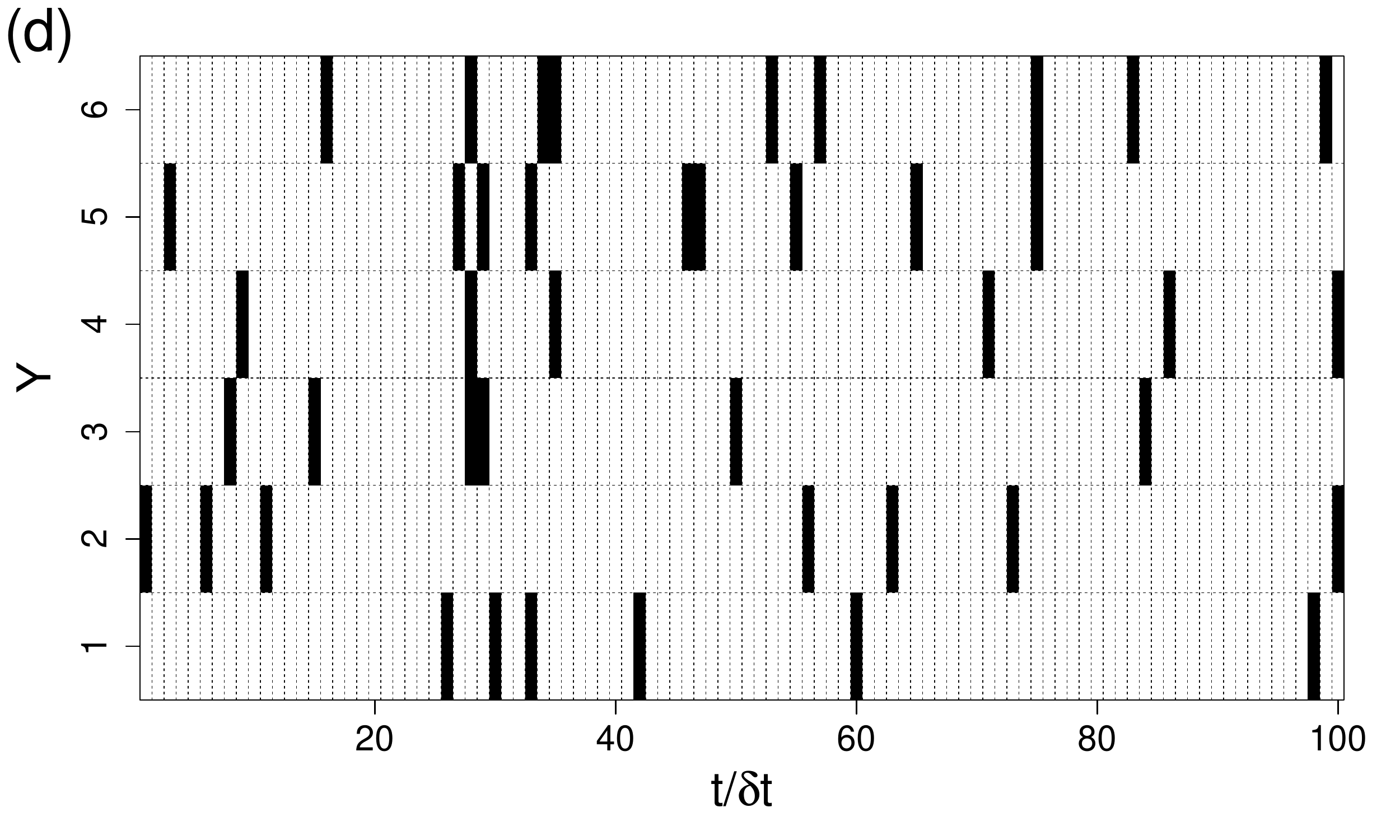}
  \caption{Sparse observation of the Lorenz-96 system for $\omega = 0.9$; (a) $y^{(1)}(t)$, (b) $y^{(3)}(t)$, and (c) $y^{(5)}(t)$.  In (a--c), the dashed line is the ground truth and the solid symbols (${\color{red}\blacktriangle}$) denote the sparse observation. An example of the data availability is shown in (d). The location of the observation is marked by the black bars.} \label{fig:Lorenz96_data_all}
\end{figure}

Figure \ref{fig:Lorenz96_data_all} (a--c) show a subset of the 6-node Lorenz-96 system for $\omega=0.9$. Figure \ref{fig:Lorenz96_data_all} (d) shows an example of the data availability. Because the probability that every $y^{(i)}$ is observed at the same time frame is $10^{-6}$ for $\omega = 0.9$, the data set does not contain a time frame at which every $y^{(i)}$ is observed.

\begin{figure}
  \centering
  \includegraphics[width=0.48\textwidth]{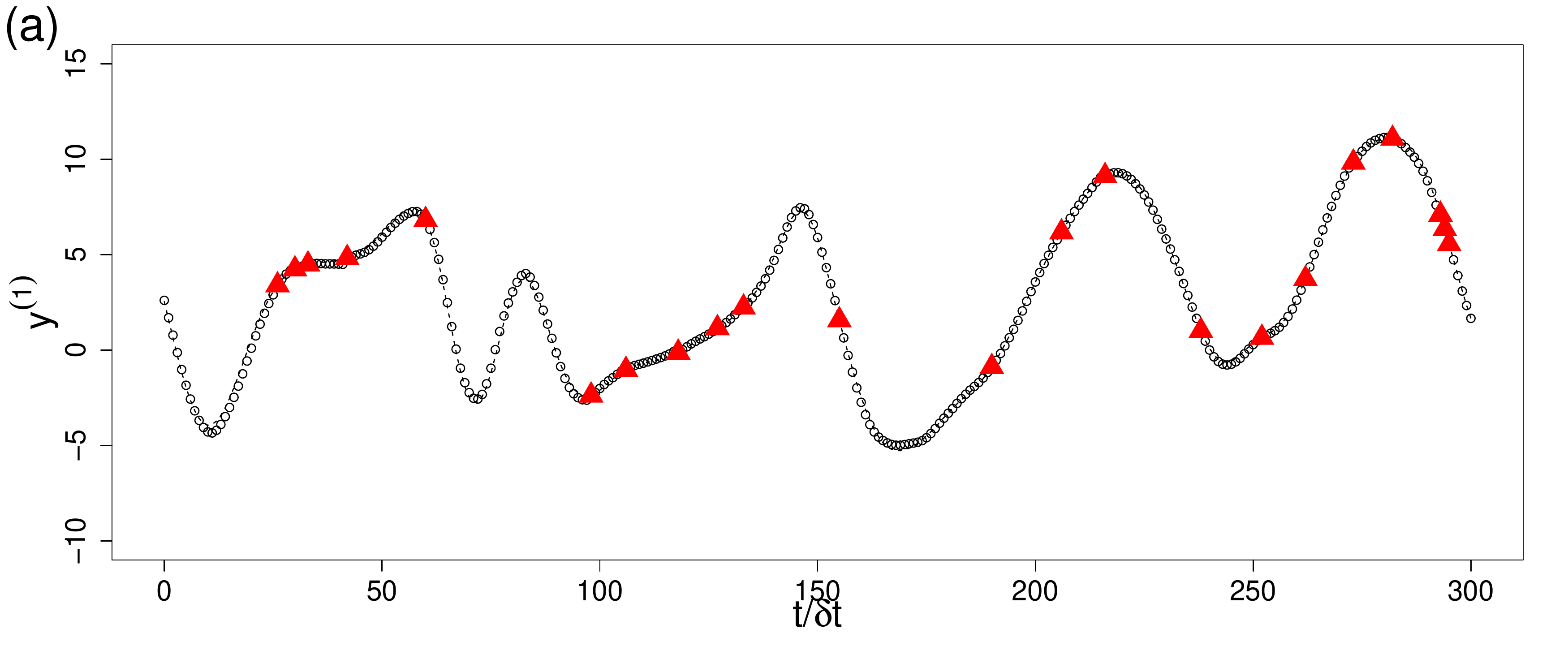}
  \includegraphics[width=0.48\textwidth]{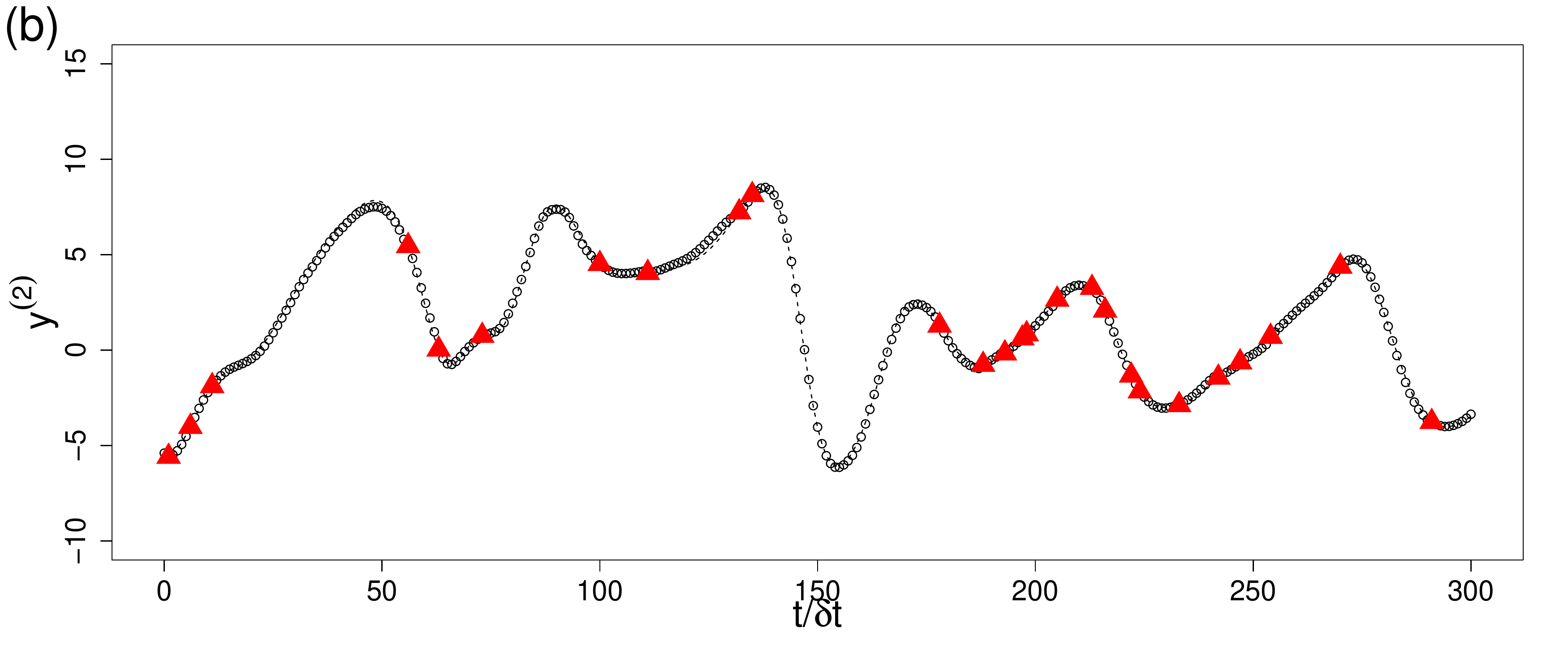}\\
  \includegraphics[width=0.48\textwidth]{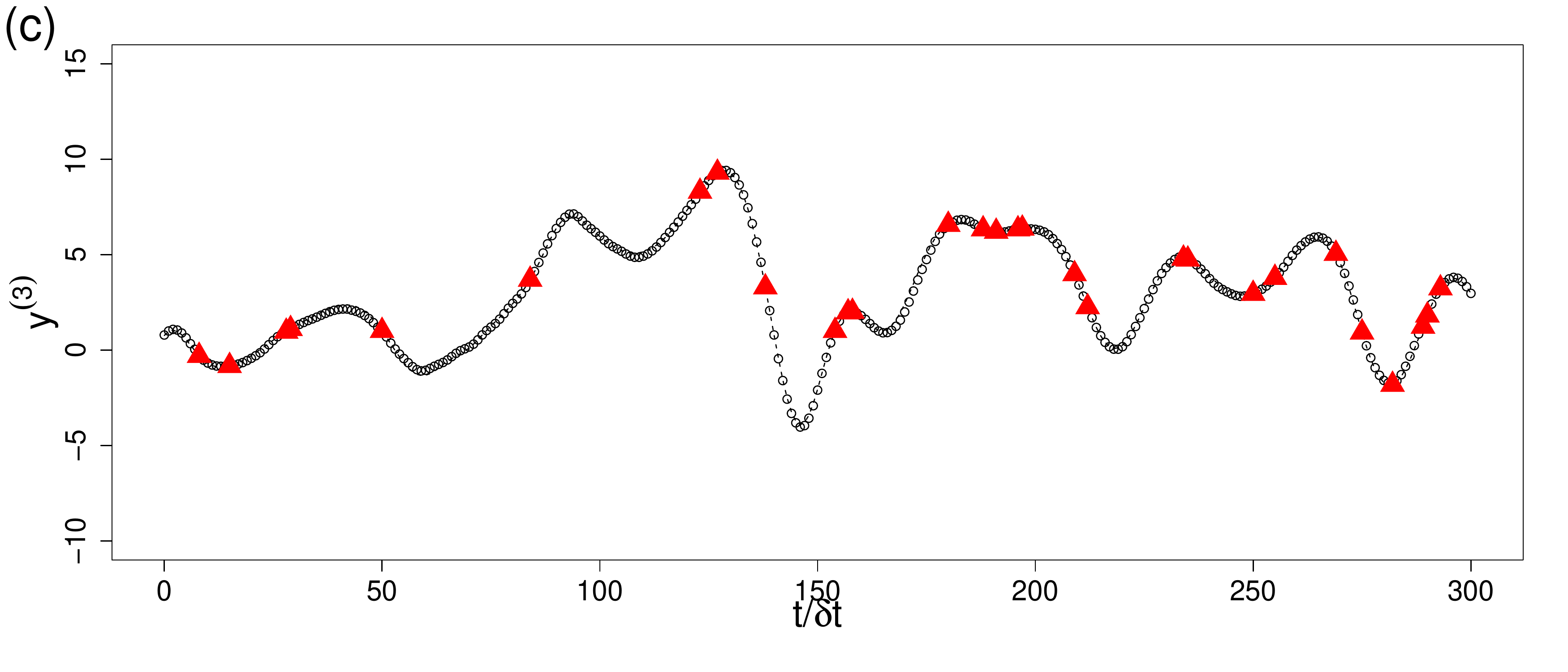}
  \includegraphics[width=0.48\textwidth]{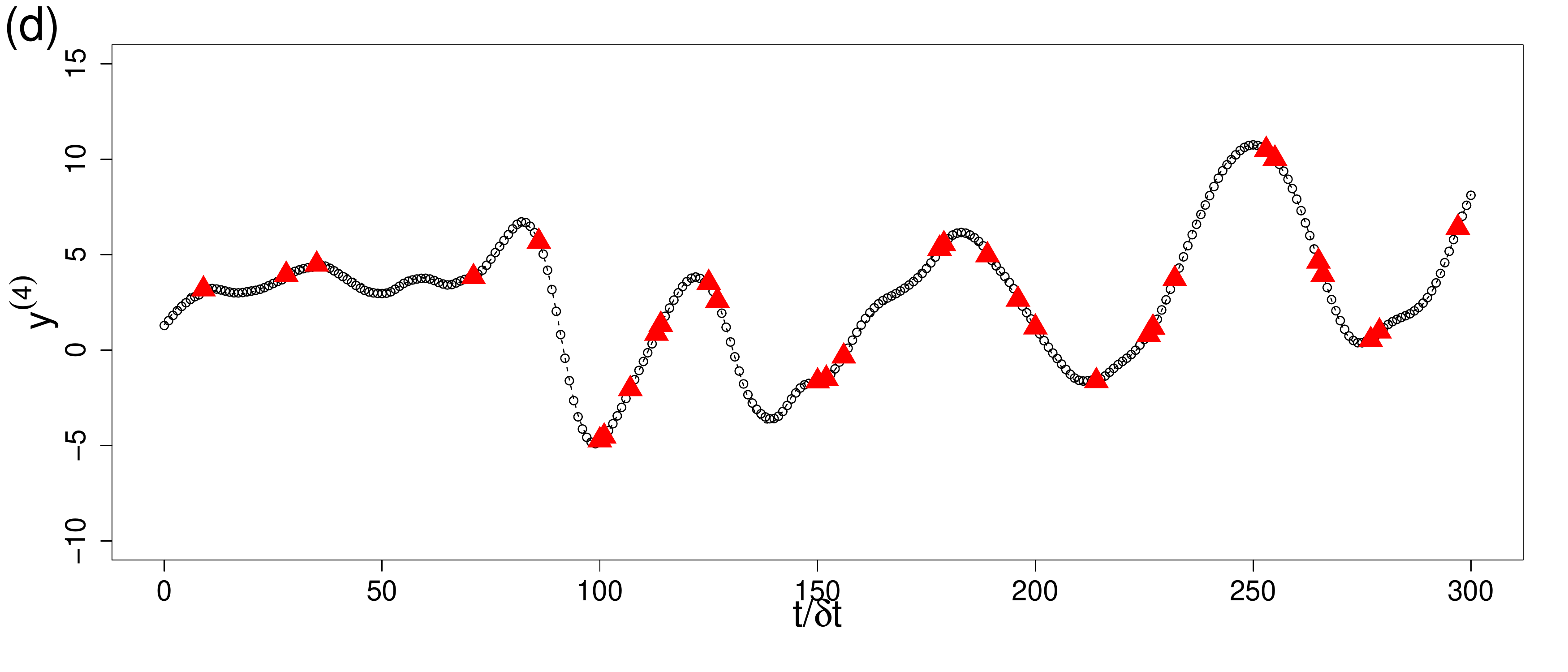}\\
  \includegraphics[width=0.48\textwidth]{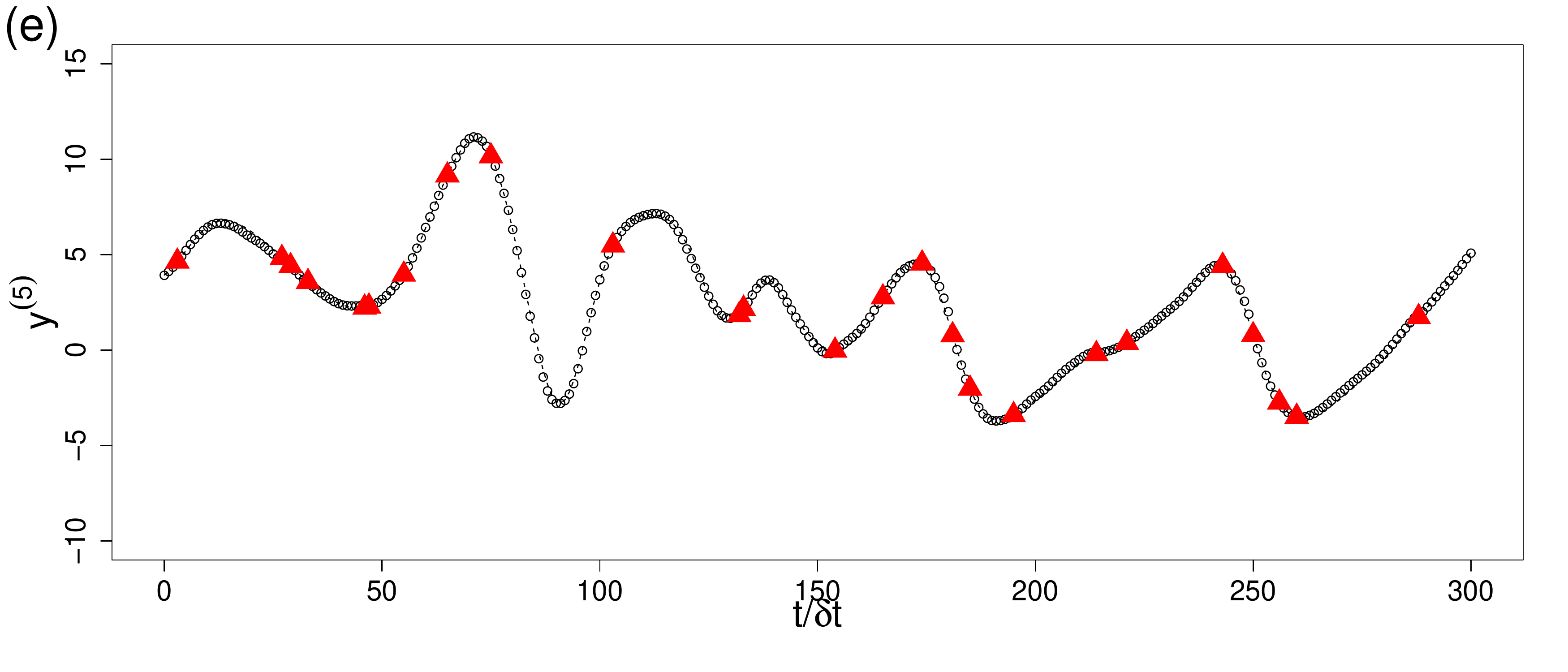}
  \includegraphics[width=0.48\textwidth]{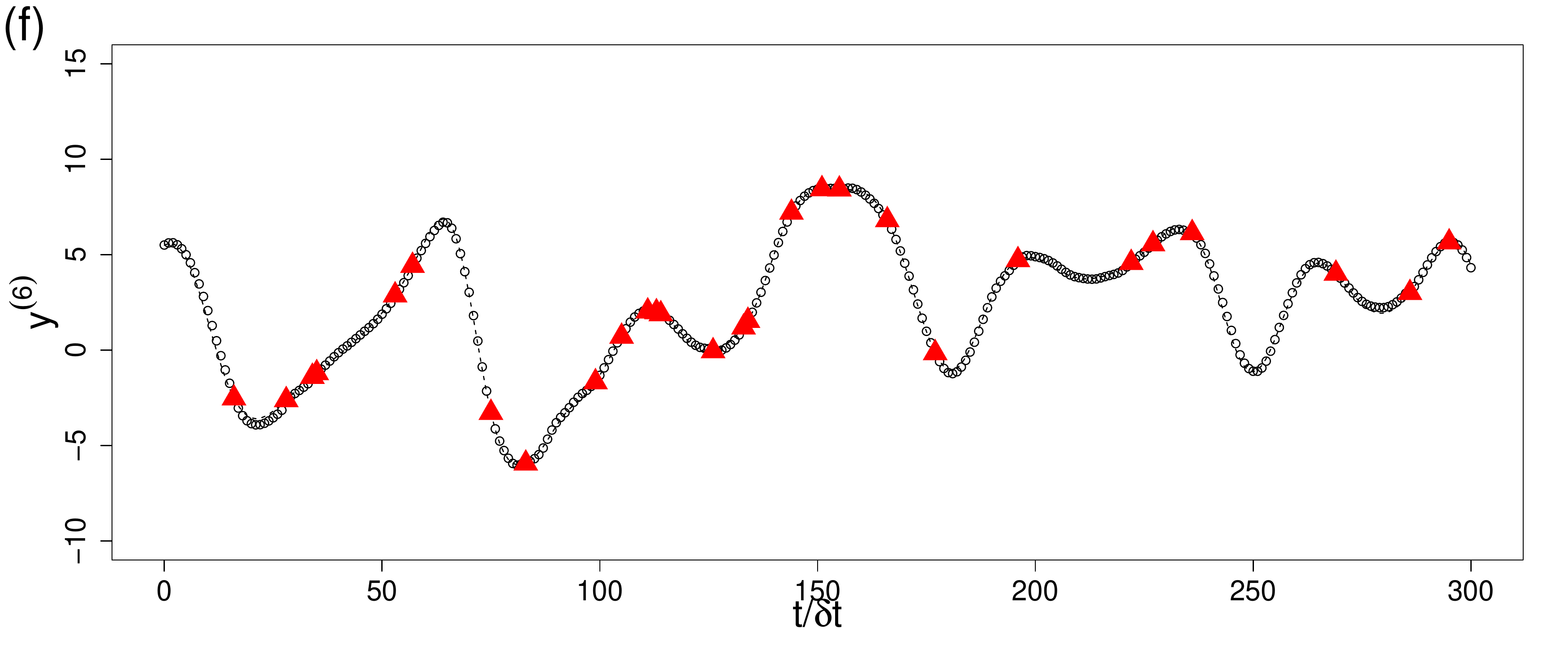}
  \caption{Reconstruction of the 6-node Lorenz-96 system from a sparse observation ($\omega = 0.9$).  The dashed line is the ground truth, the solid symbols (${\color{red}\blacktriangle}$) denote the sparse observation, and the reconstructed dynamics is shown as the hollow symbols ($\circ$).} \label{fig:Lorenz96_reconst_all}
\end{figure}

Figure \ref{fig:Lorenz96_reconst_all} shows the reconstructed 6-node Lorenz-96 system from the sparse observation of $\omega = 0.9$. The parameters of the fixed-point ESN are $\alpha = 0.2$ and $\beta = 10^{-7}$. It is again shown that $\bm{Y}^R$ is almost indistinguishable from $\bm{Y}^*$. The normalized RMSEs are $\sigma_{lin} = 0.11$ and $\sigma_{csp} = 0.10$. When the fixed-point ESN is used to recover the dynamics from a sparser data set of $\omega = 0.95$, the normalized RMSEs are significantly increased to $\sigma_{lin} = 0.57$ and $\sigma_{csp} = 0.46$.

\subsection{Forced van der Pol oscillator}\label{sec:VDP}

Now, we consider a forced van der Pol oscillator, which is given by the following equations,
\begin{align}
\frac{dy_1}{dt} &= y_2, \label{eqn:VDP_1}\\
\frac{dy_2}{dt} &= \gamma_1 (1-y_1^2)y_2 - y_1 + u(t). \label{eqn:VDP_2}
\end{align} 
The exogenous forcing, $u(t)$, is given by an Ornstein-Uhlenbeck process as
\begin{equation} \label{eqn:VDP_OU}
du = -\gamma_2 u dt + \gamma_3 dW,
\end{equation}
in which $W$ is the Wiener process.
The parameters used in this simulation are, $\gamma_1 = 2$, $\gamma_2 = 0.2$, and $\gamma_3 = 5\sqrt{2\gamma_2}$. Similar to the partial observation of the Lorenz-63 system, $y_2$ is not given and the data set consists of  $(y_1(t),u(t))$, which is sampled at every $\delta t = 0.5$. Hereafter, we use $y_t$ to denote $y_1(t)$. The length of the time series is $T = 5 \times 10^4$.

\begin{figure}
  \centering
  \includegraphics[width=0.48\textwidth]{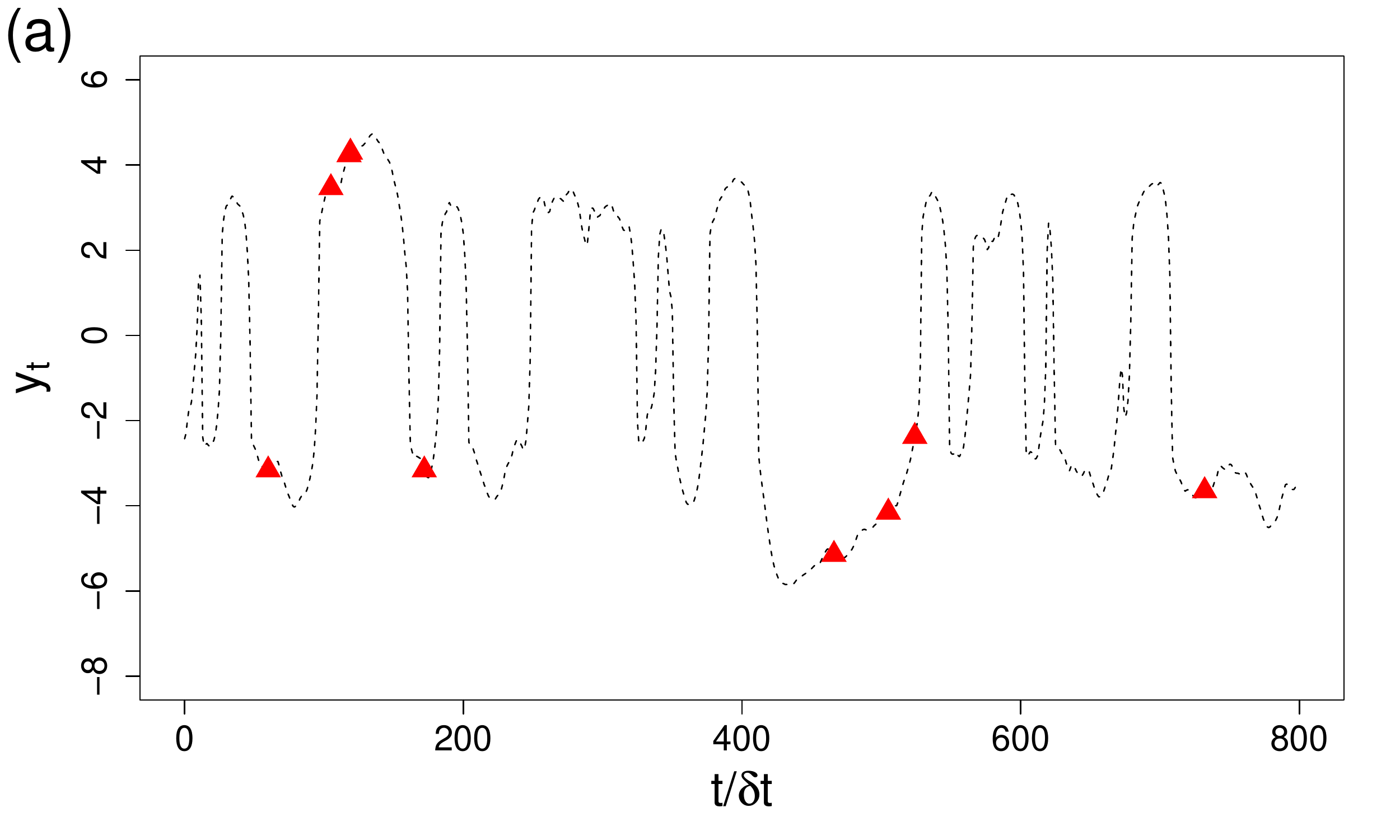}
  \includegraphics[width=0.48\textwidth]{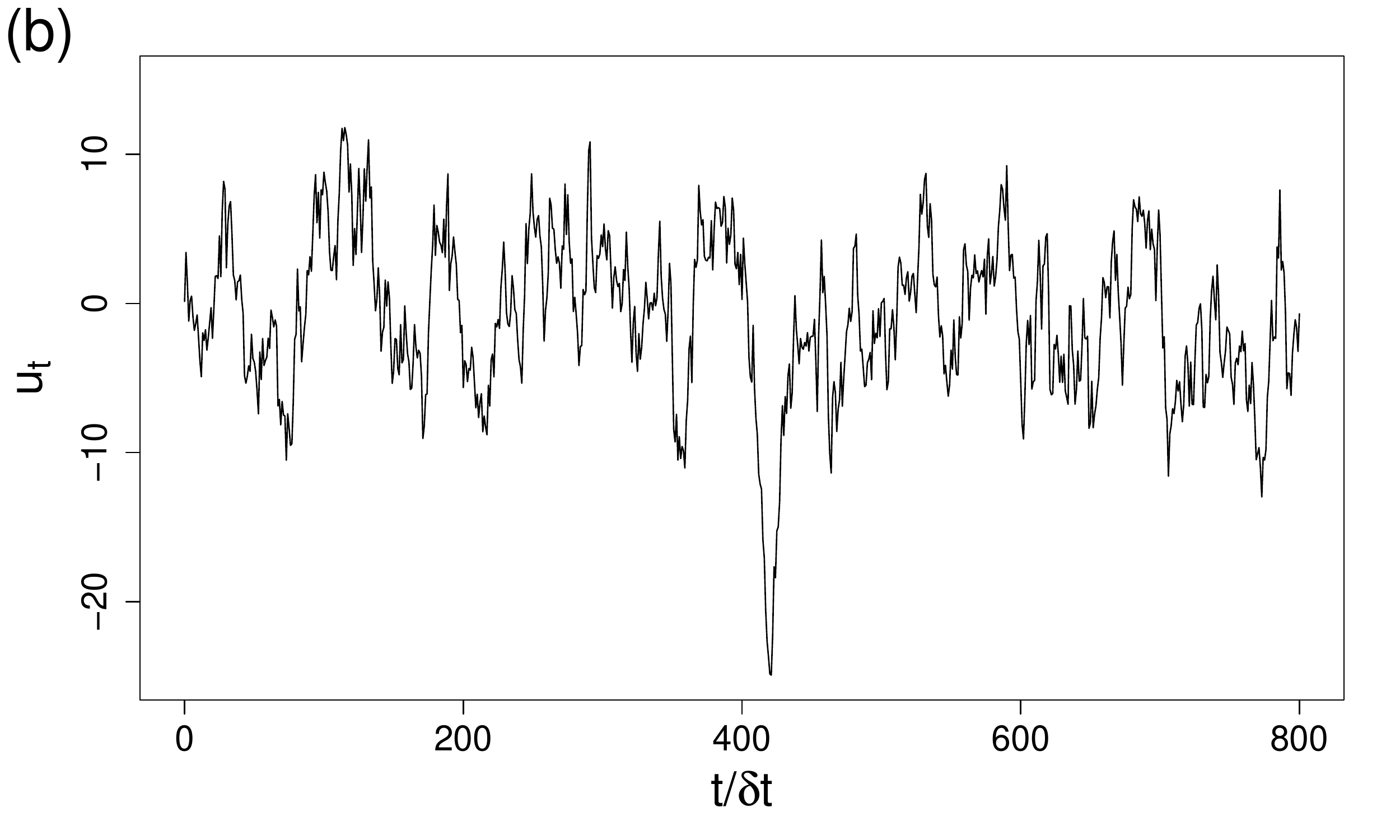}
  \caption{ The sparse data set for the forced van der Pol oscillator. (a) shows the target variable, $y_1(t)$, for the missing fraction $\omega = 0.99$. The dashed line is the ground truth ($\bm{Y}^*$) and the sparse observation ($\bm{Y}^O$) is denoted by the solid triangles ({\color{red}$\blacktriangle$}). (b) shows the corresponding exogenous forcing, $u_t$.} \label{fig:VDP_data}
\end{figure}

In sections \ref{sec:MG} -- \ref{sec:Lorenz_96}, we consider the reconstruction of chaotic time series. As shown in figure \ref{fig:Lorenz63_reconst_X}, the accuracy of the reconstruction becomes lower as the interval of the missing information increases. The decrease in the accuracy seems to be related with the chaotic nature of the time series. On the other hand, the unforced van der Pol oscillator has a stable limit cycle. The dynamics of the forced van der Pol oscillator in (\ref{eqn:VDP_1}--\ref{eqn:VDP_2}) depends on the restoring force to the limit cycle and excursion due to the exogenous forcing, $u_t$. Hence, in theory, if we know the governing equations and the full history of $u_t$, we can find $\bm{Y}^*$ no matter how long the interval between the observations.

\begin{figure}
  \centering
  \includegraphics[width=0.8\textwidth]{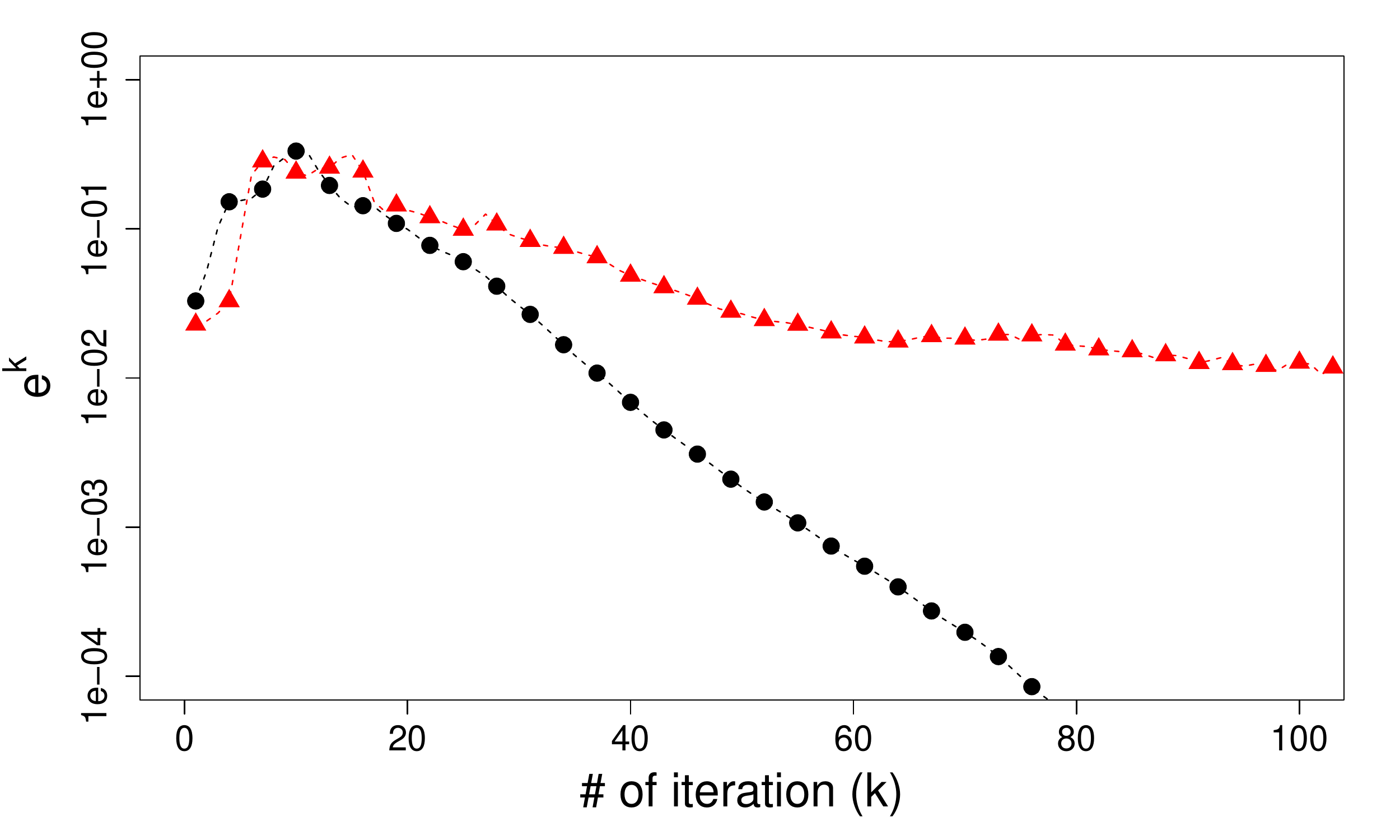}
  \caption{Changes in the $l_2$-improvement with respect to the number of iterations for $\omega = 0.98$ ($\bullet$) and 0.99 ($\color{red}\blacktriangle$). } \label{fig:VDP_converge}
\end{figure}

The sparse observation of the van der Pol oscillator is shown in figure \ref{fig:VDP_data}. Here, we consider an extremely sparse data, in which 99\% of the observations are randomly removed. Since $T=5\times10^4$, the number of observations is only 500. Although the average interval of the missing data is $100 \delta t$, due to the random removal, it is not uncommon to find a missing interval of $200\delta t \sim 300 \delta t$. The exogenous forcing is shown in figure \ref{fig:VDP_data} (b). While the Ornstein-Uhlenbeck process is computed with $dt = 2.5\times10^{-3}$ to numerically integrate (\ref{eqn:VDP_1}--\ref{eqn:VDP_2}), $u_t$ is downsampled with the sampling interval of $\delta t = 0.5$. In a sense, we introduce an uncertainty, in which the sub-timescale information of $u_t$ is unknown. Hence, the time series becomes a Markov process, $y_{t+1} \sim p(y_{t+1}|y_t,u_t)$. Although the subscale fluctuations of $u_t$ are not significant, it breaks the assumption of the fixed-point ESN that there exists a deterministic mapping, $y_{t+1} = f(y_t,u_t)$, which guarantees the existence of a fixed point.

Figure \ref{fig:VDP_converge} shows the $l_2$-improvement in terms of the iteration count. In both cases, $\alpha = 0.2$ and $\beta = 10^{-6}$ are used. It is shown that, for $\omega = 0.98$, the fixed-point ESN quickly converges to a solution. The $l_2$-improvement becomes $\simeq 10^{-6}$ by the 100-th iteration. When only 1\% of the data is available ($\omega = 0.99$), the convergence becomes much slower. The $l_2$-improvement reaches $e^k \simeq 0.008$ at the iteration number $k = 160$, and there is no further reduction of $e^k$ afterward. 

\begin{figure}
  \centering
  \includegraphics[width=0.48\textwidth]{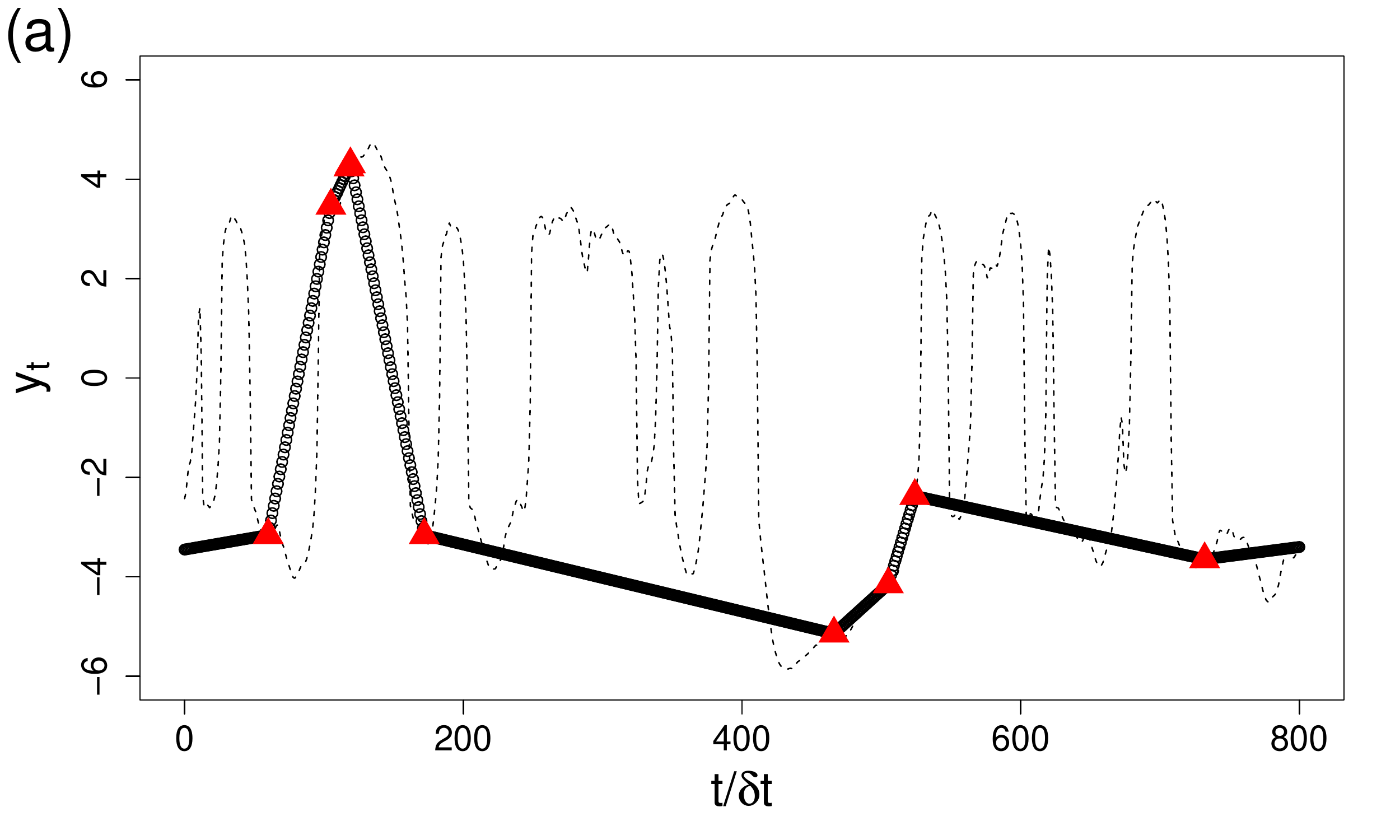}
  \includegraphics[width=0.48\textwidth]{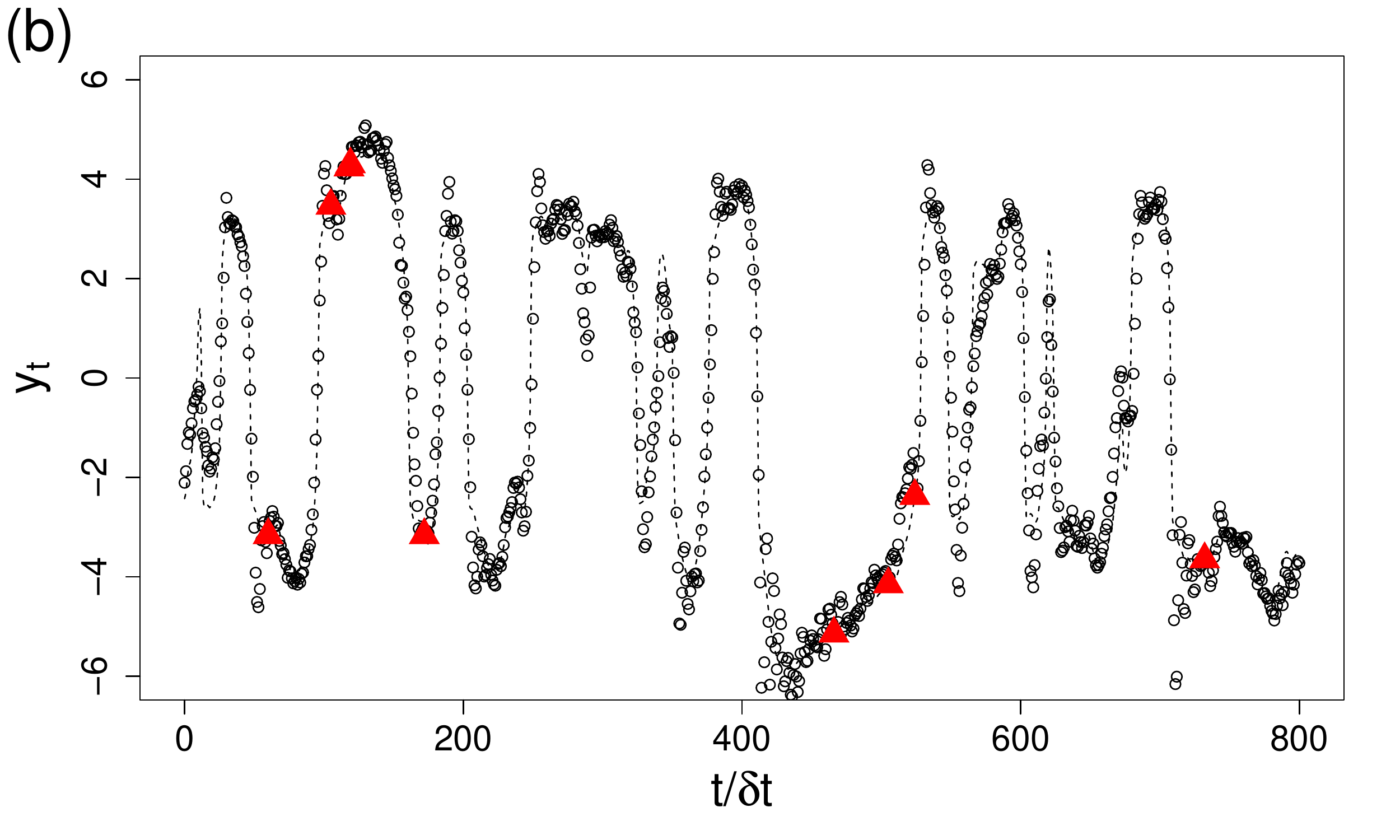}
  \caption{Reconstruction from the sparse observation of the forced van der Pol oscillator. $u_t$ is fully observed, while 99\% of $y_t$ is missing. The dashed line is the ground truth ($\bm{Y}^*$) and the sparse observation ($\bm{Y}^O$) is the solid triangles ({\color{red}$\blacktriangle$}). The hollow circles ($\circ$) are the reconstructed trajectory, $\bm{Y}^{R,\,k}$, at (a) $k =0$, and (b) 150.} \label{fig:VDP_iter}
\end{figure}

Figure \ref{fig:VDP_iter} shows the reconstructed dynamics for $\omega = 0.99$. Even with the extremely sparse data, the fixed-point ESN is able to reconstruct the nonlinear dynamics. The normalized RMSE is $\sigma_{lin} = 0.25$ for $\omega = 0.99$ and $\sigma_{lin} = 0.21$ for $\omega = 0.98$. It is observed that, although the reconstructed trajectory, $\bm{Y}^R$, closely follows the dynamics, $\bm{Y}^R$ is much noisier compared to the previous experiments on the deterministic time series, which seems to be related with the uncertainty in $u_t$.

\section{Concluding remarks} \label{sec:conclusions}

In this study, a novel model-free method is developed to reconstruct nonlinear dynamics from a temporally sparse observation, in which a large fraction of the time series data is randomly missing, or unobserved. 
Based on the assumptions of the noise-free observation and universal function approximation capability of a recurrent neural network, we show that the reconstruction problem can be solved by a fixed-point problem. The fixed-point method consists of two major components; outer-iteration to find the fixed-point solution and inner-loop to compute the parameters of the recurrent neural network. In theory, any recurrent neural network architecture can be used for the inner-loop, but the computational cost of solving the inner-loop with a stochastic gradient method will make it impractical. Hence, we employ the echo state network for the recurrent neural network, of which parameter can be computed by solving a simple ordinary least square problem. The fixed-point method is simple to implement and can be solved with $O(N_s^2 \times T)$ floating point operations, because it does not require to compute the derivatives of the nonlinear operator. Although it may be possible to develop a gradient-based optimization algorithm to solve the problem, computing the derivatives of the echo state network will require additional memory of $O(N_s \times T)$ and floating operation of $O(N_s^2 \times T^2)$, which makes it difficult to use in practice.

The proposed method, referred as the fixed-point ESN, is tested against time series data generated by chaotic dynamical systems. For the Mackey-Glass time series, it is shown that the fixed-point ESN can accurately reconstruct the nonlinear dynamics even when the 95\% of the data are randomly removed. For the missing fraction $\omega = 0.95$, the average time interval between two consecutive observations is $\Delta T = 20$, while the characteristic period of the Mackey-Glass time series is $T_c \simeq 50$ \cite{Gers01}. The Lorenz-63 and 6-node Lorenz-96 systems are used to demonstrate the capability of the fixed-point ESN for the reconstruction of nonlinear dynamics of multivariate time series. In both cases, the fixed-point ESN provides very good approximation of the complex nonlinear dynamics up to $\omega = 0.9$. Finally, we use a forced van der Pol oscillator, of which dynamics is largely determined by the exogenous forcing, to investigate the behavior of the fixed-point ESN. It is shown that the fixed-point ESN can learn the relation between the exogenous forcing and the dynamical system with only 1\% of the time series data. 

It is demonstrated that the echo state network is capable of learning the dynamics from only a partial observation of a complex nonlinear dynamical system. 
In this study, we limit our interest only to relatively low dimensional dynamical systems. It is a subject of further investigation how the nonlinear reconstruction method can be extended to a high dimensional dynamical system with complex correlation structures, such as a spatio-temporal process. 
In many practical applications, observations are typically corrupted by a sensor noise, which makes the observation a stochastic process. Although, it is shown that the fixed-point ENS still can learn the nonlinear dynamics for a small noise, the accuracy of the fixed-point ESN is not guaranteed under a small signal-to-noise ratio. Finally, although it is empirically shown that the fixed-point ESN is very effective in the nonlinear reconstruction, our understanding on the theoretical understanding, \emph{e.g.}, accuracy and convergence, is far from complete.

\appendix

\section{Convergence analysis of fixed-point ESN} \label{sec:convergence}

It is challenging to theoretically show the convergence of a fixed point iteration of a nonlinear map. One of the widely used method to analyze the convergence is based on the Banach fixed-point theorem, which provides a \emph{sufficient} condition. The Banach fixed point theorem requires a contraction mapping;
\begin{equation}
\| \bm{\mathcal{G}}(\bm{Y}^a,\bm{U}) - \bm{\mathcal{G}}(\bm{Y}^b,\bm{U})\| \le L \| \bm{Y}^a - \bm{Y}^b\|,
\end{equation}
in which $0 \le L < 1$ is a Lipschitz constant. In \cite{Lu18}, it is \emph{assumed} that ESN provides a contraction mapping. However, it is a too strong assumption and in general such a globally uniform contraction is not expected for a nonlinear map. Instead, here, we explore a local convergence around a fixed point, $\bm{Y}^*$.

Let $\mathcal{J}_\mathcal{G}(\bm{Y})$ denote the Jacobian of $\bm{\mathcal{G}}(\bm{Y},\bm{U})$ in (\ref{eqn:esn_fixed}). From the contraction mapping theorem, it can be shown that, if  $\|\mathcal{J}_{\mathcal{G}}(\bm{Y}^*)\| < 1$ for a subordinate matrix norm $\| \cdot \|$, there is a neighborhood around the fixed point, $\bm{Y}^*$, in which the fixed-point iteration converges to $\bm{Y}^*$. 

For simplicity, here we assume that $y_t$ is univariate, i.e., $N_y=1$. But, extending the analysis to a multivariate time series is straightforward. The Jacobian of the fixed point iteration is given as
\begin{align}
[\mathcal{J}_\mathcal{G}(\bm{Y})]_{ij} &= \frac{\partial}{\partial y_j}\left[ \alpha y_i + (1-\alpha)\bm{\theta}^T\bm{S}_i\right] \nonumber \\
&= \alpha \delta_{ij} + (1-\alpha)\left\{ \frac{\bm{\partial \theta}^T}{\partial y_j}\bm{S}_i + \bm{\theta}^T \frac{\partial \bm{S}_i}{\partial y_j} \right\}, \label{eqn:Jacobian}
\end{align}
in which $\delta_{ij}$ is the Kronecker delta. 

In the second term on the right hand side (RHS) of (\ref{eqn:Jacobian}),
\begin{align}
\frac{\partial \bm{\theta}}{\partial y_j} &= \frac{\partial}{\partial y_j} \left\{ \left( \sum_{l\in \bm{M}_1}\bm{S}_l\bm{S}_l^T + \beta\bm{I}\right)^{-1}  \left( \sum_{l\in \bm{M}_1} \bm{S}_l y^O_l \right)\right\} \nonumber \\
&= -\bm{D} \left(\frac{\partial}{\partial y_j}\sum_{l\in\bm{M}_1} \bm{S}_l\bm{S}_l^T\right)\left( \bm{D} \sum_{l\in\bm{M}_1} \bm{S}_l y^O_l \right) + \bm{D} \left(\frac{\partial}{\partial y_j} \sum_{l\in\bm{M}_1} \bm{S}_l y^O_l\right). \label{eqn:dtheta-1}
\end{align}
Here, $\bm{M}_1$ is a set of nonzero entries of $\bm{M}$, i.e., the summation is over the available observations, and $\bm{D} = (\sum \bm{S}\bm{S}^T + \beta \bm{I})^{-1}$. By definition, 
\[
\bm{\theta} = \bm{D}\sum_{l\in\bm{M}_1} \bm{S}_ly^O_l,
\]
and, from (\ref{eqn:y_star-1}), $y^O_l = {\bm{\theta}^*}^T\bm{S}^*_l$. Then, (\ref{eqn:dtheta-1}) can be written as
\begin{equation}
\frac{\partial \bm{\theta}}{\partial y_j} = -\bm{D} \left[ \left( \frac{\partial}{\partial y_j} \sum_{l\in\bm{M}_1} \bm{S}_l\bm{S}_l^T \right) \bm{\theta} - \left(\frac{\partial}{\partial y_j} \sum_{l\in\bm{M}_1} \bm{S}_l{\bm{S}^*_l}^T \right) \bm{\theta}^* \right].
\end{equation}
At a fixed point,  $\bm{Y} = \bm{Y}^*$, we have $\bm{S}_l = \bm{S}^*_l$ and $\bm{\theta} = \bm{\theta}^*$, which makes $\partial \bm{\theta} / \partial y_j = 0$. Therefore, the second term on the RHS of  (\ref{eqn:Jacobian}) vanishes.

In the last term on the RHS of (\ref{eqn:Jacobian}), because $\bm{S}_i = (1,\bm{s}_i^T)^T$, it is sufficient to investigate the behavior of $\partial \bm{s}_i / \partial y_j$. From (\ref{eqn:esn-internal}), $\bm{s}_i$ depends only on the past trajectory, $\bm{s}_i = f(y_0,\cdots,y_{i-1},\bm{u}_0,\cdots\bm{u}_{i-1})$. Thus,
\[
\frac{\partial \bm{s}_i}{\partial y_j} = \bm{0},~~\text{for}~~ i \le j.
\]
For $i = j+1$,
\begin{align}
\frac{\partial \bm{s}_{j+1}}{\partial y_j} &= \frac{\partial}{\partial y_j}\left\{  \lambda \bm{s}_j + (1-\lambda)\Psi(\bm{s}_j,y_j,\bm{u}_j)\right\} \nonumber \\
&=(1-\lambda)\bm{Z}_j\bm{B}_y,
\end{align}
in which $\bm{Z}_j$ is a diagonal matrix,
\[
\text{diag}(\bm{Z}_j)_p = \cosh^{-2} \left( \sum_{q=1}^{N_s}A_{pq}s_{j_q} + B_{y_p}y_j + \sum_{q=1}^{N_u} B_{u_{pq}}u_{j_q} \right).
\]
Similarly, when $i > j+1$,
\begin{align}
\frac{\partial \bm{s}_i}{\partial y_j} &= \frac{\partial}{\partial y_j}\left\{  \lambda \bm{s}_{i-1} + (1-\lambda)\Psi(\bm{s}_{i-1},y_{i-1},\bm{u}_{i-1})\right\} \nonumber \\
&=\left\{ \lambda \bm{I}+(1-\lambda)\bm{Z}_{i-1}\bm{A}\right\}\left[ \left(\prod_{k=j+1}^{i-2} \frac{\partial \bm{s}_{k+1}}{\partial \bm{s}_k}\right) \frac{\partial \bm{s}_{j+1}}{\partial y_j} \right]^T \nonumber \\
&=(1-\lambda)\bm{K}^{i-1}_{j+1} \bm{Z}_j \bm{B}_y.
\end{align}
Here,
\begin{equation} \label{eqn:Klm}
\bm{K}_l^m =
\begin{cases}
\bm{I}, & \text{if}~~ m < l,\\
\prod_{k=l}^m \left[ \lambda \bm{I} + (1-\lambda)\bm{Z}_k\bm{A}\right], & \text{otherwise}.\\
\end{cases}
\end{equation}

To sum up, the Jacobian, $\mathcal{J}_\mathcal{G}(\bm{Y}^*)$, is a lower triangular matrix, of which element is
\begin{equation} \label{eqn:Jacobian_G}
\left[\mathcal{J}_\mathcal{G}(\bm{Y}^*)\right]_{ij} = 
\begin{cases}
0&~\text{if}~~ i < j, \\
\alpha & ~\text{if}~~ i = j,\\
(1-\alpha)(1-\lambda)\widetilde{\bm{\theta}}^T\bm{K}^{i-1}_{j+1} \bm{Z}^*_j\bm{B}_y & ~\text{if}~~i > j.
\end{cases}
\end{equation}
Here, $\widetilde{\bm{\theta}} = (\theta_2,\cdots,\theta_{N_s+1})$ is a submatrix of $\bm{\theta}$, which corresponds to $\bm{s}_t$.
The $l_1$-norm of $\mathcal{J}_\mathcal{G}(\bm{Y}^*)$ is defined as
\begin{align}
\|\mathcal{J}_\mathcal{G}(\bm{Y}^*)\|_1 &= \max_{1 \le j \le T} \sum_{i=1}^T \big| \left[\mathcal{J}_\mathcal{G}(\bm{Y}^*)\right]_{ij} \big| \nonumber \\
&= \max_{1 \le j \le T} \left\{ \alpha +  (1-\alpha)(1-\lambda)\sum_{i=j+1}^T \big| \widetilde{\bm{\theta}^*}^T \bm{K}^{i-1}_{j+1} \bm{Z}^*_j\bm{B}_y \big| \right\}.
\end{align}
Let $J$ be the column index of the maximum absolute summation. The sufficient condition for a local convergence, $\|\mathcal{J}_\mathcal{G}(\bm{Y}^*)\|_1 < 1$, yields
\begin{equation} \label{eqn:suff_cond}
   \sum_{i=J+1}^T \big| \widetilde{\bm{\theta}^*}^T \bm{K}^{i-1}_{J+1} \bm{Z}^*_{J}\bm{B}_y \big| < \frac{1}{1-\lambda}.
\end{equation}
However, evaluating (\ref{eqn:suff_cond}) a priori is not possible, because of its dependence on $\bm{\theta}^*$ and $\bm{Z}^*$.

The sufficient condition (\ref{eqn:suff_cond}) involves a summation over the length of the time series data, $T$, which is usually very long, e.g., $T\sim O(10^4)$. For (\ref{eqn:suff_cond}) to hold, $\bm{K}^{i-1}_{J+1}$, should be a monotonically decreasing sequence, which requires
\begin{equation}
\frac{\lambda+1}{\lambda-1} < \text{diag}(\bm{A})_i < 1,~~\text{for}~~1 \le i \le N_s.
\end{equation}

For (\ref{eqn:suff_cond}) to hold, it requires that $|\bm{K}^{i-1}_{J+1}| $, which dictates the sensitivity of the latent state in the future ($t > J$)  to the changes in the latent state at $t=J$,  decays rapidly in time. It is challenging to evaluate $\bm{K}^{i-1}_{J+1}$ for a general case. Since the temporal decay of $\bm{K}^{i-1}_{J+1}$ depends on the diagonal matrix $\bm{Z}_J$, where $0\le\text{diag}(\bm{Z}_J)_i \le 1$ for $i = 1,\cdots,N_s$, let assume $\bm{Z} = \bm{I}$. Also, note that, since $\bm{A}$ and $\bm{B}$ are generated by independent uniform distributions, 
\begin{equation}
\left[ \bm{A}\bm{B} \right]_i = \sum_{j=1}^{N_s} A_{ij}B_j = N_s E[ \zeta_A \zeta_B ] = N_s E[ \zeta_A ]E[\zeta_B ] = 0, ~\text{as}~N_s \rightarrow \infty,
\end{equation}
where $\zeta_A \sim \mathcal{U}(-\xi_A,\xi_A)$ and $\zeta_B \sim \mathcal{U}(-\xi_B,\xi_B)$. Then, the left hand side of (\ref{eqn:suff_cond}) is
\begin{equation}
 \sum_{i=J+1}^T \big| \widetilde{\bm{\theta}^*}^T   \bm{K}^{i-1}_{J+1} \bm{B}_y \big| = \left( \sum_{i=J+1}^T \lambda^{i-J-1} \right) \big| \widetilde{\bm{\theta}^*}^T  \bm{B}_y \big| = \frac{1-\lambda^{T-J}}{1-\lambda} \big| \widetilde{\bm{\theta}^*}^T  \bm{B}_y \big|.
\end{equation}
Now, the sufficient condition for the convergence is simply
\begin{equation}
(1-\lambda^{T-J}) \big| \widetilde{\bm{\theta}^*}^T  \bm{B}_y  \big| < 1.
\end{equation}
Note that $\widetilde{\bm{\theta}^*}^T  \bm{B}_y$ corresponds to a weighted average of a zero-mean independent uniform random variables, which behaves $\widetilde{\bm{\theta}^*}^T  \bm{B}_y \simeq 0$ as $N_s \rightarrow \infty$. Although the convergence analysis is performed in a limited scope, it shows the effects of the temporal relaxation parameter, $\lambda$, on the convergence. It is consistent with the finding in figure \ref{fig:MG_param} (a) that the fixed-point ESN fails to converge when $\lambda$ becomes small.
 

\bibliographystyle{elsarticle-num}
\bibliography{ESN_ref}

\end{document}